\documentclass[10pt,twocolumn,letterpaper]{article}

\usepackage[pagenumbers]{cvpr} 

\usepackage{cuted} 

\usepackage{currfile} 

\usepackage{caption} 

\usepackage{microtype}

\usepackage{placeins}
\usepackage{longtable}   

\usepackage{amsmath,amssymb,bm}
\usepackage{booktabs,multirow}
\usepackage{xspace}
\usepackage{siunitx}

\DeclareMathOperator*{\argmin}{arg\,min}

\DeclareMathOperator{\KL}{KL}

\newcommand{\full}{f}                                  
\newcommand{\gates}{\mathbf{m}}                         
\newcommand{\circuit}{f_{\gates}}                        
\newcommand{\pfull}{p_{\full}}
\newcommand{\pcirc}{p_{\gates}}
\newcommand{\opts}{\mathcal{A}}                         

\newcommand{\Dclean}{D_{\mathrm{clean}}}
\newcommand{\Dint}{D_{\mathrm{int}}}

\newcommand{\Dval}{\widehat{D}}                         
\newcommand{\size}{S}                                   

\newcommand{\tol}{\varepsilon}
\newcommand{\Fsur}{F}                                   

\newcommand{\roleS}{\textsc{s}}
\newcommand{\roleI}{\textsc{i}}
\newcommand{\roleV}{\textsc{v}}
\newcommand{\roleQ}{\textsc{q}}
\newcommand{\roleT}{\textsc{t}}
\newcommand{\roleR}{\textsc{r}}

\newcommand{\haidermean}{Haider-Mean\xspace}           
            
\newcommand{\gqa}{{GQA}\xspace}
\newcommand{\pope}{POPE\xspace}
\newcommand{\mme}{MME\xspace}
\newcommand{\aid}{AI2D\xspace}
\newcommand{\vsr}{VSR\xspace}
\newcommand{\qwen}[1]{Qwen3-VL-#1\xspace}
\newcommand{\gemma}{Gemma-3-4B\xspace}
\newcommand{\internvl}{InternVL3-8B\xspace}

\newcommand{\figbox}[2]{%
  \fbox{\parbox[c][#1][c]{\dimexpr\linewidth-2\fboxsep-2\fboxrule\relax}{\centering\small\color{gray!80!black}#2}}}

\newcommand{\figpdf}[3]{\IfFileExists{img/#1.pdf}{\includegraphics[width=\textwidth]{img/#1.pdf}}{%
  \IfFileExists{../img/#1.pdf}{\includegraphics[width=\textwidth]{img/#1.pdf}}{\figbox{#2}{#3}}}}

\definecolor{cvprblue}{rgb}{0.21,0.49,0.74}
\usepackage[pagebackref,breaklinks,colorlinks,allcolors=cvprblue]{hyperref}

\def\confName{CVPR}
\def\confYear{2027}

\title{Executable Decision Circuits of Multimodal Large Language Models}

\author{%
  Hikaru Shijo\textsuperscript{1}\quad
  Yutaka Yoshihama\textsuperscript{1}\quad
  Yasunori Ishii\textsuperscript{2}\quad
  Takayoshi Yamashita\textsuperscript{3}\\
  {\normalsize\textsuperscript{1}Panasonic Automotive Systems Co., Ltd.\quad
  \textsuperscript{2}Panasonic Holdings Corporation\quad
  \textsuperscript{3}Chubu University}\\
  {\footnotesize\texttt{\{shijo.hikaru, yoshihama.yutaka, ishii.yasunori\}@jp.panasonic.com}\quad
  \texttt{takayoshi@fsc.chubu.ac.jp}}%
}

\begin{document}
\raggedbottom
\setlength{\dbltextfloatsep}{10pt plus 2pt minus 2pt}
\setlength{\textfloatsep}{10pt plus 2pt minus 2pt}
\maketitle
\setlength{\stripsep}{8pt plus 2pt minus 2pt}
\begingroup
  \renewcommand{\figpdf}[3]{\includegraphics[width=0.95\textwidth]{img/#1.pdf}}
  \begin{strip}
\centering
\figpdf{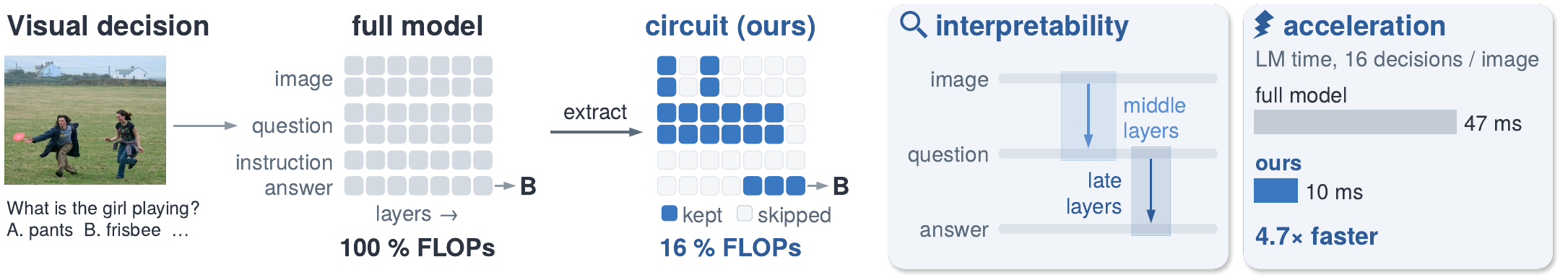}{4cm}{}
\captionsetup{hypcap=false}
\captionof{figure}{\textbf{Executable decision circuits for analysis and inference.} \qwen{2B}, \gqa, $\tol=0.25$. Blue: retained computation; grey: mean-ablated. Right: information flow and language-model time for 16 decisions per image.}
\captionsetup{hypcap=true}
\label{fig:teaser}
\end{strip}

\endgroup
\begin{abstract}
Multimodal large language models (MLLMs) typically execute their full computation even for discrete visual decisions.
To investigate the computation that reproduces visual decisions and its internal structure, we extract standalone decision circuits from MLLMs.
With the original model weights fixed, we selectively retain attention and MLP updates by token role and mean-ablate removed updates using precomputed role-wise means.
To preserve image-dependent decisions, we additionally use the original model's responses to image interventions as a learning signal.
Across five models and four benchmarks, the extracted circuits largely reproduce the original models' decisions on the target tasks while reducing language-model FLOPs by up to 95.9\%.
On Qwen3-VL-8B, for 16 decisions per image, the circuits accelerate language-model inference by up to 10.9$\times$ and end-to-end inference, including vision encoding, by up to 2.7$\times$.
Circuit analysis reveals that retained updates occupy different depths according to token role, while retained image-to-image reads vary across tasks and models.
The same extracted circuits enable both analysis of the computational structure that reproduces visual decisions and practical inference acceleration.
\end{abstract}

\section{Introduction}
\label{sec:intro}

Many visual tasks, including object existence, attribute recognition and relation recognition, can be solved as discrete decisions without generating textual descriptions.
Yet multimodal large language models (MLLMs) typically execute their full computation even for these decisions~\cite{visualjev2026}.
\textbf{How much of the original model's computation suffices to reproduce its visual decisions, and {how is that computation distributed across token roles and layers}?}
We extract computational subgraphs underlying visual decisions as standalone circuits, connecting analysis of their internal structure with inference acceleration (\cref{fig:teaser}).

To address these questions, we turn to circuit discovery, which identifies computational subgraphs that reproduce specific behaviours.
For language models, automatic circuit discovery methods select connections between components, attention heads and MLPs, and account for token positions~\cite{conmy2023acdc,bhaskar2024edgepruning,haider2026multigranular,haklay2025peap}.
Garc\'ia-Carrasco et al.~\cite{garciacarrasco2025circuits} further demonstrate the extraction of standalone task-specific circuits from language models.

Meanwhile, studies of MLLM internals examine visual information flow and component functions through attention analysis, activation patching and circuit tracing~\cite{zhang2025crossmodal,kaduri2025whatsinimage,li2026fcct,yang2026circuittracing}.
However, a component's contribution to decisions in the original model is distinct from the need to execute its updates during inference.
For visual decisions in MLLMs, which updates and information reads across roles and layers suffice to reproduce decisions remains underexplored from the joint perspectives of standalone circuit structure and measured inference efficiency.

We propose a method for extracting standalone, task-specific decision circuits from pretrained MLLMs that reproduce the original model's responses on a particular visual decision task.
Recognising the distinct roles of image, question and readout positions, we select attention-head and MLP updates separately by role, while sharing MLP neuron selection across roles.
This distinguishes which components are retained from the token roles at which their updates are executed.
We keep the original model weights fixed and replace removed updates with precomputed role-wise means, allowing the circuit to run without an additional full-model pass at inference time.
During extraction, we reduce retained computation subject to a fidelity constraint on the original model's answer distribution.
To preserve the original model's responses to image changes, we additionally use its answer distributions on inputs with replaced images as a learning signal.

We analyse retained updates and cross-role information reads to relate the computational structure underlying visual decisions to computation that can be omitted.

Across five models from Qwen3-VL, Gemma 3 and InternVL3 and four benchmarks, the extracted circuits largely reproduce the original model's decisions while retaining only 4.1--42.2\% of its language-model FLOPs.
Comparisons with existing circuit extraction and pruning methods show favourable fidelity--computation trade-offs.
On two 8B models, for 16 decisions per image, the circuits accelerate language-model inference by up to 10.9$\times$ and end-to-end inference, including vision encoding, by 1.7--2.7$\times$.

Question and readout computation follow a recurring depth arrangement across models and tasks, while image-to-image reads vary with both.

Our contributions are threefold:
{
\begin{enumerate}
    \item We introduce a novel framework for extracting standalone
    visual-decision circuits from MLLMs using role-wise selection
    and image interventions, without updating model weights.

    \item We identify reproducible role-wise depth patterns in
    the extracted circuits: earlier question heads and later
    readout heads, alongside task- and model-dependent visual
    computation.

    \item We demonstrate favourable fidelity--computation
    trade-offs and measured end-to-end inference speed-ups
    across five models and four benchmarks.
\end{enumerate}
}

\begin{figure*}[t]
\centering
\includegraphics[width=\textwidth]{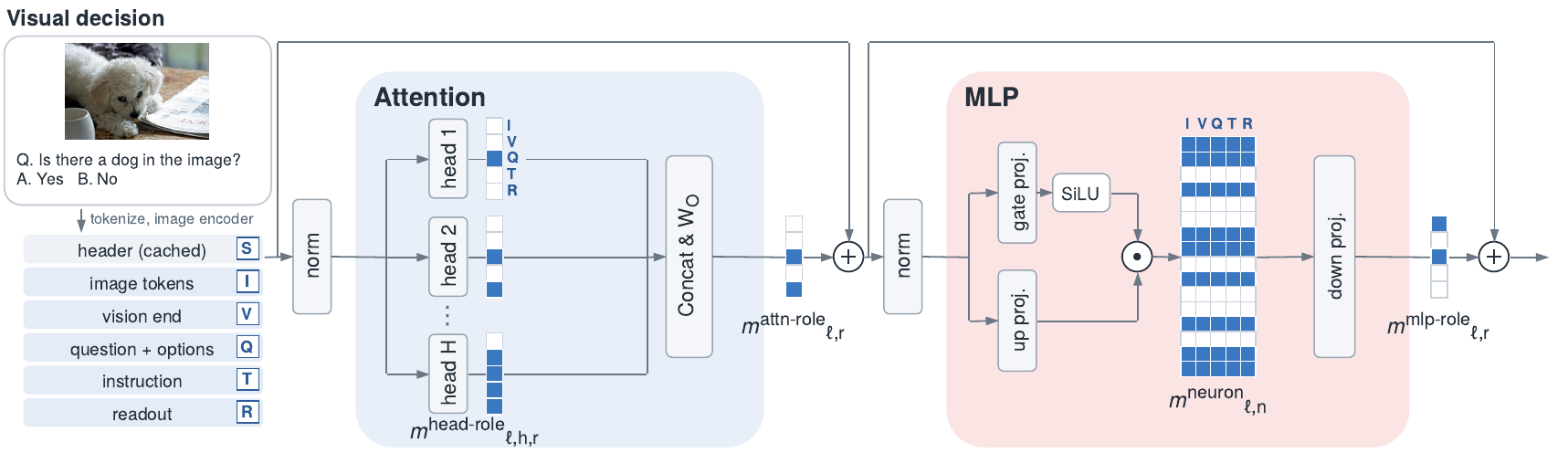}
\caption{{\textbf{Role-aware hierarchical circuit selection in one transformer layer.}
Left: token roles defined by the decision prompt (S, the fixed header, is cached and not gated).
Right: hierarchical gates select attention-head and MLP updates by role, while neuron selection is shared across roles.
Filled gate cells indicate retained updates; empty cells indicate replacement by precomputed role-wise means.
Shared attention-block, head, and MLP-block gates are omitted for clarity.}}
\label{fig:overview}
\end{figure*}

\section{Related Work}
\label{sec:related}

\paragraph{Circuit discovery.}
{Circuit discovery identifies computational subgraphs that reproduce specific model behaviours. Prior work on language models has explored edge-based selection~\cite{bhaskar2024edgepruning,conmy2023acdc}, multi-granular component selection~\cite{haider2026multigranular}, position-aware discovery~\cite{haklay2025peap}, and standalone circuit extraction~\cite{garciacarrasco2025circuits}. For visual decisions in MLLMs, matching the full model's outputs on ordinary inputs alone may not preserve its responses to image changes. We therefore learn role-conditioned, standalone circuits using image interventions as an additional fidelity signal while keeping the original model weights fixed. We analyse the retained computation and measure inference speed-ups using the same executable circuits.}

\paragraph{Understanding visual computation in MLLMs.}
Studies of MLLM internals examine visual--language information flow, object representations and the functions of attention heads and FFNs~\cite{golovanevsky2025notice,kaduri2025whatsinimage,li2026fcct,shukor2024skipping,zhang2025crossmodal}.
Zhang et al.~\cite{zhang2025crossmodal} analyse how visual information reaches readout positions through question positions; Nikankin et al.~\cite{nikankin2025samecircuits} compare circuits for visual and textual inputs to study modality-specific computation.
Yang et al.~\cite{yang2026circuittracing} trace computational paths with transcoders and attribution graphs and validate them through causal interventions.
Fan et al.~\cite{fan2026visualtokens} classify visual tokens as sink, dead and alive, showing that image-specific semantics concentrate in a subset of tokens and that internal visual computation is redundant on many standard tasks.
These studies provide insights into visual processing, component contributions and computation omission.
However, identifying a component's contribution does not imply that its computation must be executed during inference.
{We extract standalone executable subgraphs sufficient to largely reproduce visual decisions.}
Through these circuits, we analyse the internal structure of visual decision-making and connect it to practical improvements in inference efficiency.

\paragraph{Efficient MLLM inference.}
Research on efficient MLLM inference has explored visual-token reduction~\cite{chen2024fastv} and selective visual computation~\cite{vica2026,operator2026,yopo2024}.
ShortV~\cite{yuan2025shortv} stops updating image tokens in layers with small contributions, reducing visual computation.
Visual Jev~\cite{visualjev2026} shares image and common-context computation across multiple decisions.
Beyond these visual computations, we also reduce the question and readout computations required for each decision.
{
Structured pruning~\cite{an2024flap,ma2023llmpruner,men2024shortgpt,sun2024wanda,xia2022cofi,xia2024sheared} primarily aims to compress models while preserving performance, using recovery training when needed.
Such pruning techniques are also used in circuit discovery, which emphasises identifying and analysing computational subgraphs that reproduce specific behaviours of the original model.
From this perspective, we extract visual decision circuits with the original model weights fixed, analyse their computational structure, and empirically evaluate inference speed-ups from executing the same circuits.
\par}

\section{Method}
\label{sec:method}

We extract standalone visual-decision circuits by selecting computation per token role while keeping model weights fixed.
The method combines a fidelity constraint, role-wise gates with mean ablation, image interventions and validation-based circuit selection (\cref{fig:overview}).

\subsection{Problem formulation}
\label{sec:formulation}

We use a decision format based on Visual Jev~\cite{visualjev2026}, reading out answers to image--question pairs from the next-token probabilities of option letters.
Let $\full$ be the full model and $\circuit$ the circuit defined by a vector of structural gates $\gates$.
An input $x=(I,Q)$ is an image $I$ and a question $Q$; both models output a distribution over a given set of
answer options $\opts$, written $\pfull(\cdot\mid x)$ and $\pcirc(\cdot\mid x)$.  We measure fidelity to the full
model by the KL divergence between the two answer distributions,
\begin{equation}
  \Dclean(\gates)=\mathbb{E}_{x\sim\mathcal{D}}
  \bigl[\KL\bigl(\pfull(\cdot\mid x)\,\|\,\pcirc(\cdot\mid x)\bigr)\bigr].
  \label{eq:dclean}
\end{equation}
We specify a fidelity tolerance and seek to reduce circuit size subject to that constraint:
\begin{equation}
  \min_{\gates}\ \size(\gates)\quad\text{s.t.}\quad \Dclean(\gates)\le\tol.
  \label{eq:problem}
\end{equation}
Here $\size$ is a sparsity term based on retained component fractions (\cref{eq:size}) and $\tol$ the KL tolerance.
We learn candidates using continuously relaxed gates and select binary circuits whose validation KL is within the tolerance.
Retained parameter fractions are computed from the circuit itself, and fidelity is evaluated on independent test data.

\subsection{Role-aware hierarchical circuit space}
\label{sec:space}

Existing circuit discovery methods select connections between components~\cite{bhaskar2024edgepruning} or components at multiple granularities~\cite{haider2026multigranular}.
Building on the latter, we incorporate the MLLM prompt structure into the search space: even within one layer, different token roles may require different computation.
We assign each token position to one of six roles: the fixed pre-image header (\roleS), image tokens (\roleI), vision-end token (\roleV), question (\roleQ), fixed instruction and chat ending (\roleT), and the readout position whose next-token logits yield the answer (\roleR); see \cref{fig:overview}.
The prompt format determines the role partition; we learn which components to retain for each role.

\paragraph{Masks.}
We use $m$ for individual masks and $\widehat m$ for their effective products; the binarisation rule is given below.
Indices $\ell,h,r,n$ denote layer, head, token role and neuron. The hierarchy is
\begin{equation}
  \begin{gathered}
    \widehat m^{\mathrm{attn}}_{\ell,h,r}
      = m^{\mathrm{attn}}_{\ell}\,m^{\mathrm{head}}_{\ell,h}\,
        m^{\mathrm{attn\text{-}role}}_{\ell,r}\,m^{\mathrm{head\text{-}role}}_{\ell,h,r},\\
    \widehat m^{\mathrm{mlp}}_{\ell,r}
      = m^{\mathrm{mlp}}_{\ell}\,m^{\mathrm{mlp\text{-}role}}_{\ell,r},\quad
    \widehat m^{\mathrm{neuron}}_{\ell,n}
      = m^{\mathrm{neuron}}_{\ell,n}.
  \end{gathered}
  \label{eq:gates}
\end{equation}
The superscripts identify the gated components: \texttt{attn} and \texttt{mlp} control whole blocks, while \texttt{head} selects heads shared across roles.
\texttt{attn-role} controls all head updates at a role, \texttt{head-role} controls individual heads at that role, and \texttt{mlp-role} controls MLP use by role; \texttt{neuron} is shared across roles.
The effective neuron mask is $\widehat m^{\mathrm{mlp}}_{\ell,r}\widehat m^{\mathrm{neuron}}_{\ell,n}$.
Role-wise masks control updates; tokens with removed updates can still supply keys and values.
The fixed prefix \roleS\ cannot attend to the image or question, so its states are input-independent.
We leave its updates unmasked and precompute its per-layer keys and values for reuse; mask learning targets the remaining roles.

\paragraph{Role-wise mean ablation.}
Mean ablation replaces removed activations with
reference means~\cite{wang2023ioi}.
We apply it by token role, using full-model role means
$\bar h_{i,r}$ precomputed from calibration data.
During training, continuous gates interpolate between
activations and these fixed means:
\begin{equation}
  \tilde h_i = \widehat m_i\,h_i + (1-\widehat m_i)\,\bar h_{i,r}.
  \label{eq:mean}
\end{equation}
{Unlike counterfactual patching~\cite{haider2026multigranular}, fixed means~\cite{garciacarrasco2025circuits} let the circuit skip removed updates without a full-model pass.}

\paragraph{Gate learning.}
Each gate follows a hard-concrete distribution~\cite{louizos2018l0} with learnable parameters $\theta$ and is sampled as a continuous relaxation during training.
At validation and inference, gates with nonzero probability above $1/2$ are set to 1, and the rest to 0 ({gate details in the Supplementary Material}).

\paragraph{Sparsity regularisation.}
For binary $\gates$, the retained fractions of role-wise heads, MLPs and shared neurons are
\begin{equation}
\begin{gathered}
\size_H(\gates)
 = \frac{\sum_{\ell,h,r}\widehat m^{\mathrm{attn}}_{\ell,h,r}}{LH|\mathcal R_g|},\\
\size_M(\gates)
 = \frac{\sum_{\ell,r}\widehat m^{\mathrm{mlp}}_{\ell,r}}{L|\mathcal R_g|},\quad
\size_N(\gates)
 = \frac{\sum_{\ell,n}m^{\mathrm{neuron}}_{\ell,n}}{LD}.
\end{gathered}
\label{eq:sparsity_groups}
\end{equation}
Here $L,H,D$ denote layers, heads per layer and MLP width; $\mathcal R_g$ excludes the fixed prefix \roleS. Sums cover each index's full range; shared neurons are counted once. The sparsity term is
\begin{equation}
  \size(\gates)=\frac{\size_H(\gates)+\size_M(\gates)+\lambda_N\,\size_N(\gates)}{2+\lambda_N}
  \label{eq:size}
\end{equation}
with $\lambda_N=3$ ({weight selection in the Supplementary Material}).
Training uses the expectation $\bar S(\theta)$, obtained by replacing each gate with its activation probability; candidate selection uses the binary circuit's $\size(\gates)$.
The retained parameter fraction $P$ used in evaluation counts the weights in use without duplication across roles.

\subsection{Image-intervention regularisation}
\label{sec:intervention}

Output matching on ordinary inputs may not preserve image-dependent responses.
We therefore keep questions and options fixed, intervene on image tokens, and match the full model's intervened output distribution.
For $\tilde x=T_\xi(x)$, we uniformly sample full image-token swaps, partial swaps of a random token fraction, or mean-token replacement. Intervention fidelity is
\begin{equation}
  \Dint(\gates)=\mathbb{E}_{x,\xi}
  \bigl[\KL\bigl(\pfull(\cdot\mid\tilde x)\,\|\,\pcirc(\cdot\mid\tilde x)\bigr)\bigr].
  \label{eq:dint}
\end{equation}
The training fidelity loss covers both inputs:
\begin{equation}
  \Fsur(\gates)=\Dclean(\gates)+\Dint(\gates),
  \label{eq:surrogate}
\end{equation}
Only clean validation KL $\Dclean^{\mathrm{val}}\le\tol$ determines feasibility.

\subsection{Constraint-adaptive optimisation and circuit selection}
\label{sec:solver}

\paragraph{Adaptive sparsity coefficient.}
Every $K$ updates, we measure the binary circuit's clean validation KL $\Dval_t$ and set
a weight inspired by the log-barrier derivative~\cite[Chapter~11]{boyd2004convex}:
\begin{equation}
  W_t=
  \begin{cases}
    \min\!\bigl(1,\ \dfrac{\tau}{\tol-\Dval_t}\bigr), & \Dval_t<\tol,\\[8pt]
    1+\kappa\,\dfrac{\Dval_t-\tol}{\tol}, & \Dval_t\ge\tol,
  \end{cases}
  \label{eq:weight}
\end{equation}
Here $\tau=0.002$, $\kappa=10$ and $W_0=\tau/\tol$. As validation KL approaches the tolerance, $W_t$ grows and reduces the sparsity coefficient $\rho_t/W_t$. It is fixed without gradients until the next check.

\paragraph{Training loss.}
Writing the training gates as $\gates_\theta$, the loss at step $t$ is
\begin{equation}
  \mathcal{L}_t(\theta)=\mathbb{E}_{\gates_\theta}\!\left[\Fsur^{\mathrm{train}}(\gates_\theta)\right]+\frac{\rho_t}{W_t}\,\bar S(\theta),
  \label{eq:loss}
\end{equation}
The first term measures output fidelity, and the second is weighted sparsity regularisation.
We estimate fidelity gradients using training batches and sampled gates, and linearly ramp $\rho_t$ from 0 to 1 over the first 1{,}500 steps.

\paragraph{Circuit selection.}
Let $\mathcal{M}=\{\gates_1,\dots,\gates_T\}$ be the saved binary circuits and
$\mathcal{M}_{\mathrm{feas}}=\{\gates_t\in\mathcal{M}:\Dval_t\le\tol\}$ the validation-feasible candidates.
We select the candidate with the smallest binary sparsity term $\size$:
\begin{equation}
  \gates^\ast=\argmin_{\gates_t\in\mathcal{M}_{\mathrm{feas}}}\size(\gates_t),
  \label{eq:pick}
\end{equation}
with ties broken by the earlier check.
Final evaluation uses test data disjoint from validation.
The selected circuit satisfies the validation KL bound $\tol$, but its test KL is not guaranteed to satisfy the same bound.

\paragraph{Read gates (analysis stage).}
\label{sec:reads}
With $\gates^\ast$ and its means fixed, we learn head-shared gates $z_{\ell,s\to r}$ per layer and ordered role pair; closing a gate removes role-$r$ queries' attention to role-$s$ keys.
We use the same loss with the mean open-gate fraction as sparsity for 3,000 updates, selecting the fewest open reads among candidates whose validation KL is at most the node circuit's value plus one standard error.

\paragraph{Execution.}
With the gates and means fixed, the executor skips removed heads, MLPs and neurons and the keys and values that no
retained head reads (\cref{fig:overview}); FLOPs, parameters and latency are evaluated in \cref{sec:experiments}.

\section{Experiments}
\label{sec:experiments}
We evaluate whether extracted circuits reproduce visual decisions with less computation and faster inference, and examine the contributions of role-wise masks and image interventions.

\begin{figure*}[t]
\centering
\includegraphics[width=0.85\textwidth]{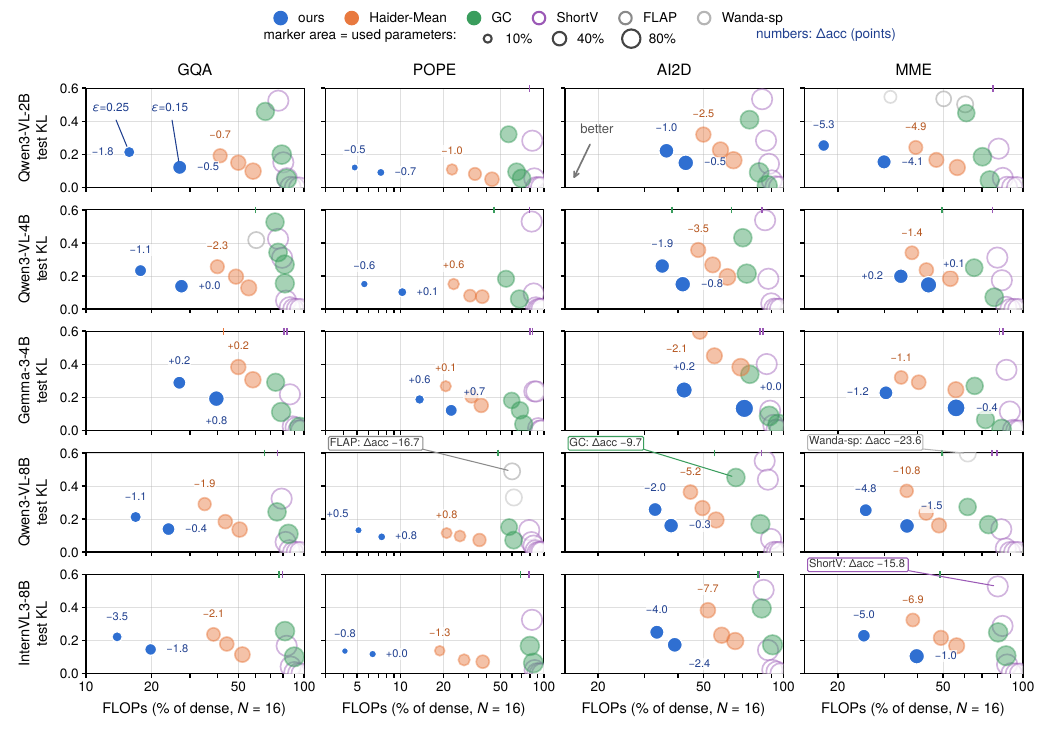}
\caption{\textbf{Fidelity against FLOPs.} Lower left is better. Test KL versus {retained language-model matrix-product FLOPs, excluding padding} (\% of dense, $N=16$). Marker area: used parameters. Blue: ours at $\tol=0.25$ (left) and $0.15$ (right); \haidermean: standalone mean-ablation adaptation. Numbers: accuracy change (points) of ours and of the smallest \haidermean circuit; boxes: for FLAP, Wanda-sp, GC and ShortV, the largest drop among that method's plotted points (KL $\le 0.6$) in rows 2--5, one panel per method. Column scales are shared; KL above 0.6 is clipped (FLAP, Wanda-sp and random omitted there; {all points are reported in the Supplementary Material}).}
\label{fig:frontier}
\end{figure*}

\subsection{Setup}
\label{sec:setup}
\paragraph{Models and data.}
We evaluate \qwen{2B}, \qwen{4B}, \qwen{8B}~\cite{bai2025qwen3vl}, \gemma~\cite{gemmateam2025gemma3} and \internvl~\cite{zhu2025internvl3} on {GQA}~\cite{hudson2019gqa}, \pope~\cite{li2023pope}, \aid~\cite{kembhavi2016ai2d} and \mme~\cite{fu2025mme}. Predictions are the highest-probability answer-option letters. Training, validation and test images are disjoint; training targets 3,000 decisions per task (950 for MME). MME uses a custom image-level split, so its accuracy is not the official aggregate score. Splits and prompts are detailed in {the Supplementary Material}.

\paragraph{Protocol and metrics.}
We use validation tolerances $\tol\in\{0.15,0.25\}$ and share hyperparameters and each model's GQA-derived training budget across tasks ({training details in the Supplementary Material}). Fidelity is measured by test KL (full model $\|$ circuit); task performance is measured by accuracy change in percentage points. Clean KL uses ordinary inputs, while Swap KL uses a different image with the question and options unchanged. Both compare the full model and circuit on the same input.
As in Visual Jev~\cite{visualjev2026}, we evaluate a setting with multiple independent decisions per image and shared visual computation.
Our main comparisons use $N=16$ decisions per image, evaluating FLOPs, retained parameter fractions and measured inference time.
Results for other values of $N$ and measurement details are provided in {the Supplementary Material}.

\subsection{Baselines}
\label{sec:baselines}
All comparisons freeze the model weights: mask learning is allowed, weight-updating recovery (including adapters) is not, since it mixes extraction with adaptation.
The original method of Haider et al.~\cite{haider2026multigranular} uses activations from corrupted inputs as replacement values and therefore does not directly yield standalone executable circuits.
To match the standalone execution setting, we replace these values with precomputed, fixed role-wise means and evaluate the resulting adaptation as \haidermean.
We also compare with GC~\cite{garciacarrasco2025circuits}, FLAP~\cite{an2024flap}, Wanda-sp~\cite{an2024flap,sun2024wanda}, ShortV~\cite{yuan2025shortv} and size-matched random circuits. Methods share the executor but retain their specified replacement rules. Discovery budgets and adaptations are reported in {the Supplementary Material}.

\begin{figure*}[t]
\centering
\includegraphics[width=\textwidth]{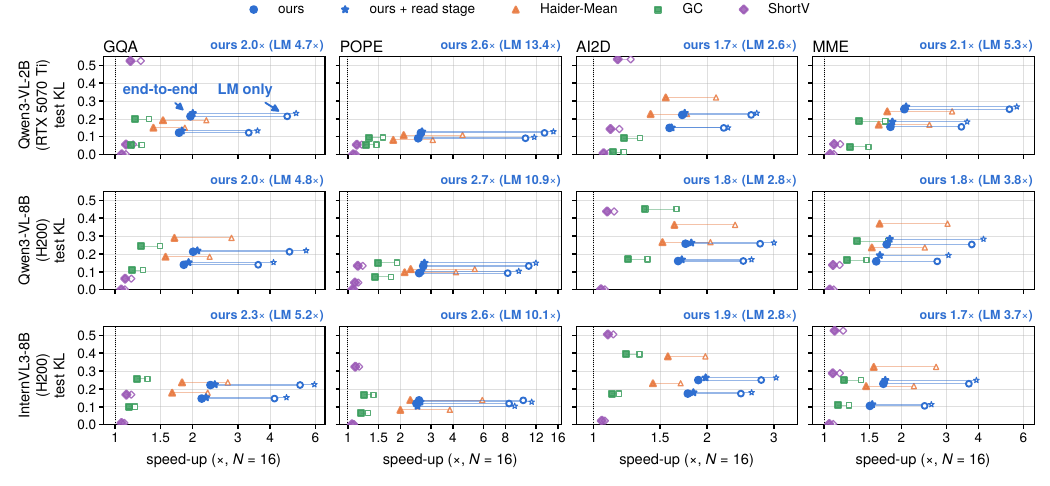}
\caption{\textbf{Fidelity against measured speed.} $N=16$ decisions per image; {Qwen3-VL-2B}: RTX 5070 Ti; 8B models: H200. Filled: end-to-end (including vision); hollow: language model. Circles: ours ($\tol=0.25,0.15$); stars: with read selection; diamonds: ShortV; triangles: \haidermean; squares: GC. Numbers above each panel: our end-to-end and language-model speed-ups at $\tol=0.25$. Horizontal scales differ by task; KL above 0.55 (ShortV with the most frozen layers) lies outside the panels. Circuit and dense reference share compilation and CUDA-graph policies.  At $N=1$, our node circuits achieve {1.35--1.78}$\times$ end-to-end speed-up ($\tol=0.25$; {full results in the Supplementary Material}).}
\label{fig:speed}
\end{figure*}

\subsection{Fidelity and computation}
\label{sec:main_results}
\Cref{fig:frontier} shows the relationship between computation and fidelity (lower left indicates fewer FLOPs and lower test KL).
Our method achieves fidelity comparable to the baselines with less computation.
For example, on {Qwen3-VL-8B} GQA, our circuit retains 16.9\,\% of the full model's FLOPs at KL 0.212.
With the same training budget, \haidermean retains 43.5\,\% at KL 0.184 and 35.0\,\% at KL 0.290.
ShortV retains most computation in settings that preserve fidelity, while FLAP and Wanda-sp lose fidelity under substantial pruning (details in {the Supplementary Material}).

{
The two tolerances illustrate the fidelity--computation trade-off.
At $\tol=0.15$, the circuits retain 6.4--71.2\,\% of language-model FLOPs while maintaining accuracy within approximately one point of the full model in 16 of 20 settings. Relaxing $\tol$ to 0.25 reduces retained FLOPs to 4.1--42.2\,\% at higher test KL in all 20 settings.
}
Retained computation also varies by task: at $\tol=0.25$, POPE retains 4.1--13.6\,\%, compared with 32.8--42.2\,\% for AI2D.
We analyse the circuit structures associated with these task differences in \cref{sec:analysis}.
Across {five} models and four tasks, evaluation with three seeds per setting yields standard deviations of {0.0}--2.2 percentage points in retained FLOPs and {0.002}--0.026 in test KL, indicating limited seed variability in both computation and fidelity ($\tol=0.25$; {full results in the Supplementary Material}).

{
At $\tol=0.15$, the largest accuracy difference is 4.1 points (Qwen3-VL-2B MME), as the constraint targets the full model's answer distribution rather than accuracy; a detailed analysis is provided in the Supplementary Material.
\par}

\begin{figure*}[t]
\centering
\includegraphics[width=\textwidth]{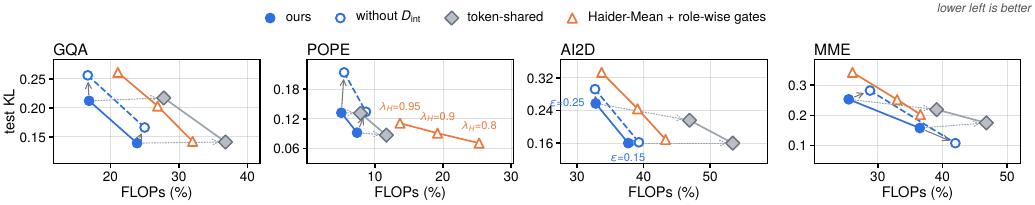}
\caption{\textbf{Ablations on \qwen{8B}.} Test KL against FLOPs. Filled: ours (role-wise gates with $\Dint$) at $\tol=0.25$ (left point) and $0.15$; hollow: without $\Dint$, arrows connecting equal tolerances; grey diamonds: token-shared gates at $\tol=0.25$ and $0.15$ (dotted arrows from ours at the same tolerance); orange triangles: \haidermean with our role-wise gates ($\lambda_H=0.95, 0.9, 0.8$ from left; same budget and means). Lower left is better; KL ranges differ by task. Test KL uses ordinary inputs.}
\label{fig:circuit_map_roles}
\label{fig:ablation_lint}
\end{figure*}

\ifdefined\analysisfloatgroup\else\begin{figure*}[t]\fi
\centering
\includegraphics[width=\textwidth]{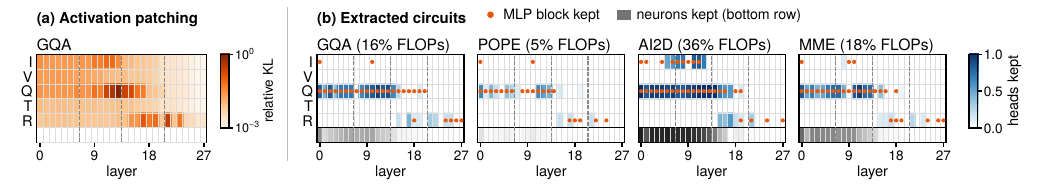}
\caption{\textbf{Patching importance and retained updates.} \qwen{2B}; rows: roles, columns: layers.
(a) \gqa: KL induced by counterfactual-image head patching, summed over heads and normalized to the map's maximum (log scale).
(b) Circuits at $\tol=0.25$: blue, retained head fraction; orange dots, retained MLP blocks; bottom strip, shared neuron retention.}
\label{fig:circuit_map}
\ifdefined\analysisfloatgroup\else\end{figure*}\fi

\subsection{Inference speed}
\label{sec:speed}
The computational savings of the extracted circuits translate into measured inference speed-ups.
We compare each circuit with an all-units-active reference on the same GPU under identical compilation and CUDA-graph policies.
{On Qwen3-VL-8B GQA at $N=16$, the all-units-active executor's language-model latency differs from Hugging Face by less than 2\,\% under CUDA Graph replay on H200 (Supplementary Material).}
{The circuit's executor gives the same decisions as the hook-based evaluation of that circuit on 99.5\,\% of all 4{,}402 test decisions ($\tol=0.25$).}
At $N=16$ and $\tol=0.25$, end-to-end inference, including the vision encoder, accelerates by 1.7--2.7$\times$ on the two 8B models and 1.7--2.6$\times$ on {Qwen3-VL-2B} (\cref{fig:speed}).
The language-model component reaches up to 10.9$\times$ on the 8B models, while the unchanged vision encoder limits the end-to-end gain.
{At $\tol=0.15$, the two 8B models still achieve 1.5--2.6$\times$ end-to-end speed-ups at $N=16$.}
{At $N=1$ and $\tol=0.25$, end-to-end speed-ups reach 1.35--1.78$\times$ across the three models.
For Qwen3-VL-8B on H200, they reach 74--89\,\% of the bound assuming zero language-model latency and unchanged vision encoding (Supplementary Material).}
{On Qwen3-VL-8B, the circuits also reduce peak GPU memory by 38--69\,\% {at $N=16$ and $\tol=0.25$}.}

Applying the read gates learned for analysis (\cref{sec:reads}) to execution further reduces language-model inference time by {7--14\,\%} on the two 8B models by skipping unused key and value computation ($N=16$, $\tol=0.25$; stars in \cref{fig:speed}).
Test KL changes by $-0.01$ to $+0.03$ relative to the circuit before read selection.
Thus, information reads identified as removable through circuit analysis can also be omitted to further accelerate inference.

\subsection{Ablations}
\label{sec:ablations}
{Role-wise selection reduces FLOPs at similar fidelity (\cref{fig:circuit_map_roles}). On Qwen3-VL-8B GQA at $\tol=0.25$, token-shared masks retain 27.8\,\% of FLOPs at KL 0.217, versus 16.9\,\% at KL 0.212 with role-wise masks.}
Even when \haidermean uses the same role-wise masks, our method requires fewer FLOPs at similar KL. The role-wise search space alone cannot explain this gap.
Removing $\Dint$ leads to poorer fidelity and increased FLOPs in many settings; on GQA at $\tol=0.25$, removing the intervention term raises Clean KL from 0.212 to 0.256 and Swap KL from 0.354 to 0.558.
These results support the contributions of role-wise selection and image interventions to extracting efficient, faithful circuits. We next analyse their internal structure through retained updates and reads.

\section{Understanding the Extracted Circuits}
\label{sec:analysis}
\ifdefined\analysisfloatgroup\else\begin{figure*}[t]\fi
\centering
\includegraphics[width=\textwidth]{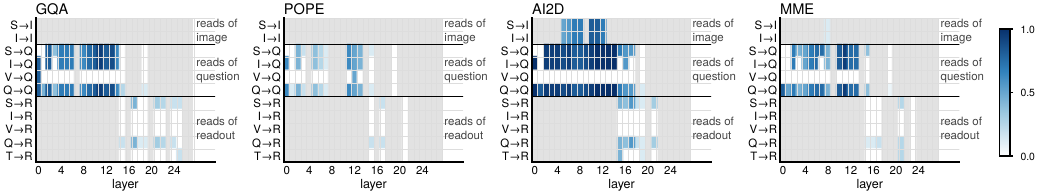}
\caption{\textbf{Selected attention reads.} \qwen{2B} on \gqa, \pope, \aid and \mme; node masks and means fixed. Rows: source $\to$ reader; columns: layers; colour: open gate $\times$ retained reader-head fraction (not attention weight); grey: no reader head. Inactive V/T readers are omitted. {Read counts and validation KL are reported in the Supplementary Material.}}
\label{fig:read_map}
\ifdefined\analysisfloatgroup\else\end{figure*}\fi

We analyse computation placement and information access in the extracted circuits and compare them with intervention results from the original model.

\smallskip
\noindent\textbf{Importance maps and sufficient circuits.}
\Cref{fig:circuit_map}(a) shows broadly distributed activation-patching importance for \qwen{2B} on \gqa, whereas computation is selectively retained in the extracted circuits (\cref{fig:circuit_map}(b)).
We select heads and MLPs by their importance rankings measured with the same patching procedure and construct circuits using our gate hierarchy and fixed means.
{Under the same validation set, gate hierarchy, replacement means and FLOPs accounting, importance-ranked circuits satisfying validation KL $\le 0.25$ retain 51--76\,\% of FLOPs across four tasks, compared with 5--36\,\% for our learned circuits (Supplementary Material). On {GQA}, the difference is 53.6\,\% versus 15.8\,\%. {These comparisons suggest that identifying compact sufficient circuits requires considering combinations of components beyond their individual importance rankings.}}

\smallskip
\noindent\textbf{Where is computation retained?}
The extracted circuits tend to place question and readout computation at different depths.
For \qwen{2B} in \cref{fig:circuit_map}(b), heads for Q, which includes questions and answer candidates, remain mainly in early to middle layers (0--19), whereas readout heads for R appear from {layer 15} onwards.
Across 20 circuits spanning five models and four tasks, Q and R head centroids lie at 22--37\,\% and 56--81\,\% of model depth, respectively; all instruction (T) and vision-end (V) updates are replaced by fixed means in 18 circuits ({complete circuit maps in the Supplementary Material}).

Image-position updates are omitted in many layers, and the retained layers vary by task and model.
On \gqa and \pope, \qwen{2B} retains no image-position attention updates, and MLP updates remain only in layers 0 and 10 in both tasks.
In contrast, \aid retains image heads in middle layers.
\gemma retains image heads in multiple layers on every task.
{The Q/R depth separation recurs across all 60 circuits from three extraction seeds, while visual computation varies across tasks and models (Supplementary Material).}

\smallskip
\noindent\textbf{How is visual information accessed?}
Tokens whose updates are removed can still supply keys and values.
We therefore freeze the node circuits and ablation means and additionally learn cross-role attention read gates (\cref{sec:reads}).
For \qwen{2B}, read groups decrease by 35--49\,\% across four tasks, while validation KL remains within the analysis-stage bound (\cref{fig:read_map}).
In all four tasks, Q reads the image in early to middle layers, and R reads Q in later layers.
{The same Image$\to$Q followed by Q$\to$R arrangement appears in the other four models' extracted circuits (Supplementary Material).
Zhang et al.~\cite{zhang2025crossmodal} identified this information-flow pattern through attention knockout in full models; here, it is retained in standalone circuits that largely reproduce the original decisions.}
Image-to-image reads, however, differ across tasks.
On \gqa and \pope, circuits with no image attention updates retain Image$\to$Q reads and largely reproduce the original model's decisions.
In contrast, \aid retains image-to-image reads across layers 5--12, which may support information integration between diagram elements.

\begin{table}[t]
\centering
\scriptsize
\setlength{\tabcolsep}{3.5pt}
\caption{\textbf{Effects of read cuts and update ablation.} Decisions changed (\%), full model / circuit
($\tol=0.25$; {all test images; 95\,\% bootstrap intervals over images are within $\pm$3 points}). Reads are masked at attention
scores; updates before the attention output and MLP down projections are replaced by fixed role means.
{Middle / late / late+final layers: 7--14 / 15--21 / 15--27 (\qwen{2B}), 9--18 / 19--27 / 19--35 (\qwen{8B}); all four tasks in the Supplementary Material.}}
\label{tab:pathcut}
\resizebox{\ifdim\width>\columnwidth\columnwidth\else\width\fi}{!}{\begin{tabular}{lrrrr}
\toprule
Decision changes (\%) & \multicolumn{2}{c}{2B} & \multicolumn{2}{c}{8B} \\
\cmidrule(lr){2-3}\cmidrule(lr){4-5}
Full model / Circuit & \gqa & \aid & \gqa & \aid \\
\midrule
\multicolumn{5}{l}{\emph{attention reads cut} (reader $\leftarrow$ source, layers)} \\
Q$\leftarrow$I, middle & 31 / 31 & 32 / 31 & 27 / 32 & 23 / 26 \\
R$\leftarrow$Q, late & 27 / 33 & 41 / 71 & 15 / 40 & 33 / 72 \\
R$\leftarrow$I, all & 1 / 1 & 3 / 2 & 1 / 0 & 1 / 1 \\
\multicolumn{5}{l}{\emph{updates replaced} (role, layers $\to$ value)} \\
T, all $\to$ mean & 10 / 0 & 14 / 0 & 9 / 0 & 13 / 0 \\
Q, middle $\to$ mean & 62 / 64 & 73 / 73 & 59 / 63 & 69 / 71 \\
R, late+final $\to$ mean & 64 / 64 & 76 / 75 & 63 / 62 & 71 / 72 \\
\bottomrule
\end{tabular}
}
\end{table}

\smallskip
\noindent\textbf{Comparing intervention responses.}
To examine whether decision-relevant computation corresponds between the extracted circuits and the original model, we block attention reads and replace role-specific updates with fixed means in both, measuring decision changes relative to each system's pre-intervention decisions (\cref{tab:pathcut}{; all four tasks in the Supplementary Material}).
{On \qwen{2B} and \qwen{8B}}, blocking Image$\to$Q or Q$\to$R reads substantially changes decisions in both systems, whereas blocking direct Image$\to$R reads changes {at most 3\,\%}.
{Replacing T updates with fixed means changes 9--14\,\% of the original models' decisions but none of the circuits'.
By contrast, replacing Q or R updates changes 59--76\,\% of full-model decisions and 62--75\,\% of circuit decisions.
}
{Cutting Q$\to$R reads changes more decisions in the circuits
(33--72\,\%) than in the full models (15--41\,\%),
particularly on Qwen3-VL-8B.
This increased sensitivity may reflect the removal of alternative computations available in the full model, leaving the extracted circuits more dependent on the retained Q$\to$R reads.
}

\section{Conclusion}
\label{sec:conclusion}
We introduced a framework for extracting standalone, executable decision circuits from pretrained MLLMs without updating their weights. Across five models and four benchmarks, the circuits largely reproduce the original models' visual decisions with substantially reduced language-model computation. This suggests that substantially less language-model computation can suffice to reproduce many of the evaluated visual decisions than is performed by the original MLLM. These compact circuits retain a recurring role-wise organization and an Image$\to$Question$\to$Readout information flow, while image-position computation varies across tasks and models. Because the same circuits are directly executable, they connect the study of sufficient computation for visual decisions with practical inference acceleration.

\paragraph{Limitations.}
We do not establish uniqueness or global minimality of the extracted circuits. Their interpretation depends on the role partition, mean-ablation scheme, and target distribution. Our analysis is restricted to static circuits within the language-model component. Future work could combine circuit extraction with input-adaptive visual-token pruning and extend extraction to the vision encoder.

\section*{LLM Usage Statement}

We used OpenAI ChatGPT, OpenAI Codex, and Anthropic Claude Code to assist with research ideation, discussions of experimental design and analysis, literature review, manuscript drafting and language editing, and code development and implementation, including figure-generation code. The authors reviewed and validated the scientific claims, references, and experimental results and take full responsibility for the content of this paper.

{
    \small
    \bibliographystyle{ieeenat_fullname}
    \bibliography{main}
}

\clearpage
\appendix
\maketitlesupplementary
\let\subsection\section

\renewcommand{\dbltopfraction}{0.95}
\renewcommand{\dblfloatpagefraction}{0.7}
\setcounter{dbltopnumber}{3}
\subsection{Reproducibility settings}
\label{sec:suppl_datasets}
\label{sec:suppl_exp_details}
\begin{table*}[t]
\centering
\small
\setlength{\tabcolsep}{8pt}
\caption{\textbf{Dataset splits used for circuit extraction and evaluation.}
Each split reports images / questions. Validation data are used for adaptive weighting and circuit selection;
test data are used for the final evaluation. Choices denotes the number of answer options per question.
Image counts refer to the image units loaded by the experiment.}
\label{tab:datasets}
\begin{tabular}{lrrrr}
\toprule
Dataset & Train (images / questions) & Val (images / questions) & Test (images / questions) & Choices \\
\midrule
{GQA} & 583 / 3{,}003 & 98 / 502 & 863 / 4{,}402 & 2--8 \\
POPE & 500 / 3{,}000 & 100 / 600 & 500 / 9{,}000 & 2 \\
MME & 475 / 950 & 119 / 238 & 593 / 1{,}186 & 2 \\
AI2D & 788 / 3{,}010 & 328 / 1{,}251 & 814 / 3{,}088 & 4 \\
\bottomrule
\end{tabular}
\end{table*}

\paragraph{Datasets and splits.}
\Cref{tab:datasets} summarises the image and question counts. Training targets approximately 3,000 questions per task (950 for MME), keeping questions from the same image together. Validation data are used for optimisation and circuit selection, while test data are reserved for evaluation.
For GQA, we follow Visual Jev's multiple-choice construction and scored test split~\cite{hudson2019gqa,visualjev2026}, excluding calibration images. Questions have 2--8 answer options. POPE uses the official random, popular and adversarial test sets~\cite{li2023pope}, with training and validation questions generated from separate COCO images following the POPE procedure.
For MME~\cite{fu2025mme}, we partition images within each of its 14 subtasks into approximately 40\% training, 10\% validation and 50\% testing, keeping both questions per image together. Reported accuracy is therefore based on our held-out test subset rather than the official aggregate score. For AI2D~\cite{kembhavi2016ai2d}, we use the publicly released test set for evaluation and hold out validation diagrams from the training data.

\begin{figure*}[t]
\centering
\includegraphics[width=0.92\textwidth]{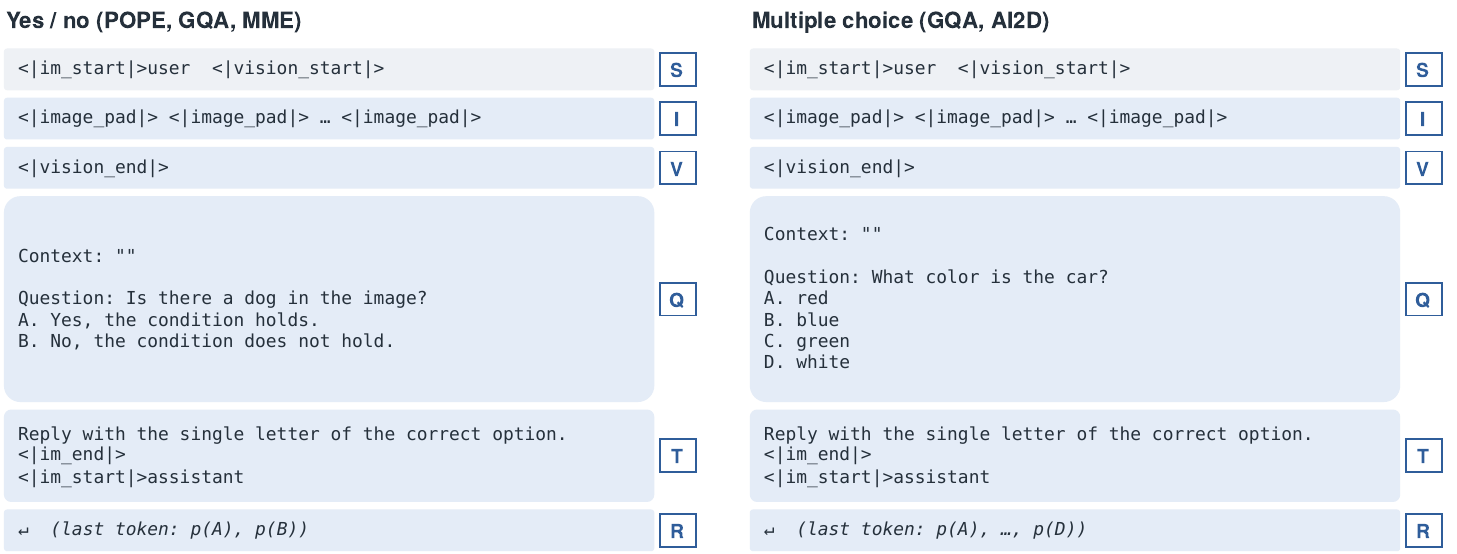}
\caption{\textbf{Decision prompts} (format A, Qwen3-VL chat tokens), {labelled by token role} as in \cref{fig:overview}: the fixed
header (S), image tokens (I), vision-end token (V), the question block with an empty \texttt{Context} line and lettered options (Q),
the fixed instruction and chat ending (T) and the readout token (R), whose next-token probabilities over the option letters are
the decision.  Yes/no questions use the two sentence options; {GQA multiple-choice questions} have 2--8 options and AI2D four.
{Gemma-3-4B} and {InternVL3-8B} wrap the same text in their own chat templates.}
\label{fig:prompts}
\end{figure*}

\paragraph{Prompt and readout.}
\label{sec:suppl_prompt}
Each prompt contains a model-specific chat header, image, empty \texttt{Context} line, question, lettered options and a fixed instruction. The decision uses next-token probabilities over option letters. Yes/no questions use two sentence options; AI2D has four options and GQA choice/query questions have 2--8 (\cref{fig:prompts}).

{Gemma-3-4B} and {InternVL3-8B} wrap the same text in their own chat templates. Following Visual Jev~\cite{visualjev2026}, we use a decision prompt, omitting its system prompt and trailing \texttt{Answer:} line.

\paragraph{Gates and calibration.}
\label{sec:suppl_gates}
\label{sec:suppl_impl}
Each hard-concrete gate has a learnable $\log\alpha_i$~\cite{louizos2018l0}:
\begin{equation}
  \begin{aligned}
    s_i &= \sigma\!\Bigl(\tfrac{\log u-\log(1-u)+\log\alpha_i}{t}\Bigr),\qquad u\sim\mathcal{U}(0,1),\\
    m_i &= \operatorname{clip}_{[0,1]}\bigl(s_i(\zeta-\gamma)+\gamma\bigr),
  \end{aligned}
  \label{eq:hc}
\end{equation}
We fix $t=2/3$, $\gamma=-0.1$ and $\zeta=1.1$. The activation probability is
$P(m_i>0)=\sigma(\log\alpha_i-t\log(-\gamma/\zeta))$; validation and inference keep gates with probability above $1/2$.
Role-wise replacement means are calibrated on 30 training images. Clean teacher outputs are cached; the teacher is evaluated on the same intervened input as the circuit when computing $\Dint$. Model weights stay frozen throughout.

\paragraph{Optimisation and selection.}
\label{sec:suppl_solver}
We use Adam~\cite{kingma2015adam}, learning rate 0.02, $\lambda_N=3$, $\tau=0.002$, $\kappa=10$ and a 1{,}500-step sparsity ramp. Each update uses one image's decisions, packed into question blocks. Binary validation runs every 200 updates; $W_t$ is held fixed until the next check, starting at $W_0=\tau/\tol$. Both tolerances, $\tol\in\{0.15,0.25\}$, share these settings. {On one RTX 5070 Ti, extracting a \qwen{2B} circuit (20{,}000 updates, including the validation checks) takes 1.3--1.9 hours per task, and the read stage (\cref{sec:reads}) {11--18 minutes}; on one H200, a \qwen{8B} circuit (\gqa, 25{,}000 updates) takes 1.7 hours and its read stage 12 minutes.}
\paragraph{Choosing the neuron weight.}
\label{sec:suppl_lambdan}
\begin{figure*}[t]
\centering
\includegraphics[width=0.70\textwidth]{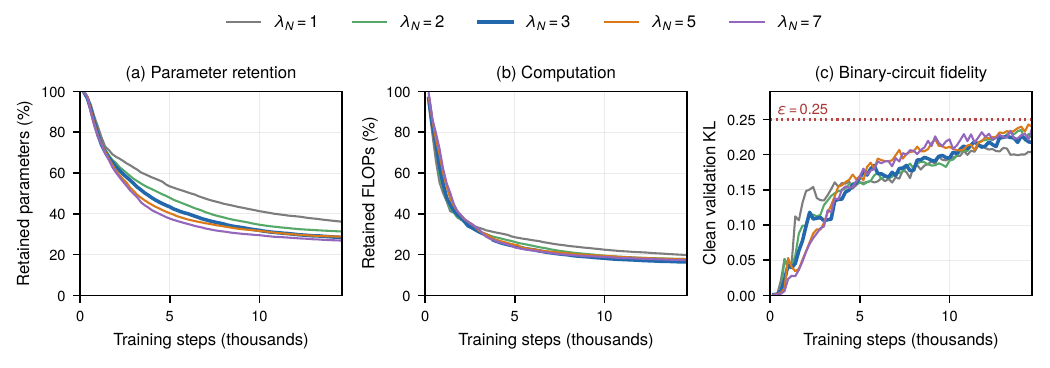}
\caption{\textbf{Neuron-weight diagnostic for the training configuration.} Qwen3-VL-2B, GQA, $\tol=0.25$, seed 0. Each setting uses learning rate 0.02 and a 1,500-step sparsity ramp. {Curves show validation results at each checkpoint.} (a) Larger $\lambda_N$ reaches low parameter retention earlier. (b) Estimated language-model FLOPs do not decrease monotonically with $\lambda_N$. Arithmetic is recomputed from saved masks at fixed mean token counts and $N=16$, excluding kernel-call overhead and width padding. (c) Clean validation KL remains below $\tol$, but larger weights can consume more of this margin. These are validation trajectories, not test-set scores.}
\label{fig:neuron_weight_pilot}
\end{figure*}

The sparsity term evaluates the retained fractions of role-wise heads, role-wise MLPs and shared neurons (\cref{eq:size}). Weighting these fractions equally does not imply equal reductions in parameter count. In pilot experiments, FLOPs stabilised early, while gradual removal of neurons within retained MLPs led to a prolonged decline in parameter retention. We therefore increase the neuron weight $\lambda_N$ to encourage faster neuron removal. As shown in \cref{fig:neuron_weight_pilot}, larger weights reduce the number of updates needed to bring parameter retention below 40\,\%. However, increasing the weight does not consistently improve final FLOPs or fidelity. We adopt $\lambda_N=3$ based on the speed of parameter reduction and the computation and fidelity of the resulting circuits. This choice is based on the evaluated settings, rather than a universally optimal value across models and tasks.

\paragraph{Choosing the step budget.}
\begin{table}[!htb]
\centering
\small
\setlength{\tabcolsep}{5pt}
\caption{\textbf{Node-stage training budgets.} One GQA pilot determines a budget $B$ shared across tasks and tolerances for each model. \Cref{fig:budget_learning} illustrates the selection rule.}
\label{tab:budget}
\begin{tabular}{lrr}
\toprule
Model & Pilot steps & Budget $B$ \\
\midrule
\qwen{2B} & 24{,}000 & 20{,}000 \\
\qwen{4B} & 24{,}000 & 21{,}000 \\
\gemma & 24{,}000 & 24{,}000 \\
\qwen{8B} & 32{,}000 & 25{,}000 \\
\internvl & 32{,}000 & 28{,}000 \\
\bottomrule
\end{tabular}
\end{table}

\begin{figure*}[t]
\centering
\includegraphics[width=0.70\textwidth]{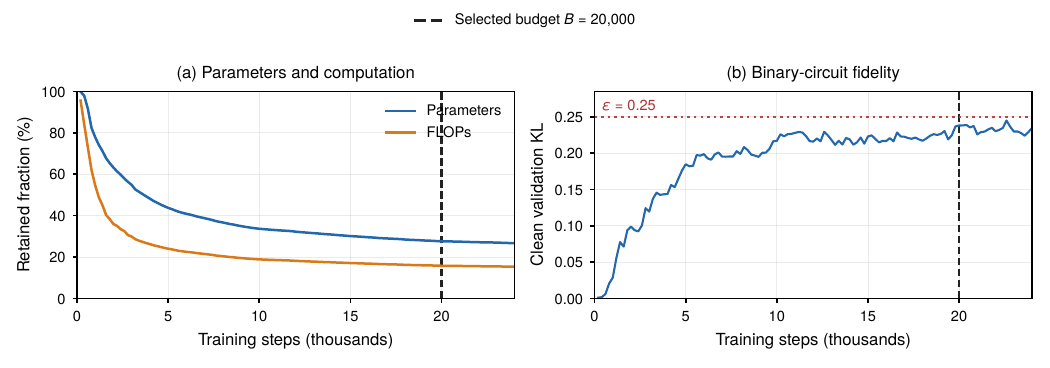}
\caption{\textbf{How the node-stage training budget is selected.} Qwen3-VL-2B, GQA, $\tol=0.25$. {Curves show validation results at each checkpoint of the 24,000-step pilot.} (a) Estimated language-model FLOPs fall faster initially than retained parameters. FLOPs are recomputed from saved masks with fixed mean token counts and 16 decisions per image, counting GEMM/BMM arithmetic without kernel-call overhead or width padding. (b) Validation KL stays below $\tol$ throughout the pilot. The dashed vertical line marks the selected training budget $B=20{,}000$; the selection rule is described in \cref{sec:suppl_solver}.}
\label{fig:budget_learning}
\end{figure*}

For each model, a GQA pilot at $\tol=0.25$ sets a node-stage budget shared across tasks and tolerances (\cref{tab:budget}). We select the first validation check $k$ such that all checks in $[k-2000,k]$ are feasible, parameter retention $P$ changes by less than one percentage point over that window, and the remaining pilot never falls more than one point below $P(k)$. The budget is $B=1000\lceil k/1000\rceil$; if no check qualifies, we use the full pilot length. The training budget is determined once per model from the validation history of the pilot run and shared across tasks and tolerances.
{
In the {Qwen3-VL-2B} pilot shown in \cref{fig:budget_learning}, the selection conditions are first satisfied at 19,800 updates, giving a training budget of 20,000 updates.
During later training, neurons are removed without changing the number of retained role-specific MLP blocks, reducing both parameter count and FLOPs.
\par}

For each run, we select the saved candidate within the training budget that satisfies the validation constraint and minimises the sparsity term $S$. Ties are broken in favour of the earlier candidate (\cref{eq:pick}). For GQA, we select the circuit from the pilot run using the same rule. Test data are reserved for evaluation.

\paragraph{Baselines and adaptations.}
\label{sec:suppl_baselines}
{All comparisons freeze model weights. Mask learning is allowed; recovery training and learned adapters are excluded because they combine extraction with weight adaptation.}
All methods use the same test questions and executor. Replacement rules differ as specified below.
{\haidermean is our adaptation of Haider et al.~\cite{haider2026multigranular}: corrupted-input activations are replaced by fixed role-wise means, enabling standalone execution. Results evaluate this adaptation, not the original patching setup.}
We denote the Haider method's hyperparameter controlling the relative weights of the task loss and sparsity term by $\lambda_H$, and evaluate $\lambda_H\in\{0.8,0.9,0.95\}$. Each model uses the same training budget as our method {in the main comparisons} (\cref{tab:budget}).
GC~\cite{garciacarrasco2025circuits} greedily selects whole heads and MLPs, using fixed means and $\alpha\in\{0.001,0.003,0.01,0.03\}$. On \qwen{4B} GQA, we also evaluate $\alpha=0.002$ and $0.005$.
FLAP~\cite{an2024flap} and Wanda-sp~\cite{an2024flap,sun2024wanda} retain 30--60\,\% of parameters using training-image activation statistics; FLAP uses bias compensation and Wanda-sp zero replacement, without weight updates.
For ShortV~\cite{yuan2025shortv}, we rank layers using 40 training images per task and evaluate different numbers of frozen layers.
Random controls match the retained head and MLP counts of our circuit in each layer and role, with retained positions selected at random.

\begin{table}[t]
\centering
\footnotesize
\setlength{\tabcolsep}{2pt}
\caption{{\textbf{The last circuit of training under different coefficient rules} (\qwen{2B}, \gqa, $\tol=0.25$, 20{,}000 updates; same loss, budget and check interval). Val / test: KL of the last circuit.}}
\label{tab:solver_last}
{\resizebox{\ifdim\width>\columnwidth\columnwidth\else\width\fi}{!}{%
\begin{tabular}{llrrrrr}
\toprule
coefficient & signal & val KL & FLOPs\,\% & par.\,\% & test KL & $\Delta$acc \\
\midrule
fixed, $\lambda=10$ & -- & 0.134 & 30.2 & 43.4 & 0.106 & $-0.3$ \\
fixed, $\lambda=30$ & -- & 0.238 & 17.1 & 28.6 & 0.197 & $-1.4$ \\
fixed, $\lambda=100$ & -- & 0.363 & 9.1 & 14.9 & 0.330 & $-5.4$ \\
dual ascent & training batch & 0.311 & 13.8 & 23.0 & 0.251 & $-3.7$ \\
dual ascent & binary, validation & 0.292 & 14.9 & 25.8 & 0.210 & $-2.5$ \\
barrier & training batch & 0.313 & 11.0 & 18.4 & 0.287 & $-4.3$ \\
barrier (ours) & binary, validation & \textbf{0.238} & 15.8 & 27.6 & 0.215 & $-1.8$ \\
\bottomrule
\end{tabular}}}
\end{table}

\paragraph{Handling the tolerance.}
\label{sec:suppl_tolerance}
{We compare how the sparsity-coefficient rule affects satisfaction of the validation KL constraint at the end of training. On \qwen{2B} \gqa at $\tol=0.25$, we keep the loss, the 20{,}000-update budget and the validation-check interval unchanged, and vary only the coefficient rule (\cref{tab:solver_last}). We evaluate the last circuit, rather than selecting the best circuit satisfying the constraint from each training trajectory.

\smallskip
\noindent\textbf{Fixed coefficient.} A coefficient of 30 yields a circuit within the tolerance (validation KL 0.238, 17.1\,\% retained FLOPs). A coefficient of 10 also satisfies the constraint but retains more computation (validation KL 0.134, 30.2\,\% FLOPs), whereas 100 exceeds the tolerance (validation KL 0.363). Using a fixed coefficient therefore requires tuning it for the target model, task and tolerance.

\smallskip
\noindent\textbf{Dual ascent.} To avoid searching for a fixed coefficient, we tested automatic coefficient adjustment through dual ascent, a widely used approach to constrained optimisation. Dual ascent increases the penalty coefficient after the KL used for the update exceeds the tolerance. The penalty therefore strengthens only after a violation, which can delay recovery. In this comparison, the final validation KL remains above the tolerance of 0.25: 0.292 when the coefficient is updated using the binary circuit's validation KL, and 0.311 when it is updated using training-batch KL. Thus, neither variant ends training with a circuit that satisfies the validation KL constraint.

\smallskip
\noindent\textbf{Barrier on the training-batch KL.} The KL measured with stochastic gates on training batches differs substantially from the binary circuit's validation KL: it is higher after 200 updates (0.60 versus 0.001), but lower after 20{,}000 updates (0.08 versus 0.31). Late in training, the training-batch KL remains below the tolerance while the binary circuit's validation KL exceeds it. The training signal therefore does not trigger sufficient reduction of the sparsity coefficient, and the last circuit violates the validation constraint (validation KL 0.313, test KL 0.287).

\smallskip
\noindent\textbf{Ours.} We set the coefficient using the binary circuit's validation KL. In this setting, all 100 validation checks satisfy the tolerance, including the final circuit (validation KL 0.238, test KL 0.215), without a separate coefficient search.\par}

\subsection{Complete experimental results}
\label{sec:suppl_main}
\begin{table}[t]
\centering
\footnotesize
\setlength{\tabcolsep}{2pt}
\caption{\textbf{Our circuits.}  FLOPs ({language-model matrix products excluding padding}; $N=16$ decisions per image) and used parameters as \% of the dense model, test KL to the full model, agreement of the argmax decisions, and the accuracy difference to {the common full-model reference}
(points).  One run per setting, returned by \cref{eq:pick}.}
\label{tab:main}
\resizebox{\ifdim\width>\columnwidth\columnwidth\else\width\fi}{!}{\begin{tabular}{llrrrrrr}
\toprule
model & task & $\tol$ & FLOPs\,\% & par.\,\% & KL & agree & $\Delta$acc \\
\midrule
Qwen3-VL-2B & GQA & 0.25 & 15.8 & 27.6 & 0.214 & 87.6 & -1.8 \\
 &  & 0.15 & 26.9 & 45.8 & 0.122 & 91.6 & -0.5 \\
 & POPE & 0.25 & 4.8 & 7.8 & 0.121 & 94.2 & -0.5 \\
 &  & 0.15 & 7.3 & 12.0 & 0.091 & 95.6 & -0.7 \\
 & AI2D & 0.25 & 36.2 & 51.1 & 0.222 & 83.2 & -1.0 \\
 &  & 0.15 & 42.8 & 57.8 & 0.149 & 87.6 & -0.5 \\
 & MME & 0.25 & 17.7 & 29.8 & 0.254 & 83.6 & -5.3 \\
 &  & 0.15 & 29.9 & 46.8 & 0.155 & 86.9 & -4.1 \\
\midrule
Qwen3-VL-4B & GQA & 0.25 & 17.8 & 31.9 & 0.232 & 89.6 & -1.1 \\
 &  & 0.15 & 27.4 & 43.5 & 0.138 & 92.5 & +0.0 \\
 & POPE & 0.25 & 5.6 & 9.5 & 0.151 & 93.6 & -0.6 \\
 &  & 0.15 & 10.3 & 16.6 & 0.102 & 95.0 & +0.1 \\
 & AI2D & 0.25 & 34.9 & 49.8 & {0.260} & 86.7 & {-1.9} \\
 &  & 0.15 & 41.7 & 57.8 & 0.150 & 90.8 & -0.8 \\
 & MME & 0.25 & 34.6 & 49.8 & 0.198 & 88.9 & +0.2 \\
 &  & 0.15 & {44.0} & 64.2 & {0.146} & 92.9 & {+0.1} \\
\midrule
Gemma-3-4B & GQA & 0.25 & 26.8 & 38.2 & 0.289 & 88.2 & +0.2 \\
 &  & 0.15 & 39.6 & 58.0 & 0.193 & 91.1 & +0.8 \\
 & POPE & 0.25 & 13.6 & 19.0 & 0.188 & 93.3 & +0.6 \\
 &  & 0.15 & 22.6 & 31.9 & 0.122 & 94.7 & +0.7 \\
 & AI2D & 0.25 & 42.2 & 63.1 & 0.245 & 89.2 & +0.2 \\
 &  & 0.15 & 71.2 & 82.1 & 0.134 & 94.4 & +0.0 \\
 & MME & 0.25 & 30.4 & 44.8 & 0.228 & 89.8 & -1.2 \\
 &  & 0.15 & 55.9 & 77.9 & 0.137 & 93.3 & -0.4 \\
\midrule
Qwen3-VL-8B & GQA & 0.25 & 16.9 & 28.2 & 0.212 & 89.8 & -1.1 \\
 &  & 0.15 & 23.9 & 37.3 & 0.139 & 91.8 & -0.4 \\
 & POPE & 0.25 & 5.1 & 8.1 & 0.132 & 93.7 & +0.5 \\
 &  & 0.15 & 7.4 & 11.2 & 0.092 & 95.1 & +0.8 \\
 & AI2D & 0.25 & 32.8 & 44.9 & 0.257 & 87.3 & -2.0 \\
 &  & 0.15 & 37.7 & 50.5 & 0.160 & 90.5 & -0.3 \\
 & MME & 0.25 & 25.5 & 38.4 & 0.253 & 87.2 & -4.8 \\
 &  & 0.15 & 36.5 & 51.3 & 0.158 & 91.6 & -1.5 \\
\midrule
InternVL3-8B & GQA & 0.25 & 13.9 & 21.2 & 0.222 & 86.9 & -3.5 \\
 &  & 0.15 & 19.8 & 29.1 & 0.146 & 89.9 & -1.8 \\
 & POPE & 0.25 & 4.1 & 5.9 & 0.136 & 92.4 & -0.8 \\
 &  & 0.15 & 6.4 & 9.1 & 0.118 & 93.7 & +0.0 \\
 & AI2D & 0.25 & 33.3 & 46.2 & 0.250 & 84.6 & -4.0 \\
 &  & 0.15 & 38.9 & 51.5 & 0.174 & 88.0 & -2.4 \\
 & MME & 0.25 & 25.1 & 36.8 & 0.229 & 84.1 & -5.0 \\
 &  & 0.15 & 39.7 & 55.9 & 0.105 & 92.8 & -1.0 \\
\bottomrule
\end{tabular}
}
\end{table}

\begin{table*}[t]
\centering
\footnotesize
\setlength{\tabcolsep}{3pt}
\caption{\textbf{Seed variance.}  Mean $\pm$ standard deviation over three extraction seeds of the $\tol=0.25$ circuits of
{\qwen{2B}, }\qwen{4B}, \gemma, \qwen{8B} and \internvl (FLOPs and parameters as \% of the dense model, test agreement, test KL, accuracy with the full
model's {common reference accuracy} in parentheses).  Seed 0 is the run of \cref{tab:main}.}
\label{tab:seeds}
\begin{tabular}{llrrrrr}
\toprule
Model & Task & FLOPs (\%) & Params (\%) & Agree (\%) & KL & Acc (\%) \\
\midrule
Qwen3-VL-2B & GQA & 15.6 $\pm$ 0.3 & 26.9 $\pm$ 0.8 & 87.9 $\pm$ 0.3 & 0.207 $\pm$ 0.006 & 83.6 $\pm$ 0.3 (85.4) \\
 Qwen3-VL-2B & POPE & 4.8 $\pm$ 0.0 & 7.8 $\pm$ 0.1 & 93.9 $\pm$ 0.4 & 0.121 $\pm$ 0.011 & 88.8 $\pm$ 0.3 (89.5) \\
 Qwen3-VL-2B & MME & 18.8 $\pm$ 0.9 & 31.2 $\pm$ 1.2 & 84.1 $\pm$ 0.4 & 0.252 $\pm$ 0.002 & 71.9 $\pm$ 0.3 (76.9) \\
 Qwen3-VL-2B & AI2D & 35.7 $\pm$ 0.5 & 50.8 $\pm$ 0.5 & 83.2 $\pm$ 0.5 & 0.227 $\pm$ 0.009 & 74.3 $\pm$ 0.2 (75.5) \\
\midrule
Qwen3-VL-4B & GQA & 18.6 $\pm$ 0.7 & 33.7 $\pm$ 1.6 & 90.0 $\pm$ 0.4 & 0.228 $\pm$ 0.006 & 86.3 $\pm$ 0.4 (87.1) \\
Qwen3-VL-4B & POPE & 5.4 $\pm$ 0.2 & 9.4 $\pm$ 0.2 & 93.9 $\pm$ 0.5 & 0.147 $\pm$ 0.014 & 89.0 $\pm$ 0.4 (89.4) \\
Qwen3-VL-4B & MME & 34.1 $\pm$ 1.4 & 51.1 $\pm$ 2.1 & 89.1 $\pm$ 0.3 & 0.204 $\pm$ 0.010 & 83.6 $\pm$ 1.1 (84.5) \\
Qwen3-VL-4B & AI2D & 33.8 $\pm$ 1.0 & 47.7 $\pm$ 1.8 & 86.9 $\pm$ 0.6 & 0.258 $\pm$ 0.004 & 80.7 $\pm$ 0.1 ({82.5}) \\
\midrule
Gemma-3-4B & GQA & 29.2 $\pm$ 2.1 & 41.6 $\pm$ 3.0 & 88.7 $\pm$ 0.5 & 0.275 $\pm$ 0.012 & 77.0 $\pm$ 0.5 (76.4) \\
Gemma-3-4B & POPE & 12.7 $\pm$ 1.0 & 17.7 $\pm$ 1.2 & 93.2 $\pm$ 0.4 & 0.177 $\pm$ 0.010 & 86.1 $\pm$ 0.2 (85.7) \\
Gemma-3-4B & MME & 29.9 $\pm$ 1.7 & 43.7 $\pm$ 3.5 & 89.6 $\pm$ 1.1 & 0.239 $\pm$ 0.026 & 77.3 $\pm$ 0.3 (78.3) \\
Gemma-3-4B & AI2D & 44.7 $\pm$ 2.2 & 64.7 $\pm$ 2.0 & 88.4 $\pm$ 1.1 & 0.258 $\pm$ 0.013 & 70.3 $\pm$ 0.6 (70.6) \\
\midrule
Qwen3-VL-8B & GQA & 15.6 $\pm$ 1.2 & 26.2 $\pm$ 1.7 & 89.1 $\pm$ 0.8 & 0.231 $\pm$ 0.019 & 85.8 $\pm$ 0.2 (87.1) \\
Qwen3-VL-8B & POPE & 4.9 $\pm$ 0.3 & 7.7 $\pm$ 0.4 & 93.6 $\pm$ 0.1 & 0.136 $\pm$ 0.003 & 88.9 $\pm$ 0.4 (88.4) \\
Qwen3-VL-8B & MME & 25.4 $\pm$ 0.5 & 37.8 $\pm$ 0.6 & 87.4 $\pm$ 0.9 & 0.258 $\pm$ 0.009 & 82.6 $\pm$ 0.8 (87.3) \\
Qwen3-VL-8B & AI2D & 32.0 $\pm$ 0.7 & 44.2 $\pm$ 0.6 & 87.0 $\pm$ 0.6 & 0.252 $\pm$ 0.007 & 81.7 $\pm$ 0.3 (83.9) \\
\midrule
InternVL3-8B & GQA & 13.2 $\pm$ 0.6 & 20.4 $\pm$ 0.7 & 86.3 $\pm$ 0.6 & 0.234 $\pm$ 0.011 & 83.3 $\pm$ 0.2 (87.0) \\
InternVL3-8B & POPE & 4.1 $\pm$ 0.1 & 6.0 $\pm$ 0.1 & 92.2 $\pm$ 0.6 & 0.144 $\pm$ 0.010 & 88.6 $\pm$ 0.2 (89.5) \\
InternVL3-8B & MME & 26.2 $\pm$ 1.0 & 37.8 $\pm$ 1.2 & 85.7 $\pm$ 1.4 & 0.210 $\pm$ 0.017 & 80.3 $\pm$ 0.5 (85.1) \\
InternVL3-8B & AI2D & 34.4 $\pm$ 0.9 & 47.8 $\pm$ 1.6 & 84.9 $\pm$ 0.3 & 0.249 $\pm$ 0.005 & 79.1 $\pm$ 0.2 (83.0) \\
\bottomrule
\end{tabular}

\end{table*}

\Cref{tab:main} reports all settings for the four main benchmarks, including FLOPs, parameters, test KL, prediction agreement and accuracy change.
FLOPs are evaluated at $N=16$; parameter counts include shared weights only once.
\Cref{tab:seeds} gives the mean and standard deviation over three extraction seeds at $\tol=0.25$ across four tasks for {five models: \qwen{2B}, }\qwen{4B}, \gemma, \qwen{8B} and \internvl.
\Cref{tab:main} reports one run per setting; for settings included in \cref{tab:seeds}, this is seed 0.

\begin{table*}[t]
\centering
\small
\setlength{\tabcolsep}{6pt}
\caption{{\textbf{Reproducibility of circuit selection by task} ($\tol=0.25$, five models, three seeds per task). Each cell reports the mean Jaccard overlap over the three seed pairs for that task. Higher values indicate more consistent selection across seeds (1: identical selections). Parentheses show the mean overlap of random circuits matched to each seed pair in the number of heads per layer and role, or neurons per layer. Role-wise comparisons include I, Q and R; neuron selection is shared across roles. No random baseline is reported for role-MLPs.}}
\label{tab:seed_structure}
\begin{tabular}{llcccc}
\toprule
model & unit & \gqa & \pope & \mme & \aid \\
\midrule
\qwen{2B} & role-heads & 0.80 (0.45) & 0.69 (0.27) & 0.60 (0.42) & 0.84 (0.63) \\
 & role-MLPs & 0.95 & 0.97 & 0.87 & 0.84 \\
 & neurons & 0.38 (0.22) & 0.28 (0.06) & 0.44 (0.34) & 0.72 (0.64) \\
\midrule
\qwen{4B} & role-heads & 0.72 (0.42) & 0.55 (0.20) & 0.63 (0.50) & 0.81 (0.60) \\
 & role-MLPs & 0.94 & 0.79 & 0.83 & 0.95 \\
 & neurons & 0.47 (0.35) & 0.22 (0.10) & 0.74 (0.66) & 0.68 (0.60) \\
\midrule
\gemma & role-heads & 0.82 (0.59) & 0.71 (0.34) & 0.77 (0.61) & 0.83 (0.77) \\
 & role-MLPs & 0.87 & 0.82 & 0.96 & 0.85 \\
 & neurons & 0.62 (0.53) & 0.40 (0.27) & 0.68 (0.61) & 0.72 (0.70) \\
\midrule
\qwen{8B} & role-heads & 0.68 (0.37) & 0.62 (0.18) & 0.59 (0.39) & 0.76 (0.54) \\
 & role-MLPs & 0.92 & 0.73 & 0.80 & 0.89 \\
 & neurons & 0.38 (0.26) & 0.22 (0.08) & 0.55 (0.47) & 0.65 (0.57) \\
\midrule
\internvl & role-heads & 0.60 (0.30) & 0.48 (0.12) & 0.57 (0.38) & 0.72 (0.55) \\
 & role-MLPs & 0.88 & 0.63 & 0.80 & 0.89 \\
 & neurons & 0.30 (0.19) & 0.18 (0.07) & 0.47 (0.40) & 0.61 (0.53) \\
\bottomrule
\end{tabular}

\end{table*}

{\Cref{tab:seed_structure} reports seed-to-seed overlap separately for each task. Across all models and tasks, the mean seed-to-seed overlap of role-heads and shared neurons exceeds that of size-matched random circuits. {Head and MLP-block selections are relatively consistent across seeds, while individual neuron selections vary (mean overlap of 0.18--0.40 across models on \pope).}\par}

{Across all 60 circuits extracted with three seeds, the depth centroid of Q heads is shallower than that of R heads, reproducing the arrangement reported in the main paper.\par}

\paragraph{Why accuracy can drop at small KL.}
\label{sec:suppl_accdrop}
{Accuracy changes by the share of decisions the circuit fixes minus the share it breaks. On MME at $\tol=0.25$, accuracy drops by 5.3, 4.8 and 5.0 percentage points for \qwen{2B}, \qwen{8B} and \internvl, respectively, despite test KL values of 0.254, 0.253 and 0.229 (\cref{tab:main}). The tolerance constrains average KL, so a loss of fidelity on a minority of question types can be obscured by this average. Thus, low average KL does not necessarily imply a small accuracy drop.\par}
Compared with \qwen{8B}, the full \internvl model makes more low-confidence predictions, for which small distribution changes can more easily change the answer; this may contribute to the accuracy drop.

All evaluated settings for our method and the baselines are reported in \cref{tab:frontier_full}.

\begin{figure*}[t]
\centering
\includegraphics[width=\textwidth]{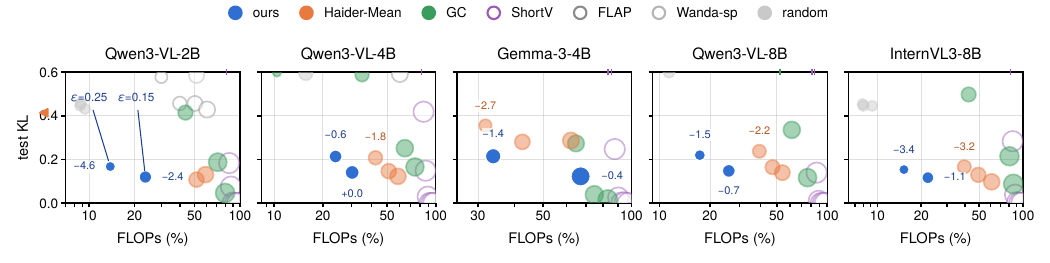}
\caption{{\textbf{Fidelity against FLOPs on \vsr.}  As \cref{fig:frontier}, one panel per model, with the FLOPs axis scaled per model.  Blue: ours at $\tol=0.25$ (left) and $0.15$ (right); numbers: accuracy change (points) of ours and of the smallest \haidermean circuit.  The \qwen{2B} \haidermean circuit at $\lambda_H=0.95$ keeps 0.1\,\% of the FLOPs (KL 0.41) and lies left of its panel.}}
\label{fig:frontier_vsr}
\end{figure*}

\paragraph{A fifth benchmark: VSR.}
\label{sec:suppl_vsr}
We evaluate spatial-relation decisions on \vsr~\cite{liu2023vsr} using the random split (3{,}000 training, 500 validation and 2{,}195 test decisions). Hyperparameters and training budgets are unchanged from the main experiments.

Across five models, $\tol=0.25$ retains 13.8--33.8\,\% of FLOPs at test KL 0.153--0.219 (\cref{fig:frontier_vsr}). Our circuits tend to lie toward the lower left relative to the baselines, achieving high fidelity with less computation on \vsr.

\begin{table}[t]
\centering
\footnotesize
\setlength{\tabcolsep}{2pt}
\caption{{\textbf{Cross-task transfer} (\qwen{8B}, $\tol=0.25$). Rows: the task a circuit is extracted on; columns: the task it is evaluated on (test KL / accuracy change in points). The circuits are used as extracted, their replacement means included. Diagonal entries for the four main tasks reproduce \cref{tab:main}.}}
\label{tab:transfer}
{\resizebox{\ifdim\width>\columnwidth\columnwidth\else\width\fi}{!}{%
\begin{tabular}{lccccc}
\toprule
{\scriptsize (test KL / $\Delta$acc)} & GQA & \pope & \aid & \mme & \vsr \\
\midrule
GQA & \textbf{0.212} / $-1.1$ & 0.141 / $-0.7$ & 1.450 / $-44.4$ & 0.648 / $-22.3$ & 0.479 / $-10.0$ \\
\pope & 4.564 / $-46.1$ & \textbf{0.132} / $+0.5$ & 6.842 / $-57.8$ & 0.924 / $-31.1$ & 1.200 / $-27.9$ \\
\aid & 1.047 / $-10.2$ & 0.075 / $-0.4$ & \textbf{0.257} / $-2.0$ & 0.270 / $-9.1$ & 0.275 / $-7.3$ \\
\mme & 3.305 / $-27.3$ & 0.134 / $+0.1$ & 5.198 / $-51.8$ & \textbf{0.253} / $-4.8$ & 0.483 / $-12.7$ \\
\vsr & 3.952 / $-36.4$ & 0.197 / $-3.5$ & 6.190 / $-57.7$ & 0.612 / $-25.2$ & \textbf{0.219} / $-1.5$ \\
\bottomrule
\end{tabular}}}
\end{table}

\paragraph{Cross-task transfer.}
\label{sec:suppl_transfer}
We apply each \qwen{8B} circuit at $\tol=0.25$ to other tasks using its original replacement means and without further training (\cref{tab:transfer}). On \pope, circuits extracted on other tasks achieve test KL of 0.075--0.197 and change accuracy by $-3.5$ to $+0.1$ points. Transfer varies substantially across targets. For example, the \aid circuit gives test KL of 0.270 on \mme and 0.275 on \vsr, with accuracy drops of 9.1 and 7.3 points, respectively. The \pope circuit loses 57.8 points on \aid. These results show that some circuits transfer across tasks, with performance depending on both the source and target tasks.

\subsection{Method comparisons and ablations}
\label{sec:suppl_ablations}

\paragraph{Role-wise selection.}
\label{sec:suppl_roles}
\begin{figure}[!htb]
\centering
\includegraphics[width=0.80\columnwidth]{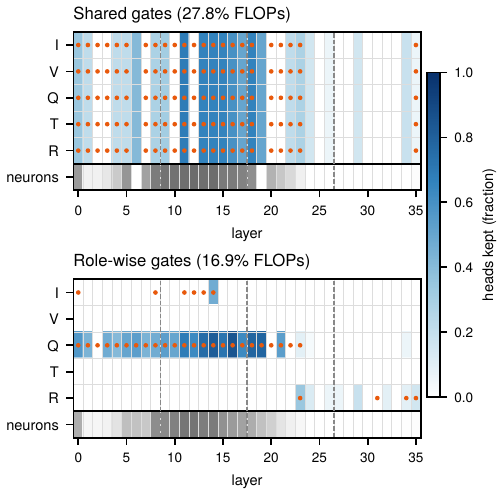}
\caption{\textbf{Role-wise masks reduce FLOPs at similar parameter retention.} Qwen3-VL-8B on GQA, $\tol=0.25$: shared (top) versus role-wise gates (bottom). FLOPs: 27.8\,\% / 16.9\,\%; parameters: 27.3\,\% / 28.2\,\%; test KL: 0.217 / 0.212. Rows are token roles, columns are layers; colour shows the retained head fraction, dots mark retained MLP blocks, and the bottom strip shows shared neuron retention.}
\label{fig:shared_role_map}
\end{figure}

\Cref{fig:shared_role_map} shows how circuit structure differs with and without role-wise gates.
Role-wise gates select execution positions by role, rather than running the same components at every position.
This selection reduces computation even when the retained parameter fractions are similar.

\paragraph{Image interventions versus more training signal.}
\label{sec:suppl_lint_data}
\begin{table}[t]
\centering
\footnotesize
\setlength{\tabcolsep}{3pt}
\renewcommand{\arraystretch}{1.12}
\caption{\textbf{Does more data replace the intervention term?}
\qwen{8B} on \gqa, $\tol=0.25$, 25{,}000 steps.
Indented rows double the training data from 3,000 to 6,000 decisions.
FLOPs and parameters are percentages of the full model.
Swap KL uses whole-image replacement on 200 test image pairs.
Bold denotes the lowest KL in each column.}
\label{tab:lint_control}
\begin{tabular*}{\columnwidth}{@{\extracolsep{\fill}}lrrrr@{}}
\toprule
 & FLOPs & Params & Clean KL & Swap KL \\
Training variant & (\%) & (\%) & $\downarrow$ & $\downarrow$ \\
\midrule
Ours & 16.9 & 28.2 & 0.212 & 0.354 \\
\quad + 2$\times$ data & 16.0 & 27.0 & {\fontencoding{OT1}\fontfamily{cmr}\fontseries{bx}\selectfont 0.210} & {\fontencoding{OT1}\fontfamily{cmr}\fontseries{bx}\selectfont 0.290} \\
\midrule
No $\Dint$ & 16.7 & 30.8 & 0.256 & 0.558 \\
\quad + 2$\times$ data & 16.3 & 29.3 & 0.243 & 0.512 \\
\bottomrule
\end{tabular*}

\end{table}

{
We test whether simply increasing training data can reproduce the improvement from image interventions on {Qwen3-VL-8B} GQA (\cref{tab:lint_control}).
Doubling the training data without $\Dint$ yields {Clean KL} of 0.243 and {Swap KL} of 0.512, both higher than 0.212 and 0.354 with $\Dint$ at the original data size.
Thus, in this setting, doubling the training data alone does not reproduce the fidelity improvement from image interventions.
\par}

\paragraph{Does $\Dint$ change the dependence on image access?}
\label{sec:suppl_dint_access}
\begin{table}[!htbp]
\centering
\footnotesize
\setlength{\tabcolsep}{3pt}
\caption{\textbf{Does the intervention term change the dependence on image access?}
{\qwen{8B}, \gqa. Clean KL before and after cutting Image$\to$Q reads in the read-stage layers of our circuit at the same tolerance.}
$\Delta$ KL = after cut $-$ before cut, computed before rounding, separately for the extracted circuit and with all image-position updates restored.
Full: full model with the same cut. All KL values are relative to the unmodified full model on the same inputs. No retraining.}
\label{tab:dint_factorial}
\begin{tabular*}{\columnwidth}{@{\extracolsep{\fill}}lrrr@{}}
\toprule
Metric & Ours & No $\Dint$ & Full \\
\midrule
\multicolumn{4}{l}{$\tol=0.25$} \\
\midrule
\multicolumn{4}{l}{\emph{Extracted circuit}} \\
Before cut & 0.212 & 0.256 & 0 \\
After cut & 0.826 & 0.725 & 0.973 \\
$\Delta$ KL & $+$0.614 & $+$0.468 & $+$0.973 \\
\multicolumn{4}{l}{\emph{All image updates restored}} \\
Before cut & 0.156 & 0.183 & -- \\
After cut & 0.822 & 0.708 & -- \\
$\Delta$ KL & $+$0.665 & $+$0.524 & -- \\
\midrule
\multicolumn{4}{l}{$\tol=0.15$} \\
\midrule
\multicolumn{4}{l}{\emph{Extracted circuit}} \\
Before cut & 0.139 & 0.166 & 0 \\
After cut & 0.830 & 0.677 & 0.930 \\
$\Delta$ KL & $+$0.690 & $+$0.511 & $+$0.930 \\
\multicolumn{4}{l}{\emph{All image updates restored}} \\
Before cut & 0.106 & 0.107 & -- \\
After cut & 0.815 & 0.664 & -- \\
$\Delta$ KL & $+$0.708 & $+$0.557 & -- \\
\bottomrule
\end{tabular*}

\end{table}

{\Cref{tab:dint_factorial} cuts reads from question positions to image positions in the \qwen{8B} \gqa circuits with and without $\Dint$. Both circuits are cut in the layers where Image$\to$Q reads remain after read selection in our circuit at the same tolerance.}
The cut increases {Clean KL} by 0.61--0.69 with $\Dint$ and 0.47--0.51 without it, compared with 0.93--0.97 for the full model.
{We also restore all image-position updates before cutting the reads to test whether the difference is due to fewer image-position updates. The KL increase remains larger with $\Dint$ after restoration. These results indicate greater sensitivity to the tested Image$\to$Q cuts under this metric. They concern perturbation magnitude relative to the unmodified full model, not agreement between post-intervention output distributions.}

\begin{table}[t]
\centering
\footnotesize
\setlength{\tabcolsep}{3pt}
\caption{{\textbf{Amount of training data} (\qwen{2B}, \gqa, $\tol=0.25$, 20{,}000 updates; validation and test decisions shared by all rows). 3,000 decisions: the main circuits, mean over three seeds; other rows: one seed.}}
\label{tab:ntrain}
{
\begin{tabular}{rrrrrr}
\toprule
train decisions & FLOPs\,\% & par.\,\% & KL & agree & $\Delta$acc \\
\midrule
300 & 21.7 & 38.9 & 0.207 & 87.1 & $-4.0$ \\
1{,}000 & 18.1 & 31.1 & 0.220 & 86.8 & $-1.9$ \\
3{,}000 & 15.6 & 26.9 & 0.207 & 87.9 & $-1.8$ \\
10{,}000 & 13.4 & 22.9 & 0.214 & 87.0 & $-1.8$ \\
\bottomrule
\end{tabular}}
\end{table}

\paragraph{Amount of training data.}
\label{sec:suppl_ntrain}
{On \qwen{2B} \gqa we vary only the number of training decisions (300, 1{,}000, 3{,}000 as in the main experiments, 10{,}000), keeping $\tol=0.25$, the 20{,}000-update budget, the replacement means and the validation (502) and test (4{,}402) decisions fixed (\cref{tab:ntrain}). {Validation data control adaptive weighting and circuit selection, but are not used to compute fidelity gradients.} Extraction uses the full model's outputs, not labels, so these counts are unlabelled image--question pairs.
Test KL (0.207--0.220) and agreement (86.8--87.9\,\%) hardly change, while smaller data give larger circuits (21.7\,\% of the FLOPs at 300 decisions, 13.4\,\% at 10{,}000).
{With 300 training decisions, training KL falls to nearly zero while validation KL approaches the tolerance early, suggesting a tendency to overfit. In response, the sparsity coefficient decreases, yielding a circuit that retains more computation.}
{With 300 training decisions, test KL and prediction agreement remain similar, but accuracy drops by 4.0 points, compared with 1.8 points at 3{,}000 decisions (\cref{tab:ntrain}).}
From 1{,}000 to 10{,}000 decisions, the accuracy drop stays within 1.8--1.9 points while retained FLOPs fall from 18.1 to 13.4\,\%.\par}

\subsection{Executor and speed measurement}
\label{sec:suppl_speed}
\begin{table}[t]
\centering
\footnotesize
\setlength{\tabcolsep}{3pt}
\caption{\textbf{End-to-end speed-up} of our circuits over the reference model with all units enabled at $N=1$, 16 and 64 decisions per
image (\qwen{2B} on one RTX 5070 Ti, \qwen{8B} and \internvl on one NVIDIA H200; the executor compiled and captured in a CUDA Graph against the same
executor with every unit kept; ratio of medians; vision encoding included). }
\label{tab:speed}
\begin{tabular}{llrrrrrr}
\toprule
model & task & \multicolumn{3}{c}{ours $\tol=0.25$} & \multicolumn{3}{c}{ours $\tol=0.15$} \\
\cmidrule(lr){3-5}\cmidrule(lr){6-8}
 & $N$ = & 1 & 16 & 64 & 1 & 16 & 64 \\
\midrule
Qwen3-VL-2B & GQA & 1.65 & 1.96 & 2.77 & 1.55 & 1.77 & 2.24 \\
 & POPE & 1.77 & 2.62 & 4.06 & 1.76 & 2.53 & 3.82 \\
 & AI2D & 1.41 & 1.72 & 2.01 & 1.31 & 1.59 & 1.83 \\
 & MME & 1.45 & 2.05 & 2.61 & 1.36 & 1.81 & 2.11 \\
\midrule
Qwen3-VL-8B & GQA & 1.48 & 2.01 & 2.94 & 1.42 & 1.85 & 2.55 \\
 & POPE & 1.66 & 2.71 & 4.88 & 1.62 & 2.56 & 4.44 \\
 & AI2D & 1.45 & 1.75 & 2.09 & 1.40 & 1.67 & 1.95 \\
 & MME & 1.36 & 1.75 & 2.38 & 1.29 & 1.60 & 2.04 \\
\midrule
InternVL3-8B & GQA & 1.66 & 2.35 & 3.59 & 1.59 & 2.17 & 3.09 \\
 & POPE & 1.78 & 2.56 & 5.10 & 1.74 & 2.47 & 4.65 \\
 & AI2D & 1.40 & 1.89 & 2.31 & 1.34 & 1.78 & 2.13 \\
 & MME & 1.35 & 1.70 & 2.42 & 1.27 & 1.51 & 1.93 \\
\bottomrule
\end{tabular}

\end{table}

\paragraph{Execution and timing protocol.}
\Cref{tab:speed} reports end-to-end speed-ups for $N=1,16,64$ decisions per image. We use bf16 and one process on an H200 ({Qwen3-VL-8B} and {InternVL3-8B}), RTX 5070 Ti ({Qwen3-VL-2B}), or RTX 5080 ({Qwen3-VL-4B} and {Gemma-3-4B}). Each ratio compares methods on the same GPU. The reference model with all units enabled and the extracted circuits use the same executor, \texttt{torch.compile} and CUDA Graph capture; warm-up is excluded. Timings are ratios of medians over 24 test images and 10 repetitions. {On Qwen3-VL-8B GQA at $N=16$, the all-units-active executor's language-model latency differs from Hugging Face by less than 2\,\% under CUDA Graph replay on H200. {Across the full test split of each task, the executor's decisions with the \qwen{8B} circuits ($\tol=0.25$) agree with the hook-based evaluation of the same circuits on 99.50\,\% (\gqa, 4{,}402 decisions), 99.40\,\% (\pope, 9{,}000), 98.65\,\% (\mme, 1{,}186) and 98.64\,\% (\aid, 3{,}088) of the decisions.}}
Image resolution is fixed at $640\times480$ for Qwen (300 image tokens), $448\times448$ for {InternVL3-8B} (256 tokens), and $896\times896$ for {Gemma-3-4B}. Question blocks use test-question lengths. The speed plots use $\lambda_H\in\{0.9,0.95\}$ for \haidermean and $\alpha\in\{0.001,0.003\}$ for GC.

\paragraph{Speed-up bound at $N=1$.}
Let $T_V,T_D$ be the measured vision-encoder and language-model times with all units enabled at $N=1$.
With the vision encoder unchanged, setting the circuit's language-model time to zero gives the speed-up bound $s_{\max}=(T_V+T_D)/T_V$.
At $\tol=0.25$, the ratios of measured end-to-end speed-up to this bound are 74--89\,\% across the four tasks for \qwen{8B} on the H200.
\begin{table}[t]
\centering
\scriptsize
\setlength{\tabcolsep}{3pt}
\caption{{\textbf{Peak GPU memory} (GB) at $N=16$ decisions per image: the dense executor and our circuits at $\tol=0.25$ / $0.15$ (one CUDA-graph replay, \texttt{torch.cuda.max\_memory\_allocated}; same measurement protocol as \cref{tab:speed}).}}
\label{tab:peak_mem}
\resizebox{\ifdim\width>\columnwidth\columnwidth\else\width\fi}{!}{\begin{tabular}{lrrrrr}
\toprule
Model & Dense & \gqa & \pope & \mme & \aid \\
\midrule
\qwen{2B} & 4.4 & 2.6 / 3.1 & 2.1 / 2.2 & 2.7 / 3.2 & 3.3 / 3.6 \\
\qwen{4B} & 8.7 & 4.4 / 5.2 & 2.8 / 3.3 & 5.8 / 7.0 & 5.8 / 6.6 \\
\gemma & 8.5 & 4.9 / 6.2 & 3.7 / 4.5 & 5.3 / 8.2 & 6.5 / 8.5 \\
\qwen{8B} & 16.7 & 7.8 / 9.3 & 5.2 / 5.7 & 9.3 / 11.3 & 10.4 / 11.5 \\
\internvl & 15.9 & 5.8 / 6.9 & 3.9 / 4.4 & 7.7 / 10.3 & 9.0 / 9.9 \\
\bottomrule
\end{tabular}
}
\end{table}

\paragraph{Peak memory.}
{\Cref{tab:peak_mem} reports peak GPU memory at $N=16$. While a circuit is timed, the decoder layers of the original model are offloaded and the executor holds its own copies of the retained weights, so the circuit's peak includes the vision encoder, the embeddings and the LM head.
At $\tol=0.25$, the circuits reduce peak memory by 23--75\,\% (38--69\,\% on \qwen{8B}).
At $\tol=0.15$, the reduction reaches 72\,\%; the largest circuit, \gemma on \aid, uses about as much memory as the dense model, as its weight copies and neuron padding offset the removed units.}

\paragraph{Neuron alignment and accounting.}
\begin{table*}[t]
\centering
\footnotesize
\setlength{\tabcolsep}{4pt}
\caption{\textbf{Effect of neuron padding on inference speed.}
Language-model speed-up at $N=16$ on one NVIDIA H200,
relative to the compiled executor with all units enabled.
The same circuits are measured without and with zero-padding
MLP widths to multiples of 128.
Reported FLOPs exclude padding.
These measurements were collected separately from
\cref{tab:speed,fig:speed}.}
\label{tab:speed_pad}
\begin{tabular}{llrrrrrr}
\toprule
model & task & ours 0.25 & ours 0.15 & \haidermean{} 0.95 & \haidermean{} 0.9 & GC 0.001 & GC 0.003 \\
\midrule
Qwen3-VL-8B & GQA & 2.48 / 4.82 & 1.51 / 3.59 & 1.09 / 2.67 & 0.83 / 2.16 & 1.28 / 1.28 & 1.46 / 1.45 \\
 & POPE & 5.51 / 10.40 & 4.07 / 8.29 & 1.66 / 4.83 & 1.43 / 3.77 & 1.73 / 1.74 & 1.91 / 1.90 \\
 & AI2D & 1.23 / 2.75 & 0.98 / 2.49 & 0.91 / 2.18 & 0.75 / 1.93 & 1.37 / 1.38 & 1.65 / 1.68 \\
 & MME & 1.48 / 3.68 & 1.05 / 2.76 & 1.08 / 2.91 & 0.90 / 2.45 & 1.45 / 1.48 & 1.79 / 1.78 \\
\midrule
InternVL3-8B & GQA & 2.48 / 5.29 & 1.80 / 4.17 & 0.95 / 2.45 & 0.73 / 2.03 & 1.20 / 1.20 & 1.33 / 1.33 \\
 & POPE & 5.10 / 10.30 & 3.66 / 8.59 & 1.48 / 4.55 & 0.94 / 3.02 & 1.31 / 1.31 & 1.40 / 1.40 \\
 & AI2D & 1.03 / 2.79 & 0.92 / 2.45 & 0.79 / 1.85 & 0.68 / 1.64 & 1.17 / 1.17 & 1.33 / 1.33 \\
 & MME & 1.50 / 3.68 & 1.02 / 2.49 & 0.96 / 2.78 & 0.63 / 2.08 & 1.25 / 1.25 & 1.38 / 1.39 \\
\bottomrule
\end{tabular}

\end{table*}

\begin{figure*}[!tp]
\centering
\input{fig/mlp_exec}
\vspace{\floatsep}
\input{fig/attn_exec}
\end{figure*}
\Cref{fig:mlp_exec,fig:attn_exec} illustrate the MLP and attention updates. The executor slices MLPs to retained neurons and zero-pads their widths to multiples of 128 for efficient GPU matrix products. Padding preserves the computed function; reported circuit FLOPs count the unpadded retained operations, whereas timing includes the padded execution. \Cref{tab:speed_pad} compares execution with and without padding: on the H200, padding speeds up ours by 1.9--2.7$\times$ and \haidermean by 2.3--3.3$\times$; GC retains whole MLPs and is unchanged. All main speed results use padding.

\begin{table*}[t]
\centering
\footnotesize
\setlength{\tabcolsep}{4pt}
\caption{\textbf{Executing the read stage.}  The circuits extracted by node selection in \cref{tab:speed} with the read gates of the analysis
stage applied (keys and values that no open read uses are skipped): test KL and speed-up over the compiled executor with all units enabled
on the H200 at $N=16$ decisions per image (language model only; end-to-end with vision encoding at $N$ = 1 / 16 / 64).}
\label{tab:speed_reads}
\begin{tabular}{llrrrrrr}
\toprule
model & task & $\tol$ & \multicolumn{2}{c}{test KL} & \multicolumn{2}{c}{LM speed-up, $N=16$} & e2e speed-up with reads \\
\cmidrule(lr){4-5}\cmidrule(lr){6-7}
 & & & nodes & + reads & nodes & + reads & $N$ = 1 / 16 / 64 \\
\midrule
Qwen3-VL-8B & GQA & 0.25 & 0.212 & 0.218 & 4.76 & 5.52 & 1.50 / 2.09 / 3.13 \\
 & GQA & 0.15 & 0.139 & 0.152 & 3.59 & 4.12 & 1.44 / 1.93 / 2.72 \\
 & POPE & 0.25 & 0.132 & 0.149 & 10.88 & 11.96 & 1.67 / 2.75 / 5.05 \\
 & POPE & 0.15 & 0.092 & 0.103 & 8.24 & 9.54 & 1.63 / 2.64 / 4.63 \\
 & MME & 0.25 & 0.253 & 0.283 & 3.77 & 4.19 & 1.37 / 1.80 / 2.52 \\
 & MME & 0.15 & 0.158 & 0.192 & 2.76 & 3.06 & 1.30 / 1.65 / 2.16 \\
 & AI2D & 0.25 & 0.257 & 0.262 & 2.76 & 3.00 & 1.46 / 1.81 / 2.20 \\
 & AI2D & 0.15 & 0.160 & 0.168 & 2.49 & 2.62 & 1.41 / 1.71 / 2.04 \\
InternVL3-8B & GQA & 0.25 & 0.222 & 0.225 & 5.23 & 5.94 & 1.69 / 2.44 / 3.93 \\
 & GQA & 0.15 & 0.146 & 0.152 & 4.16 & 4.63 & 1.62 / 2.25 / 3.33 \\
 & POPE & 0.25 & 0.136 & 0.127 & 10.12 & 11.30 & 1.80 / 2.61 / 5.34 \\
 & POPE & 0.15 & 0.118 & 0.104 & 8.39 & 9.04 & 1.78 / 2.51 / 4.79 \\
 & MME & 0.25 & 0.229 & 0.249 & 3.67 & 3.93 & 1.36 / 1.73 / 2.55 \\
 & MME & 0.15 & 0.105 & 0.113 & 2.47 & 2.62 & 1.28 / 1.54 / 2.02 \\
 & AI2D & 0.25 & 0.250 & 0.263 & 2.78 & 3.04 & 1.41 / 1.98 / 2.48 \\
 & AI2D & 0.15 & 0.174 & 0.180 & 2.46 & 2.62 & 1.35 / 1.84 / 2.27 \\
\bottomrule
\end{tabular}

\end{table*}

\paragraph{Executing the read stage.}
The read gates used for analysis (\cref{sec:reads}) are learned while keeping the circuit extracted by node selection fixed.
Read selection requires the selected circuit's validation KL to be at most the frozen circuit's validation KL plus one standard error.
The circuit after read selection is also executable: the executor skips the keys
and values of the tokens that no open read uses and the attention columns of unread roles.
{
\Cref{tab:speed_reads} compares inference times before and after read selection for the circuits of \cref{tab:speed}, measured on the same H200 under identical timing conditions.
At $N=16$ and $\tol=0.25$, read selection reduces language-model inference time by 7--14\,\% and increases end-to-end speed-up by 1--5\,\% relative to that before read selection.
}
{The change in test KL relative to that before read selection ranges from $-0.01$ to $+0.03$.} At $\tol=0.15$, language-model time falls by 5--14\,\%.  The FLOPs reported in
\cref{sec:main_results} do not include the additional computation savings from cutting reads. Read selection therefore provides further inference acceleration beyond the reported FLOPs reduction.
For a consistent comparison across methods, the main results use circuits extracted by node selection, before read selection.
\begin{figure*}[!tp]
\centering
\includegraphics[width=\textwidth]{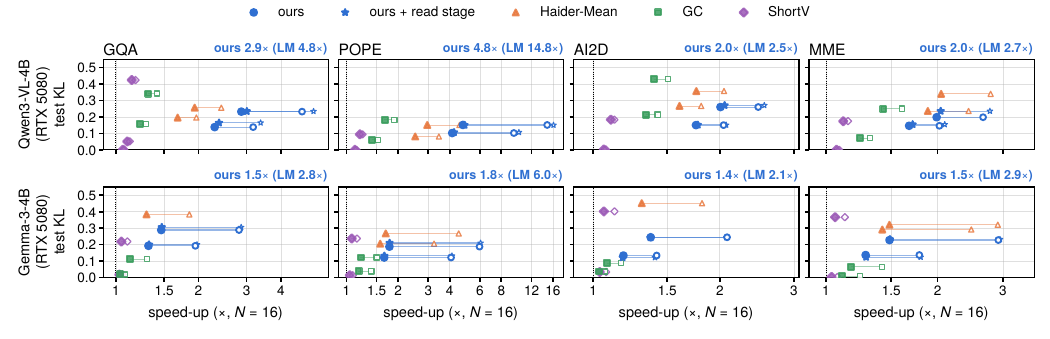}
\caption{\textbf{Fidelity against measured speed on \qwen{4B} and \gemma} (RTX 5080, $N=16$), in the format of \cref{fig:speed}:
test KL against the speed-up over the dense executor; filled: end-to-end, including vision encoding; hollow: language model only.
Circles: ours ($\tol=0.25,0.15$); stars: with read selection; triangles: \haidermean ($\lambda_H=0.95,0.9$); squares: GC; diamonds: ShortV; numbers above each panel as in \cref{fig:speed}.  Horizontal scales differ by task; KL above 0.55
(ShortV with the most frozen layers, and \haidermean on {Gemma-3-4B} GQA and AI2D) lies outside the panels. Same compilation and CUDA-graph
policy as \cref{fig:speed}; {Gemma-3-4B} images are $896\times896$, so its vision encoder dominates the end-to-end time.}
\label{fig:speed_4b_g3}
\end{figure*}

\paragraph{Vision overhead and additional models.}
Vision encoding costs 19--33\,ms on {Qwen3-VL-8B}, {14.3--29.2\,ms} on {InternVL3-8B} and {80--89\,ms} on {Gemma-3-4B}. Because vision-encoder time remains, end-to-end speed-ups are smaller than language-model speed-ups. The image encoding is shared across the $N$ decisions. For example, {Qwen3-VL-8B} POPE at $\tol=0.25$ achieves language-model speed-ups of 7.1/10.9/12.8$\times$ at $N=1/16/64$. At $N=16$, across both $\tol$ settings, {Qwen3-VL-4B} achieves 1.7--4.8$\times$ end-to-end (2.0--14.8$\times$ language-model only), and {Gemma-3-4B} 1.2--1.8$\times$ (1.4--6.0$\times$; \cref{fig:speed_4b_g3}). ShortV, measured with the same H200 protocol, retains 75--92\,\% of the FLOPs and achieves 1.05--1.14$\times$ end-to-end (1.07--1.22$\times$ language-model only): freezing image-position updates still requires computing the keys and values read by later tokens.

\subsection{Circuit structure across models}
\label{sec:suppl_maps}
\begin{figure*}[!tp]
\centering
\def\sidemodel#1#2{\sbox0{#2}\makebox[1.3em]{\raisebox{0.5\ht0}{\rotatebox[origin=c]{90}{\footnotesize\sffamily\bfseries #1}}}\usebox0}
\ifdefined\ltjsetparameter\def\mapw{0.8}\else\def\mapw{0.88}\fi
\sidemodel{\qwen{2B}}{\includegraphics[width=\mapw\textwidth]{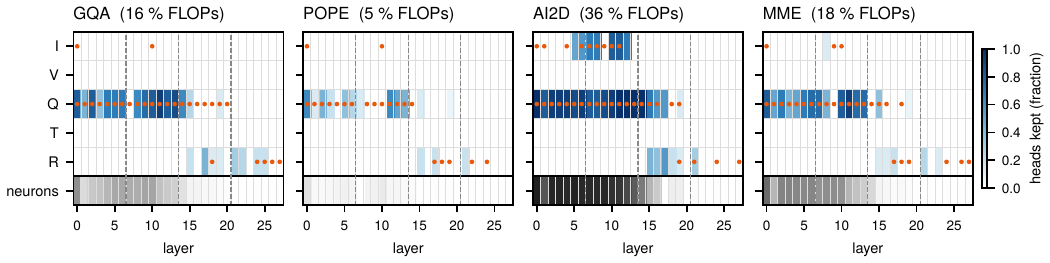}}\\[2pt]
\sidemodel{\qwen{4B}}{\includegraphics[width=\mapw\textwidth]{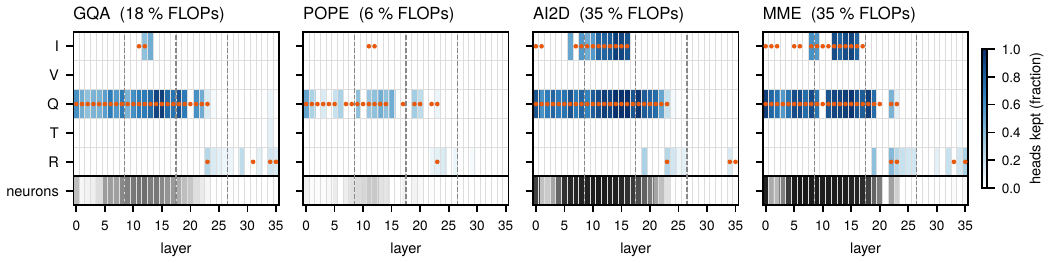}}\\[2pt]
\sidemodel{\gemma}{\includegraphics[width=\mapw\textwidth]{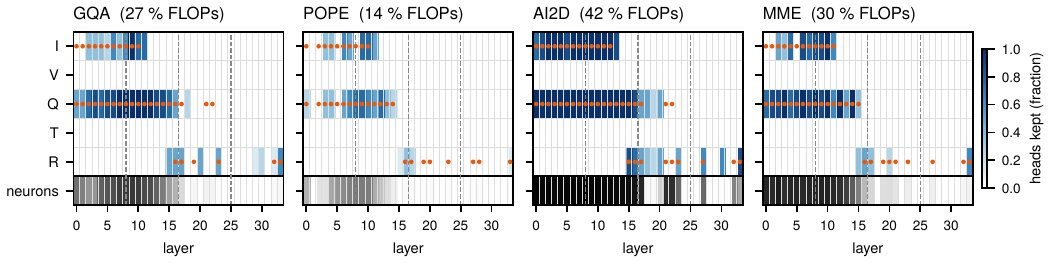}}\\[2pt]
\sidemodel{\qwen{8B}}{\includegraphics[width=\mapw\textwidth]{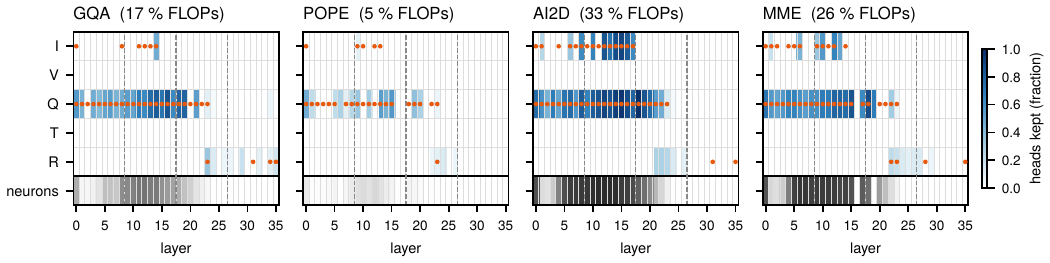}}\\[2pt]
\sidemodel{\internvl}{\includegraphics[width=\mapw\textwidth]{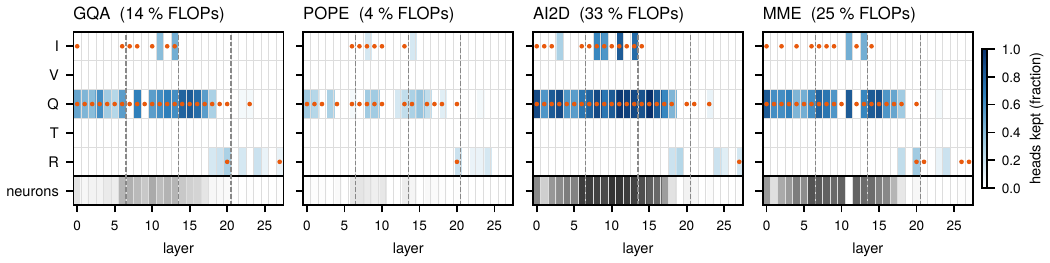}}
\caption{\textbf{Retained updates by role and layer in every model.}  The $\tol=0.25$ circuits of \qwen{2B}, \qwen{4B},
\gemma, \qwen{8B} and \internvl (rows) on \gqa, \pope, \aid and \mme (columns), in the format of \cref{fig:circuit_map}:
rows of a panel are the token roles (I image, V vision-end token, Q question, T instruction, R readout), columns the layers,
colour the fraction of heads kept for the role, orange dot a kept MLP block, bottom strip the kept neurons.  Panel titles give
the FLOPs fraction.}
\label{fig:circuit_map_all}
\end{figure*}

\paragraph{Circuit maps.}
\Cref{fig:circuit_map_all} shows all 20 circuits extracted from five models and four tasks at $\tol=0.25$.
The maps illustrate the patterns described in \cref{sec:analysis}: the role-wise arrangement of Q and R computation and task- and model-dependent computation at image positions.

\begin{figure*}[p]
\centering
\def\sidemodel#1#2{\sbox0{#2}\makebox[1.3em]{\raisebox{0.5\ht0}{\rotatebox[origin=c]{90}{\footnotesize\sffamily\bfseries #1}}}\usebox0}
\sidemodel{\qwen{2B}}{\includegraphics[width=0.955\textwidth]{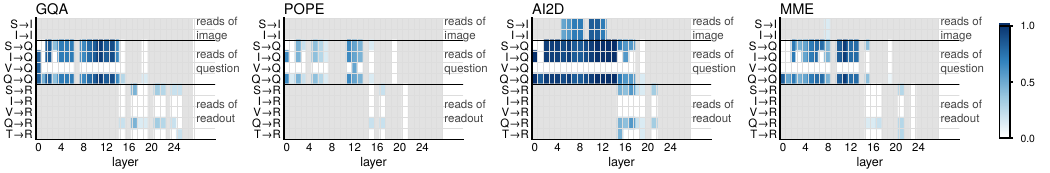}}\\[1pt]
\sidemodel{\qwen{4B}}{\includegraphics[width=0.955\textwidth]{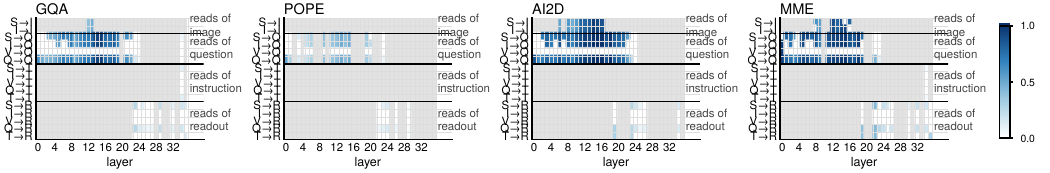}}\\[1pt]
\sidemodel{\gemma}{\includegraphics[width=0.955\textwidth]{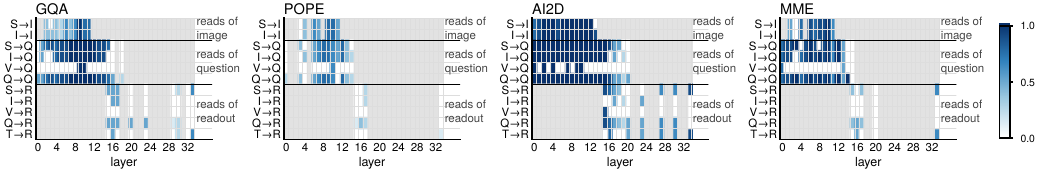}}\\[1pt]
\sidemodel{\qwen{8B}}{\includegraphics[width=0.955\textwidth]{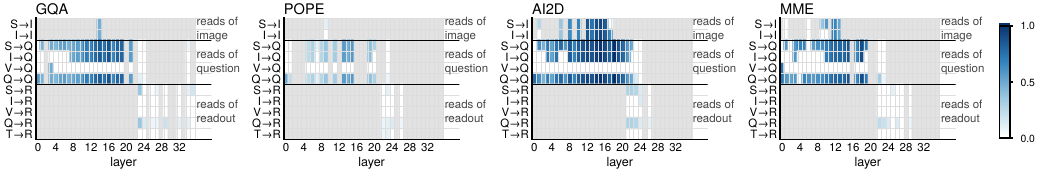}}\\[1pt]
\sidemodel{\internvl}{\includegraphics[width=0.955\textwidth]{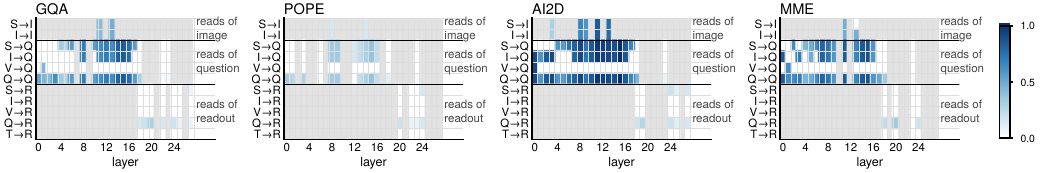}}
\caption{\textbf{Attention reads after edge-mask learning in all models.}  The $\tol=0.25$ circuits of
\qwen{2B} (as in \cref{fig:read_map}), \qwen{4B}, \gemma, \qwen{8B} and \internvl (top to bottom) on \gqa, \pope, \aid and \mme, in the format of
\cref{fig:read_map}: rows are source $\to$ reader role pairs grouped by the reader, columns are layers, colour is the
open gate times the fraction of the layer's heads kept for the reader, grey marks layers where the reader keeps no
head; reader roles without any head are omitted.  Gate counts and KLs are in \cref{tab:read_stage_all}.}
\label{fig:read_map_all}
\end{figure*}

\begin{table*}[t]
\centering
\caption{\textbf{The read stage in all models} ($\tol=0.25$, \cref{fig:read_map_all}).  Gates whose reader keeps a head in the layer
(before) and those left open after learning (after); the validation KL of the frozen circuit, the selection
bound (frozen KL plus one standard error), the validation KL of the selected candidate, the test KL after learning,
and the layers in which the question reads the image and the readout reads the image directly (count in
parentheses).}
\label{tab:read_stage_all}
{\footnotesize\setlength{\tabcolsep}{3pt}\begin{tabular}{llrrrrrll}
\toprule
model & task & open reads (before $\to$ after) & val KL before & val bound & val KL after & test KL after & I$\to$Q open & I$\to$R open \\
\midrule
Qwen3-VL-2B & GQA & 104 $\to$ 54 & 0.239 & 0.265 & 0.253 & 0.231 & 0-14 (12) & none \\
 & POPE & 63 $\to$ 32 & 0.105 & 0.122 & 0.118 & 0.128 & 0-13 (8) & none \\
 & AI2D & 120 $\to$ 78 & 0.247 & 0.260 & 0.259 & 0.232 & 0-16 (15) & none \\
 & MME & 87 $\to$ 44 & 0.249 & 0.285 & 0.276 & 0.268 & 2-13 (9) & none \\
Qwen3-VL-4B & GQA & 154 $\to$ 78 & 0.229 & 0.259 & 0.236 & 0.234 & 5-19 (13) & none \\
 & POPE & 81 $\to$ 42 & 0.148 & 0.175 & 0.154 & 0.150 & 5-20 (10) & none \\
 & AI2D & 150 $\to$ {94} & 0.247 & 0.264 & 0.251 & {0.270} & 3-21 (17) & none \\
 & MME & 153 $\to$ {81} & 0.248 & 0.295 & 0.290 & {0.235} & 0-18 (13) & none \\
Gemma-3-4B & GQA & 137 $\to$ 83 & 0.235 & 0.265 & 0.253 & 0.306 & 1-14 (12) & 16-17 (2) \\
 & POPE & 82 $\to$ 48 & 0.240 & 0.272 & 0.268 & 0.210 & 4-14 (10) & 17-17 (1) \\
 & AI2D & 162 $\to$ 117 & 0.250 & 0.265 & 0.250 & 0.243 & 0-19 (18) & 17-23 (3) \\
 & MME & 107 $\to$ 66 & 0.245 & 0.281 & 0.275 & 0.232 & 0-13 (13) & none \\
Qwen3-VL-8B & GQA & 134 $\to$ 70 & 0.235 & 0.264 & 0.234 & 0.218 & 8-21 (13) & none \\
 & POPE & 78 $\to$ 42 & 0.155 & 0.178 & 0.173 & 0.149 & 5-19 (9) & none \\
 & AI2D & 145 $\to$ 91 & 0.247 & 0.264 & 0.254 & 0.262 & 3-21 (17) & 24-24 (1) \\
 & MME & 131 $\to$ 58 & 0.245 & 0.286 & 0.267 & 0.283 & 3-18 (9) & none \\
InternVL3-8B & GQA & 115 $\to$ 51 & 0.220 & 0.245 & 0.223 & 0.225 & 8-16 (7) & none \\
 & POPE & 76 $\to$ 36 & 0.122 & 0.140 & 0.120 & 0.127 & 8-16 (7) & none \\
 & AI2D & 115 $\to$ 67 & 0.241 & 0.256 & 0.252 & 0.263 & 0-16 (14) & none \\
 & MME & 96 $\to$ 48 & 0.248 & 0.283 & 0.282 & 0.249 & 0-16 (11) & none \\
\bottomrule
\end{tabular}
}
\end{table*}

\paragraph{How is visual information accessed?}
\label{sec:suppl_reads_all}
To examine information access in the 20 circuits, we freeze the extracted node masks and additionally learn role-pair read gates at each layer (\cref{sec:reads}).
We train the gates for 3,000 updates using the same rule as in node selection.
We bound validation KL by that of the circuit before read selection plus one standard error, and select the candidate with the fewest open gates satisfying this bound.
\Cref{fig:read_map_all} shows the read structures obtained through this additional learning.

Consistent with \cref{sec:analysis}, the maps show Image$\to$Q reads in lower and middle layers followed by Q$\to$R reads in deeper layers.
Closing 35--56\,\% of the gates whose reader retains a head (28--41\,\% for {Gemma-3-4B}) changes test KL by at most 0.04 relative to the circuit before read selection.
Thus, the extracted circuits largely maintain fidelity with fewer reads. The learned structure is one sufficient read structure for reproducing decisions.

\begin{table*}[t]
\centering
\scriptsize
\setlength{\tabcolsep}{4pt}
\caption{{\textbf{Read-cut and update-replacement results for all four tasks.} Decisions changed (\%), full model / circuit at $\tol=0.25$ on all test images (95\,\% bootstrap intervals over images are within $\pm$3 points). Layer windows: middle 7--14, late 15--21, final 22--27 (\qwen{2B}); 9--18, 19--27, 28--35 (\qwen{8B}); ``up to late'' starts at layer 0. ``Outside circuit'': layers in which the circuit retains no image-position head or MLP. In the circuit, T's updates are already role means.}}
\label{tab:pathcut_full}
\resizebox{\ifdim\width>\textwidth\textwidth\else\width\fi}{!}{\begin{tabular}{lrrrrrrrr}
\toprule
 & \multicolumn{4}{c}{\qwen{2B}} & \multicolumn{4}{c}{\qwen{8B}} \\
\cmidrule(lr){2-5}\cmidrule(lr){6-9}
 & \gqa & \pope & \mme & \aid & \gqa & \pope & \mme & \aid \\
\midrule
\multicolumn{9}{l}{\emph{attention reads cut} (reader $\leftarrow$ source, layers)} \\
Q$\leftarrow$I, middle & 31 / 31 & 27 / 36 & 25 / 23 & 32 / 31 & 27 / 32 & 26 / 48 & 24 / 30 & 23 / 26 \\
Q$\leftarrow$I, up to late & 38 / 36 & 50 / 45 & 36 / 33 & 36 / 35 & 33 / 35 & 25 / 53 & 33 / 37 & 31 / 31 \\
R$\leftarrow$Q, late & 27 / 33 & 34 / 54 & 32 / 66 & 41 / 71 & 15 / 40 & 7 / 48 & 7 / 41 & 33 / 72 \\
R$\leftarrow$I, all & 1 / 1 & 0 / 0 & 1 / 0 & 3 / 2 & 1 / 0 & 0 / 0 & 1 / 1 & 1 / 1 \\
T$\leftarrow$Q, all & 2 / 0 & 0 / 0 & 2 / 0 & 3 / 0 & 1 / 0 & 1 / 0 & 1 / 0 & 2 / 0 \\
\multicolumn{9}{l}{\emph{updates replaced} (role, layers $\to$ value)} \\
T, all $\to$ mean & 10 / 0 & 7 / 0 & 14 / 0 & 14 / 0 & 9 / 0 & 19 / 0 & 10 / 0 & 13 / 0 \\
I, outside circuit $\to$ mean & 8 / 0 & 6 / 0 & 11 / 0 & 9 / 0 & 6 / 0 & 6 / 0 & 6 / 0 & 6 / 0 \\
T + I outside $\to$ mean & 14 / 0 & 12 / 0 & 18 / 0 & 16 / 0 & 13 / 0 & 25 / 0 & 15 / 0 & 16 / 0 \\
Q, middle $\to$ mean & 62 / 64 & 50 / 54 & 35 / 34 & 73 / 73 & 59 / 63 & 35 / 48 & 40 / 41 & 69 / 71 \\
R, late+final $\to$ mean & 64 / 64 & 45 / 46 & 35 / 34 & 76 / 75 & 63 / 62 & 45 / 48 & 42 / 41 & 71 / 72 \\
\bottomrule
\end{tabular}
}
\end{table*}

\paragraph{Intervention responses across tasks.}
\label{sec:suppl_pathcut}
{\Cref{tab:pathcut_full} extends \cref{tab:pathcut} to all four tasks on \qwen{2B} and \qwen{8B}.
In every setting, replacing T updates, image updates outside the circuit, or their combination with fixed means changes 6--25\,\% of the original model's decisions but none of the circuit's, whereas replacing Q or R updates changes 34--76\,\% in both.
Cutting Q$\to$R reads changes more decisions in the circuits (33--72\,\%) than in the original models (7--41\,\%); the gap is largest on \qwen{8B} \pope and \mme (7\,\% versus 41--48\,\%).}

\subsection{Attribution and patching maps}
\label{sec:suppl_attribution}
\begin{figure*}[!t]
\centering
\input{fig/attribution_maps}
\vspace{\floatsep}
\input{fig/ranking}
\end{figure*}
\Cref{fig:attribution_maps} compares our \qwen{2B} circuits ($\tol=0.25$) with linear AP-IG maps~\cite{hanna2024eapig} (component selection as in~\cite{nikankin2025samecircuits}) and activation patching from a counterfactual image~\cite{wang2023ioi,zhang2024patching}. Both maps highlight the question role in early and middle layers. AP-IG assigns 43--59\,\% of head importance to the readout role, compared with 6--18\,\% for patching. Patching from another image or using role means gives similar rankings (Spearman 0.85--0.93).

\Cref{fig:ranking} applies our gate hierarchy, role means, validation split and FLOPs accounting to both rankings. At validation KL $\le 0.25$, ranked circuits retain 51--76\,\% of FLOPs and 53--91\,\% of parameters. Ours retain 5--36\,\% and 8--51\,\%, respectively, using 1.6--11$\times$ fewer FLOPs than the best ranking per task. {On \gqa, the ranked circuit retains 53.6\,\% of FLOPs and ours retains 15.8\,\%.} {Learning combinations of components helps recover compact circuits that reproduce decisions.}

\clearpage
\onecolumn
{\footnotesize\setlength{\tabcolsep}{4pt}
\begin{longtable}{lllrrrrrrrr}
\caption{\textbf{All points of \cref{fig:frontier}.} Test split, $N=16$ decisions per image. For ShortV, $L_f$ denotes the number of frozen layers. FLOPs and parameters as \% of the dense model; agreement of the argmax decisions; accuracy of the circuit and of the full model and their difference (points). Random controls were run on \qwen{2B}, \qwen{4B} and \gemma, and FLAP and Wanda-sp on every model except \internvl; FLAP is not available for \qwen{4B} \pope. {Full-model accuracy and accuracy differences use a common reference per model and task; KL and agreement use the full-model predictions from each evaluation.}}\label{tab:frontier_full}\\
\toprule
Model & Task & Method & Setting & FLOPs & Params & Test KL & Agree & Acc & Full acc & $\Delta$acc \\
\midrule
\endfirsthead
\toprule
Model & Task & Method & Setting & FLOPs & Params & Test KL & Agree & Acc & Full acc & $\Delta$acc \\
\midrule
\endhead
\bottomrule
\endfoot
Qwen3-VL-2B & GQA & Ours & $\tol=0.25$ & 15.8 & 27.6 & 0.214 & 87.6 & 83.6 & 85.4 & -1.8 \\
 &  & Ours & $\tol=0.15$ & 26.9 & 45.8 & 0.122 & 91.6 & 84.9 & 85.4 & -0.5 \\
 &  & Haider-Mean & $\lambda_H=0.8$ & 58.3 & 58.0 & 0.099 & 91.9 & 85.8 & 85.4 & +0.4 \\
 &  & Haider-Mean & $\lambda_H=0.9$ & 49.9 & 49.4 & 0.150 & 89.5 & 84.0 & 85.4 & -1.4 \\
 &  & Haider-Mean & $\lambda_H=0.95$ & 41.2 & 41.0 & 0.192 & 88.8 & 84.7 & 85.4 & -0.7 \\
 &  & GC & $\alpha=0.001$ & 83.1 & 89.7 & 0.052 & 95.5 & 84.4 & 85.4 & -1.0 \\
 &  & GC & $\alpha=0.003$ & 78.9 & 86.2 & 0.199 & 89.1 & 82.9 & 85.4 & -2.5 \\
 &  & GC & $\alpha=0.01$ & 66.4 & 74.3 & 0.460 & 74.6 & 70.9 & 85.4 & -14.5 \\
 &  & ShortV & $L_f=6$ & 94.8 & 100.0 & 0.000 & 99.8 & 85.4 & 85.4 & +0.0 \\
 &  & ShortV & $L_f=10$ & 90.6 & 100.0 & 0.001 & 99.5 & 85.4 & 85.4 & +0.0 \\
 &  & ShortV & $L_f=14$ & 86.5 & 100.0 & 0.004 & 99.1 & 85.5 & 85.4 & +0.1 \\
 &  & ShortV & $L_f=17$ & 83.3 & 100.0 & 0.056 & 95.3 & 84.6 & 85.4 & -0.8 \\
 &  & ShortV & $L_f=20$ & 80.2 & 100.0 & 0.152 & 91.4 & 83.0 & 85.4 & -2.4 \\
 &  & ShortV & $L_f=24$ & 76.1 & 100.0 & 0.525 & 82.1 & 76.8 & 85.4 & -8.6 \\
 &  & FLAP & keep 0.6 & 60.4 & 61.1 & 1.193 & 35.2 & 36.4 & 85.4 & -49.0 \\
 &  & FLAP & keep 0.5 & 50.1 & 51.2 & 1.273 & 35.3 & 36.5 & 85.4 & -48.9 \\
 &  & FLAP & keep 0.4 & 39.9 & 41.3 & 1.322 & 35.3 & 36.5 & 85.4 & -48.9 \\
 &  & FLAP & keep 0.3 & 30.7 & 31.3 & 1.227 & 35.3 & 36.5 & 85.4 & -48.9 \\
 &  & Wanda-sp & keep 0.6 & 62.6 & 62.5 & 3.314 & 35.2 & 36.5 & 85.4 & -48.9 \\
 &  & Wanda-sp & keep 0.5 & 51.6 & 51.7 & 1.709 & 35.5 & 36.3 & 85.4 & -49.1 \\
 &  & Wanda-sp & keep 0.4 & 42.7 & 42.5 & 1.966 & 37.4 & 37.2 & 85.4 & -48.2 \\
 &  & Wanda-sp & keep 0.3 & 31.8 & 31.8 & 3.255 & 35.5 & 36.1 & 85.4 & -49.3 \\
 &  & Random & seed 1 & 11.6 & 27.9 & 2.684 & 35.3 & 36.5 & 85.4 & -48.9 \\
 &  & Random & seed 0 & 10.8 & 27.9 & 2.394 & 35.3 & 36.5 & 85.4 & -48.9 \\
 &  & Random & seed 2 & 10.8 & 27.9 & 2.570 & 35.3 & 36.5 & 85.4 & -48.9 \\
\midrule
Qwen3-VL-2B & POPE & Ours & $\tol=0.25$ & 4.8 & 7.8 & 0.121 & 94.2 & 89.0 & 89.5 & -0.5 \\
 &  & Ours & $\tol=0.15$ & 7.3 & 12.0 & 0.091 & 95.6 & 88.8 & 89.5 & -0.7 \\
 &  & Haider-Mean & $\lambda_H=0.8$ & 43.5 & 43.5 & 0.050 & 96.7 & 90.5 & 89.5 & +1.0 \\
 &  & Haider-Mean & $\lambda_H=0.9$ & 33.1 & 33.1 & 0.082 & 95.4 & 90.1 & 89.5 & +0.6 \\
 &  & Haider-Mean & $\lambda_H=0.95$ & 22.9 & 23.8 & 0.109 & 94.2 & 88.5 & 89.5 & -1.0 \\
 &  & GC & $\alpha=0.001$ & 69.9 & 77.0 & 0.055 & 96.9 & 89.3 & 89.5 & -0.2 \\
 &  & GC & $\alpha=0.003$ & 64.6 & 69.4 & 0.095 & 95.8 & 88.8 & 89.5 & -0.7 \\
 &  & GC & $\alpha=0.01$ & 56.8 & 61.9 & 0.322 & 89.4 & 84.2 & 89.5 & -5.3 \\
 &  & ShortV & $L_f=6$ & 95.6 & 100.0 & 0.001 & 99.7 & 89.5 & 89.5 & +0.0 \\
 &  & ShortV & $L_f=10$ & 92.0 & 100.0 & 0.001 & 99.7 & 89.5 & 89.5 & +0.0 \\
 &  & ShortV & $L_f=14$ & 88.5 & 100.0 & 0.002 & 99.5 & 89.6 & 89.5 & +0.1 \\
 &  & ShortV & $L_f=17$ & 85.8 & 100.0 & 0.055 & 96.6 & 88.6 & 89.5 & -0.9 \\
 &  & ShortV & $L_f=20$ & 83.1 & 100.0 & 0.283 & 91.0 & 84.3 & 89.5 & -5.2 \\
 &  & ShortV & $L_f=24$ & 79.6 & 100.0 & 1.814 & 72.9 & 67.5 & 89.5 & -22.0 \\
 &  & FLAP & keep 0.6 & 60.5 & 61.0 & 1.124 & 45.2 & 50.0 & 89.5 & -39.5 \\
 &  & FLAP & keep 0.5 & 50.3 & 51.3 & 0.698 & 45.5 & 49.9 & 89.5 & -39.6 \\
 &  & FLAP & keep 0.4 & 39.9 & 41.2 & 0.938 & 45.2 & 50.0 & 89.5 & -39.5 \\
 &  & FLAP & keep 0.3 & 30.4 & 31.4 & 0.874 & 45.3 & 50.0 & 89.5 & -39.5 \\
 &  & Wanda-sp & keep 0.6 & 62.3 & 62.3 & 1.546 & 45.6 & 49.9 & 89.5 & -39.6 \\
 &  & Wanda-sp & keep 0.5 & 51.5 & 51.6 & 1.498 & 45.1 & 49.9 & 89.5 & -39.6 \\
 &  & Wanda-sp & keep 0.4 & 42.7 & 42.6 & 0.912 & 48.8 & 50.6 & 89.5 & -38.9 \\
 &  & Wanda-sp & keep 0.3 & 31.8 & 31.8 & 1.525 & 49.2 & 50.3 & 89.5 & -39.2 \\
 &  & Random & seed 1 & 4.2 & 8.1 & 0.660 & 54.8 & 50.0 & 89.5 & -39.5 \\
 &  & Random & seed 2 & 4.1 & 8.2 & 0.649 & 54.8 & 50.0 & 89.5 & -39.5 \\
 &  & Random & seed 0 & 3.8 & 8.2 & 0.650 & 54.8 & 50.0 & 89.5 & -39.5 \\
\midrule
Qwen3-VL-2B & AI2D & Ours & $\tol=0.25$ & 36.2 & 51.1 & 0.222 & 83.2 & 74.5 & 75.5 & -1.0 \\
 &  & Ours & $\tol=0.15$ & 42.8 & 57.8 & 0.149 & 87.6 & 75.0 & 75.5 & -0.5 \\
 &  & Haider-Mean & $\lambda_H=0.8$ & 64.8 & 64.9 & 0.165 & 86.3 & 75.6 & 75.5 & +0.1 \\
 &  & Haider-Mean & $\lambda_H=0.9$ & 57.9 & 56.7 & 0.227 & 84.3 & 75.0 & 75.5 & -0.5 \\
 &  & Haider-Mean & $\lambda_H=0.95$ & 49.9 & 49.0 & 0.320 & 80.2 & 73.0 & 75.5 & -2.5 \\
 &  & GC & $\alpha=0.001$ & 86.9 & 92.7 & 0.012 & 97.2 & 75.3 & 75.5 & -0.2 \\
 &  & GC & $\alpha=0.003$ & 81.0 & 87.3 & 0.091 & 90.5 & 74.0 & 75.5 & -1.5 \\
 &  & GC & $\alpha=0.01$ & 74.5 & 78.6 & 0.410 & 76.4 & 68.3 & 75.5 & -7.2 \\
 &  & ShortV & $L_f=6$ & 96.3 & 100.0 & 0.004 & 98.2 & 75.6 & 75.5 & +0.1 \\
 &  & ShortV & $L_f=10$ & 93.4 & 100.0 & 0.007 & 97.7 & 75.3 & 75.5 & -0.2 \\
 &  & ShortV & $L_f=14$ & 90.4 & 100.0 & 0.043 & 93.9 & 75.6 & 75.5 & +0.1 \\
 &  & ShortV & $L_f=17$ & 88.2 & 100.0 & 0.143 & 88.1 & 74.4 & 75.5 & -1.1 \\
 &  & ShortV & $L_f=20$ & 86.0 & 100.0 & 0.284 & 81.3 & 70.6 & 75.5 & -4.9 \\
 &  & ShortV & $L_f=24$ & 83.1 & 100.0 & 0.534 & 73.3 & 65.5 & 75.5 & -10.0 \\
 &  & FLAP & keep 0.6 & 60.2 & 60.9 & 1.205 & 28.0 & 26.3 & 75.5 & -49.2 \\
 &  & FLAP & keep 0.5 & 50.0 & 51.1 & 1.174 & 25.7 & 25.6 & 75.5 & -49.9 \\
 &  & FLAP & keep 0.4 & 39.8 & 41.1 & 1.199 & 23.0 & 24.6 & 75.5 & -50.9 \\
 &  & FLAP & keep 0.3 & 30.7 & 31.2 & 1.198 & 23.5 & 25.0 & 75.5 & -50.5 \\
 &  & Wanda-sp & keep 0.6 & 62.1 & 62.0 & 3.685 & 24.6 & 24.9 & 75.5 & -50.6 \\
 &  & Wanda-sp & keep 0.5 & 51.7 & 51.7 & 2.044 & 25.9 & 25.1 & 75.5 & -50.4 \\
 &  & Wanda-sp & keep 0.4 & 42.5 & 42.4 & 2.471 & 24.8 & 23.8 & 75.5 & -51.7 \\
 &  & Wanda-sp & keep 0.3 & 31.5 & 31.5 & 1.440 & 26.4 & 25.5 & 75.5 & -50.0 \\
 &  & Random & seed 1 & 24.2 & 51.6 & 1.283 & 24.3 & 24.5 & 75.5 & -51.0 \\
 &  & Random & seed 0 & 24.0 & 51.9 & 1.217 & 24.3 & 24.5 & 75.5 & -51.0 \\
 &  & Random & seed 2 & 20.9 & 51.8 & 1.244 & 24.3 & 24.5 & 75.5 & -51.0 \\
\midrule
Qwen3-VL-2B & MME & Ours & $\tol=0.25$ & 17.7 & 29.8 & 0.254 & 83.6 & 71.6 & 76.9 & -5.3 \\
 &  & Ours & $\tol=0.15$ & 29.9 & 46.8 & 0.155 & 86.9 & 72.8 & 76.9 & -4.1 \\
 &  & Haider-Mean & $\lambda_H=0.8$ & 56.5 & 56.0 & 0.121 & 89.6 & 76.1 & 76.9 & -0.8 \\
 &  & Haider-Mean & $\lambda_H=0.9$ & 47.1 & 46.9 & 0.167 & 88.7 & 75.2 & 76.9 & -1.7 \\
 &  & Haider-Mean & $\lambda_H=0.95$ & 39.4 & 39.3 & 0.242 & 84.5 & 72.0 & 76.9 & -4.9 \\
 &  & GC & $\alpha=0.001$ & 74.9 & 84.0 & 0.043 & 95.5 & 76.1 & 76.9 & -0.8 \\
 &  & GC & $\alpha=0.003$ & 70.3 & 77.6 & 0.186 & 86.8 & 73.2 & 76.9 & -3.7 \\
 &  & GC & $\alpha=0.01$ & 61.1 & 66.8 & 0.450 & 70.3 & 54.1 & 76.9 & -22.8 \\
 &  & ShortV & $L_f=6$ & 95.0 & 100.0 & 0.002 & 99.2 & 77.2 & 76.9 & +0.3 \\
 &  & ShortV & $L_f=10$ & 91.0 & 100.0 & 0.002 & 99.3 & 77.2 & 76.9 & +0.3 \\
 &  & ShortV & $L_f=14$ & 87.0 & 100.0 & 0.004 & 99.0 & 77.1 & 76.9 & +0.2 \\
 &  & ShortV & $L_f=17$ & 83.9 & 100.0 & 0.058 & 95.3 & 75.2 & 76.9 & -1.7 \\
 &  & ShortV & $L_f=20$ & 80.9 & 100.0 & 0.235 & 88.3 & 68.9 & 76.9 & -8.0 \\
 &  & ShortV & $L_f=24$ & 76.9 & 100.0 & 0.672 & 79.1 & 62.7 & 76.9 & -14.2 \\
 &  & FLAP & keep 0.6 & 60.5 & 61.0 & 0.503 & 65.4 & 50.1 & 76.9 & -26.8 \\
 &  & FLAP & keep 0.5 & 50.2 & 51.3 & 0.536 & 65.6 & 49.9 & 76.9 & -27.0 \\
 &  & FLAP & keep 0.4 & 39.9 & 41.4 & 0.655 & 35.5 & 50.0 & 76.9 & -26.9 \\
 &  & FLAP & keep 0.3 & 30.8 & 31.1 & 0.744 & 34.7 & 50.0 & 76.9 & -26.9 \\
 &  & Wanda-sp & keep 0.6 & 62.3 & 62.3 & 2.148 & 34.4 & 49.9 & 76.9 & -27.0 \\
 &  & Wanda-sp & keep 0.5 & 51.7 & 51.7 & 1.308 & 45.1 & 51.0 & 76.9 & -25.9 \\
 &  & Wanda-sp & keep 0.4 & 42.7 & 42.6 & 0.894 & 47.4 & 50.3 & 76.9 & -26.6 \\
 &  & Wanda-sp & keep 0.3 & 31.6 & 31.5 & 0.548 & 59.9 & 51.4 & 76.9 & -25.5 \\
 &  & Random & seed 1 & 11.7 & 30.2 & 0.686 & 65.5 & 50.0 & 76.9 & -26.9 \\
 &  & Random & seed 0 & 11.3 & 30.2 & 0.778 & 65.5 & 50.0 & 76.9 & -26.9 \\
 &  & Random & seed 2 & 10.7 & 30.0 & 0.782 & 65.5 & 50.0 & 76.9 & -26.9 \\
\midrule
Qwen3-VL-4B & GQA & Ours & $\tol=0.25$ & 17.8 & 31.9 & 0.232 & 89.6 & 86.0 & 87.1 & -1.1 \\
 &  & Ours & $\tol=0.15$ & 27.4 & 43.5 & 0.138 & 92.5 & 87.1 & 87.1 & +0.0 \\
 &  & Haider-Mean & $\lambda_H=0.8$ & 55.7 & 55.5 & 0.128 & 93.4 & 86.7 & 87.1 & -0.4 \\
 &  & Haider-Mean & $\lambda_H=0.9$ & 48.7 & 48.7 & 0.196 & 91.0 & 86.2 & 87.1 & -0.9 \\
 &  & Haider-Mean & $\lambda_H=0.95$ & 40.0 & 40.2 & 0.256 & 88.4 & 84.8 & 87.1 & -2.3 \\
 &  & GC & $\alpha=0.001$ & 81.5 & 83.2 & 0.155 & 93.6 & 85.9 & 87.1 & -1.2 \\
 &  & GC & $\alpha=0.002$ & 81.4 & 83.5 & 0.269 & 89.9 & 84.4 & 87.1 & -2.7 \\
 &  & GC & $\alpha=0.003$ & 75.9 & 78.1 & 0.342 & 88.1 & 82.5 & 87.1 & -4.6 \\
 &  & GC & $\alpha=0.005$ & 73.5 & 75.7 & 0.526 & 84.6 & 79.8 & 87.1 & -7.3 \\
 &  & GC & $\alpha=0.01$ & 59.9 & 64.8 & 0.881 & 62.6 & 62.4 & 87.1 & -24.7 \\
 &  & ShortV & $L_f=6$ & 95.8 & 100.0 & 0.000 & 99.9 & 87.2 & 87.1 & +0.1 \\
 &  & ShortV & $L_f=10$ & 92.5 & 100.0 & 0.000 & 99.9 & 87.2 & 87.1 & +0.1 \\
 &  & ShortV & $L_f=14$ & 89.2 & 100.0 & 0.001 & 99.6 & 87.2 & 87.1 & +0.1 \\
 &  & ShortV & $L_f=18$ & 85.9 & 100.0 & 0.011 & 98.4 & 86.7 & 87.1 & -0.4 \\
 &  & ShortV & $L_f=22$ & 82.6 & 100.0 & 0.051 & 96.6 & 86.8 & 87.1 & -0.3 \\
 &  & ShortV & $L_f=26$ & 79.2 & 100.0 & 0.310 & 90.8 & 83.6 & 87.1 & -3.5 \\
 &  & ShortV & $L_f=30$ & 75.9 & 100.0 & 0.424 & 87.9 & 81.6 & 87.1 & -5.5 \\
 &  & FLAP & keep 0.6 & 60.3 & 60.6 & 0.417 & 80.1 & 77.5 & 87.1 & -9.6 \\
 &  & FLAP & keep 0.5 & 50.4 & 51.1 & 1.632 & 46.1 & 46.3 & 87.1 & -40.8 \\
 &  & FLAP & keep 0.4 & 40.5 & 41.7 & 4.690 & 36.3 & 36.5 & 87.1 & -50.6 \\
 &  & FLAP & keep 0.3 & 31.9 & 31.7 & 4.610 & 36.3 & 36.5 & 87.1 & -50.6 \\
 &  & Wanda-sp & keep 0.6 & 62.2 & 62.2 & 1.795 & 36.7 & 36.8 & 87.1 & -50.3 \\
 &  & Wanda-sp & keep 0.5 & 51.9 & 51.9 & 1.786 & 37.0 & 36.9 & 87.1 & -50.2 \\
 &  & Wanda-sp & keep 0.4 & 42.1 & 42.1 & 2.258 & 36.4 & 36.7 & 87.1 & -50.4 \\
 &  & Wanda-sp & keep 0.3 & 32.3 & 32.3 & 1.625 & 36.2 & 36.0 & 87.1 & -51.1 \\
 &  & Random & seed 2 & 13.1 & 32.5 & 3.742 & 36.3 & 36.5 & 87.1 & -50.6 \\
 &  & Random & seed 0 & 13.0 & 32.4 & 3.381 & 36.3 & 36.5 & 87.1 & -50.6 \\
 &  & Random & seed 1 & 12.9 & 32.6 & 4.048 & 36.3 & 36.5 & 87.1 & -50.6 \\
\midrule
Qwen3-VL-4B & POPE & Ours & $\tol=0.25$ & 5.6 & 9.5 & 0.151 & 93.6 & 88.8 & 89.4 & -0.6 \\
 &  & Ours & $\tol=0.15$ & 10.3 & 16.6 & 0.102 & 95.0 & 89.5 & 89.4 & +0.1 \\
 &  & Haider-Mean & $\lambda_H=0.8$ & 37.2 & 38.1 & 0.075 & 96.0 & 89.4 & 89.4 & +0.0 \\
 &  & Haider-Mean & $\lambda_H=0.9$ & 30.6 & 30.9 & 0.081 & 95.8 & 89.6 & 89.4 & +0.2 \\
 &  & Haider-Mean & $\lambda_H=0.95$ & 23.5 & 23.7 & 0.151 & 94.0 & 90.0 & 89.4 & +0.6 \\
 &  & GC & $\alpha=0.001$ & 67.6 & 73.4 & 0.061 & 97.0 & 89.1 & 89.4 & -0.3 \\
 &  & GC & $\alpha=0.003$ & 54.5 & 62.9 & 0.183 & 92.3 & 86.2 & 89.4 & -3.2 \\
 &  & GC & $\alpha=0.01$ & 45.1 & 47.7 & 0.629 & 58.0 & 61.7 & 89.4 & -27.7 \\
 &  & ShortV & $L_f=6$ & 96.5 & 100.0 & 0.000 & 100.0 & 89.4 & 89.4 & +0.0 \\
 &  & ShortV & $L_f=10$ & 93.6 & 100.0 & 0.000 & 99.9 & 89.5 & 89.4 & +0.1 \\
 &  & ShortV & $L_f=14$ & 90.8 & 100.0 & 0.000 & 99.7 & 89.5 & 89.4 & +0.1 \\
 &  & ShortV & $L_f=18$ & 88.0 & 100.0 & 0.010 & 99.0 & 89.3 & 89.4 & -0.1 \\
 &  & ShortV & $L_f=22$ & 85.1 & 100.0 & 0.094 & 96.7 & 87.9 & 89.4 & -1.5 \\
 &  & ShortV & $L_f=26$ & 82.3 & 100.0 & 0.529 & 89.9 & 82.9 & 89.4 & -6.5 \\
 &  & ShortV & $L_f=30$ & 79.5 & 100.0 & 1.415 & 80.5 & 73.9 & 89.4 & -15.5 \\
 &  & Wanda-sp & keep 0.6 & 62.2 & 62.2 & 0.955 & 44.3 & 49.9 & 89.4 & -39.5 \\
 &  & Wanda-sp & keep 0.5 & 51.9 & 51.9 & 1.897 & 44.2 & 50.0 & 89.4 & -39.4 \\
 &  & Wanda-sp & keep 0.4 & 42.1 & 42.2 & 2.281 & 44.2 & 50.0 & 89.4 & -39.4 \\
 &  & Wanda-sp & keep 0.3 & 32.3 & 32.3 & 1.739 & 44.5 & 50.3 & 89.4 & -39.1 \\
 &  & Random & seed 2 & 4.3 & 9.7 & 0.984 & 55.8 & 50.0 & 89.4 & -39.4 \\
 &  & Random & seed 1 & 4.2 & 9.7 & 0.982 & 55.8 & 50.0 & 89.4 & -39.4 \\
 &  & Random & seed 0 & 4.0 & 9.6 & 0.907 & 55.8 & 50.0 & 89.4 & -39.4 \\
\midrule
Qwen3-VL-4B & AI2D & Ours & $\tol=0.25$ & 34.9 & 49.8 & {0.260} & 86.7 & 80.6 & {82.5} & {-1.9} \\
 &  & Ours & $\tol=0.15$ & 41.7 & 57.8 & 0.150 & 90.8 & 81.7 & 82.5 & -0.8 \\
 &  & Haider-Mean & $\lambda_H=0.8$ & 61.5 & 61.2 & 0.194 & 89.5 & 81.6 & 82.5 & -0.9 \\
 &  & Haider-Mean & $\lambda_H=0.9$ & 54.1 & 54.2 & 0.267 & 86.6 & 80.4 & 82.5 & -2.1 \\
 &  & Haider-Mean & $\lambda_H=0.95$ & 47.7 & 47.4 & 0.357 & 83.2 & 79.0 & 82.5 & -3.5 \\
 &  & GC & $\alpha=0.001$ & 72.7 & 83.9 & 0.214 & 89.8 & 80.3 & 82.5 & -2.2 \\
 &  & GC & $\alpha=0.003$ & 70.3 & 75.6 & 0.431 & 82.8 & 76.4 & 82.5 & -6.1 \\
 &  & GC & $\alpha=0.01$ & 63.7 & 71.7 & 0.969 & 66.3 & 63.4 & 82.5 & -19.1 \\
 &  & GC & $\alpha=0.03$ & 38.0 & 40.4 & 1.779 & 24.9 & 24.5 & 82.5 & -58.0 \\
 &  & ShortV & $L_f=6$ & 97.1 & 100.0 & 0.001 & 99.7 & 82.5 & 82.5 & +0.0 \\
 &  & ShortV & $L_f=10$ & 94.7 & 100.0 & 0.001 & 99.7 & 82.5 & 82.5 & +0.0 \\
 &  & ShortV & $L_f=14$ & 92.4 & 100.0 & 0.002 & 99.4 & 82.7 & 82.5 & +0.2 \\
 &  & ShortV & $L_f=18$ & 90.0 & 100.0 & 0.034 & 96.4 & 82.0 & 82.5 & -0.5 \\
 &  & ShortV & $L_f=22$ & 87.7 & 100.0 & 0.183 & 92.0 & 80.4 & 82.5 & -2.1 \\
 &  & ShortV & $L_f=26$ & 85.3 & 100.0 & 0.537 & 82.0 & 74.7 & 82.5 & -7.8 \\
 &  & ShortV & $L_f=30$ & 83.0 & 100.0 & 0.814 & 74.2 & 69.0 & 82.5 & -13.5 \\
 &  & FLAP & keep 0.6 & 60.3 & 61.0 & 0.948 & 56.4 & 53.5 & 82.5 & -29.0 \\
 &  & FLAP & keep 0.5 & 50.4 & 51.3 & 3.209 & 25.5 & 24.9 & 82.5 & -57.6 \\
 &  & FLAP & keep 0.4 & 40.2 & 41.4 & 4.325 & 24.9 & 24.5 & 82.5 & -58.0 \\
 &  & FLAP & keep 0.3 & 31.8 & 31.6 & 4.918 & 24.8 & 24.5 & 82.5 & -58.0 \\
 &  & Wanda-sp & keep 0.6 & 62.3 & 62.2 & 1.347 & 37.3 & 36.9 & 82.5 & -45.6 \\
 &  & Wanda-sp & keep 0.5 & 51.9 & 51.9 & 2.930 & 24.7 & 24.5 & 82.5 & -58.0 \\
 &  & Wanda-sp & keep 0.4 & 42.2 & 42.2 & 2.846 & 25.4 & 25.2 & 82.5 & -57.3 \\
 &  & Wanda-sp & keep 0.3 & 32.3 & 32.3 & 1.999 & 25.2 & 25.7 & 82.5 & -56.8 \\
 &  & Random & seed 0 & 24.1 & 50.1 & 1.480 & 24.9 & 24.5 & 82.5 & -58.0 \\
 &  & Random & seed 2 & 23.5 & 50.3 & 1.496 & 24.9 & 24.5 & 82.5 & -58.0 \\
 &  & Random & seed 1 & 23.0 & 50.1 & 1.640 & 24.9 & 24.5 & 82.5 & -58.0 \\
\midrule
Qwen3-VL-4B & MME & Ours & $\tol=0.25$ & 34.6 & 49.8 & 0.198 & 88.9 & 84.7 & 84.5 & +0.2 \\
 &  & Ours & $\tol=0.15$ & {44.0} & 64.2 & {0.146} & 92.9 & 84.6 & {84.5} & {+0.1} \\
 &  & Haider-Mean & $\lambda_H=0.8$ & 53.2 & 53.2 & 0.184 & 90.6 & 84.9 & 84.5 & +0.4 \\
 &  & Haider-Mean & $\lambda_H=0.9$ & 43.2 & 43.4 & 0.236 & 87.5 & 82.0 & 84.5 & -2.5 \\
 &  & Haider-Mean & $\lambda_H=0.95$ & 38.1 & 38.3 & 0.340 & 85.8 & 83.1 & 84.5 & -1.4 \\
 &  & GC & $\alpha=0.001$ & 77.9 & 79.0 & 0.071 & 96.3 & 84.1 & 84.5 & -0.4 \\
 &  & GC & $\alpha=0.003$ & 65.4 & 69.4 & 0.250 & 88.7 & 78.8 & 84.5 & -5.7 \\
 &  & GC & $\alpha=0.01$ & 49.4 & 54.4 & 0.655 & 60.0 & 50.1 & 84.5 & -34.4 \\
 &  & ShortV & $L_f=6$ & 96.0 & 100.0 & 0.000 & 99.8 & 84.3 & 84.5 & -0.2 \\
 &  & ShortV & $L_f=10$ & 92.8 & 100.0 & 0.000 & 99.7 & 84.4 & 84.5 & -0.1 \\
 &  & ShortV & $L_f=14$ & 89.6 & 100.0 & 0.000 & 99.8 & 84.3 & 84.5 & -0.2 \\
 &  & ShortV & $L_f=18$ & 86.4 & 100.0 & 0.013 & 98.8 & 84.5 & 84.5 & +0.0 \\
 &  & ShortV & $L_f=22$ & 83.2 & 100.0 & 0.174 & 94.4 & 83.6 & 84.5 & -0.9 \\
 &  & ShortV & $L_f=26$ & 80.0 & 100.0 & 0.311 & 90.1 & 80.3 & 84.5 & -4.2 \\
 &  & ShortV & $L_f=30$ & 76.8 & 100.0 & 0.835 & 84.0 & 72.8 & 84.5 & -11.7 \\
 &  & FLAP & keep 0.6 & 60.2 & 60.6 & 0.617 & 67.0 & 71.1 & 84.5 & -13.4 \\
 &  & FLAP & keep 0.5 & 50.4 & 51.3 & 3.123 & 40.2 & 50.0 & 84.5 & -34.5 \\
 &  & FLAP & keep 0.4 & 40.5 & 41.8 & 2.397 & 40.2 & 50.0 & 84.5 & -34.5 \\
 &  & FLAP & keep 0.3 & 31.8 & 31.7 & 2.088 & 40.2 & 50.0 & 84.5 & -34.5 \\
 &  & Wanda-sp & keep 0.6 & 62.3 & 62.2 & 1.492 & 40.2 & 50.0 & 84.5 & -34.5 \\
 &  & Wanda-sp & keep 0.5 & 51.8 & 51.9 & 2.399 & 41.2 & 49.5 & 84.5 & -35.0 \\
 &  & Wanda-sp & keep 0.4 & 42.1 & 42.2 & 1.532 & 42.7 & 51.0 & 84.5 & -33.5 \\
 &  & Wanda-sp & keep 0.3 & 32.3 & 32.3 & 2.592 & 40.1 & 49.9 & 84.5 & -34.6 \\
 &  & Random & seed 2 & 23.7 & 50.4 & 1.591 & 59.8 & 50.0 & 84.5 & -34.5 \\
 &  & Random & seed 0 & 23.5 & 50.3 & 1.468 & 59.8 & 50.0 & 84.5 & -34.5 \\
 &  & Random & seed 1 & 22.8 & 50.4 & 0.856 & 59.8 & 50.0 & 84.5 & -34.5 \\
\midrule
Gemma-3-4B & GQA & Ours & $\tol=0.25$ & 26.8 & 38.2 & 0.289 & 88.2 & 76.6 & 76.4 & +0.2 \\
 &  & Ours & $\tol=0.15$ & 39.6 & 58.0 & 0.193 & 91.1 & 77.2 & 76.4 & +0.8 \\
 &  & Haider-Mean & $\lambda_H=0.8$ & 58.2 & 57.9 & 0.307 & 88.3 & 76.7 & 76.4 & +0.3 \\
 &  & Haider-Mean & $\lambda_H=0.9$ & 49.9 & 49.4 & 0.384 & 87.3 & 76.6 & 76.4 & +0.2 \\
 &  & Haider-Mean & $\lambda_H=0.95$ & 42.7 & 41.9 & 0.699 & 84.8 & 76.3 & 76.4 & -0.1 \\
 &  & GC & $\alpha=0.001$ & 95.3 & 95.7 & 0.019 & 98.4 & 76.5 & 76.4 & +0.1 \\
 &  & GC & $\alpha=0.003$ & 78.4 & 84.0 & 0.112 & 96.1 & 76.4 & 76.4 & +0.0 \\
 &  & GC & $\alpha=0.01$ & 74.0 & 72.6 & 0.292 & 91.3 & 75.1 & 76.4 & -1.3 \\
 &  & ShortV & $L_f=6$ & 96.7 & 100.0 & 0.001 & 99.9 & 76.5 & 76.4 & +0.1 \\
 &  & ShortV & $L_f=10$ & 94.0 & 100.0 & 0.001 & 99.8 & 76.6 & 76.4 & +0.2 \\
 &  & ShortV & $L_f=14$ & 91.3 & 100.0 & 0.018 & 99.1 & 76.6 & 76.4 & +0.2 \\
 &  & ShortV & $L_f=18$ & 88.6 & 100.0 & 0.025 & 99.0 & 76.6 & 76.4 & +0.2 \\
 &  & ShortV & $L_f=22$ & 86.0 & 100.0 & 0.219 & 96.1 & 76.7 & 76.4 & +0.3 \\
 &  & ShortV & $L_f=26$ & 83.3 & 100.0 & 1.502 & 84.6 & 72.9 & 76.4 & -3.5 \\
 &  & ShortV & $L_f=30$ & 80.6 & 100.0 & 1.702 & 77.6 & 69.2 & 76.4 & -7.2 \\
 &  & FLAP & keep 0.6 & 60.6 & 60.9 & 2.050 & 51.9 & 50.8 & 76.4 & -25.6 \\
 &  & FLAP & keep 0.5 & 51.2 & 50.8 & 2.868 & 35.3 & 36.5 & 76.4 & -39.9 \\
 &  & FLAP & keep 0.4 & 41.4 & 40.9 & 2.432 & 35.3 & 36.5 & 76.4 & -39.9 \\
 &  & FLAP & keep 0.3 & 31.4 & 31.0 & 3.095 & 35.3 & 36.5 & 76.4 & -39.9 \\
 &  & Wanda-sp & keep 0.6 & 61.7 & 61.6 & 2.138 & 35.8 & 36.4 & 76.4 & -40.0 \\
 &  & Wanda-sp & keep 0.5 & 51.0 & 51.0 & 4.034 & 35.3 & 36.5 & 76.4 & -39.9 \\
 &  & Wanda-sp & keep 0.4 & 43.1 & 42.9 & 2.846 & 34.7 & 35.8 & 76.4 & -40.6 \\
 &  & Wanda-sp & keep 0.3 & 32.5 & 32.3 & 2.875 & 35.4 & 36.3 & 76.4 & -40.1 \\
 &  & Random & seed 0 & 18.0 & 38.5 & 3.317 & 35.3 & 36.5 & 76.4 & -39.9 \\
 &  & Random & seed 2 & 16.5 & 38.6 & 3.361 & 36.9 & 37.0 & 76.4 & -39.4 \\
 &  & Random & seed 1 & 16.1 & 38.5 & 3.362 & 35.3 & 36.5 & 76.4 & -39.9 \\
\midrule
Gemma-3-4B & POPE & Ours & $\tol=0.25$ & 13.6 & 19.0 & 0.188 & 93.3 & 86.3 & 85.7 & +0.6 \\
 &  & Ours & $\tol=0.15$ & 22.6 & 31.9 & 0.122 & 94.7 & 86.4 & 85.7 & +0.7 \\
 &  & Haider-Mean & $\lambda_H=0.8$ & 36.6 & 42.8 & 0.152 & 94.3 & 86.1 & 85.7 & +0.4 \\
 &  & Haider-Mean & $\lambda_H=0.9$ & 31.3 & 34.7 & 0.206 & 93.0 & 86.5 & 85.7 & +0.8 \\
 &  & Haider-Mean & $\lambda_H=0.95$ & 20.7 & 22.0 & 0.268 & 91.7 & 85.8 & 85.7 & +0.1 \\
 &  & GC & $\alpha=0.001$ & 72.4 & 73.2 & 0.038 & 98.8 & 86.2 & 85.7 & +0.5 \\
 &  & GC & $\alpha=0.003$ & 68.2 & 66.9 & 0.122 & 96.3 & 86.2 & 85.7 & +0.5 \\
 &  & GC & $\alpha=0.01$ & 59.7 & 58.9 & 0.182 & 93.5 & 85.6 & 85.7 & -0.1 \\
 &  & ShortV & $L_f=6$ & 96.6 & 100.0 & 0.000 & 100.0 & 85.7 & 85.7 & +0.0 \\
 &  & ShortV & $L_f=10$ & 94.0 & 100.0 & 0.000 & 100.0 & 85.7 & 85.7 & +0.0 \\
 &  & ShortV & $L_f=14$ & 91.3 & 100.0 & 0.014 & 99.5 & 85.8 & 85.7 & +0.1 \\
 &  & ShortV & $L_f=18$ & 88.6 & 100.0 & 0.236 & 96.7 & 85.3 & 85.7 & -0.4 \\
 &  & ShortV & $L_f=22$ & 85.9 & 100.0 & 0.237 & 97.0 & 85.5 & 85.7 & -0.2 \\
 &  & ShortV & $L_f=26$ & 83.2 & 100.0 & 0.955 & 89.5 & 83.5 & 85.7 & -2.2 \\
 &  & ShortV & $L_f=30$ & 80.5 & 100.0 & 2.878 & 71.6 & 69.5 & 85.7 & -16.2 \\
 &  & FLAP & keep 0.6 & 59.9 & 60.8 & 2.086 & 53.5 & 54.8 & 85.7 & -30.9 \\
 &  & FLAP & keep 0.5 & 51.0 & 50.8 & 2.837 & 48.7 & 50.0 & 85.7 & -35.7 \\
 &  & FLAP & keep 0.4 & 41.3 & 40.9 & 2.035 & 48.7 & 50.0 & 85.7 & -35.7 \\
 &  & FLAP & keep 0.3 & 31.3 & 30.9 & 2.150 & 48.7 & 50.0 & 85.7 & -35.7 \\
 &  & Wanda-sp & keep 0.6 & 61.5 & 61.4 & 3.555 & 48.7 & 50.0 & 85.7 & -35.7 \\
 &  & Wanda-sp & keep 0.5 & 51.1 & 51.1 & 1.167 & 49.9 & 49.8 & 85.7 & -35.9 \\
 &  & Wanda-sp & keep 0.4 & 43.1 & 42.9 & 1.194 & 49.1 & 49.1 & 85.7 & -36.6 \\
 &  & Wanda-sp & keep 0.3 & 32.4 & 32.3 & 1.756 & 48.9 & 50.0 & 85.7 & -35.7 \\
 &  & Random & seed 0 & 9.1 & 19.7 & 0.821 & 48.7 & 50.0 & 85.7 & -35.7 \\
 &  & Random & seed 1 & 8.0 & 19.6 & 0.819 & 48.7 & 50.0 & 85.7 & -35.7 \\
 &  & Random & seed 2 & 7.8 & 19.6 & 0.965 & 48.7 & 50.0 & 85.7 & -35.7 \\
\midrule
Gemma-3-4B & AI2D & Ours & $\tol=0.25$ & 42.2 & 63.1 & 0.245 & 89.2 & 70.8 & 70.6 & +0.2 \\
 &  & Ours & $\tol=0.15$ & 71.2 & 82.1 & 0.134 & 94.4 & 70.6 & 70.6 & +0.0 \\
 &  & Haider-Mean & $\lambda_H=0.8$ & 69.0 & 69.0 & 0.383 & 83.6 & 70.5 & 70.6 & -0.1 \\
 &  & Haider-Mean & $\lambda_H=0.9$ & 54.9 & 54.6 & 0.452 & 82.9 & 69.6 & 70.6 & -1.0 \\
 &  & Haider-Mean & $\lambda_H=0.95$ & 48.4 & 47.3 & 0.597 & 79.8 & 68.5 & 70.6 & -2.1 \\
 &  & GC & $\alpha=0.001$ & 95.5 & 95.9 & 0.037 & 97.3 & 70.8 & 70.6 & +0.2 \\
 &  & GC & $\alpha=0.003$ & 88.4 & 89.1 & 0.088 & 95.6 & 70.3 & 70.6 & -0.3 \\
 &  & GC & $\alpha=0.01$ & 74.7 & 75.6 & 0.340 & 90.8 & 69.2 & 70.6 & -1.4 \\
 &  & ShortV & $L_f=6$ & 96.8 & 100.0 & 0.001 & 99.6 & 70.6 & 70.6 & +0.0 \\
 &  & ShortV & $L_f=10$ & 94.2 & 100.0 & 0.005 & 99.2 & 70.6 & 70.6 & +0.0 \\
 &  & ShortV & $L_f=14$ & 91.6 & 100.0 & 0.034 & 98.2 & 70.8 & 70.6 & +0.2 \\
 &  & ShortV & $L_f=18$ & 89.1 & 100.0 & 0.120 & 96.2 & 70.7 & 70.6 & +0.1 \\
 &  & ShortV & $L_f=22$ & 86.5 & 100.0 & 0.402 & 93.0 & 69.9 & 70.6 & -0.7 \\
 &  & ShortV & $L_f=26$ & 83.9 & 100.0 & 2.952 & 70.2 & 60.1 & 70.6 & -10.5 \\
 &  & ShortV & $L_f=30$ & 81.4 & 100.0 & 3.586 & 60.9 & 56.3 & 70.6 & -14.3 \\
 &  & FLAP & keep 0.6 & 61.1 & 60.8 & 2.555 & 20.7 & 24.9 & 70.6 & -45.7 \\
 &  & FLAP & keep 0.5 & 51.5 & 50.9 & 2.341 & 21.5 & 25.1 & 70.6 & -45.5 \\
 &  & FLAP & keep 0.4 & 41.6 & 41.0 & 2.055 & 20.8 & 25.1 & 70.6 & -45.5 \\
 &  & FLAP & keep 0.3 & 31.5 & 31.0 & 3.437 & 20.1 & 24.5 & 70.6 & -46.1 \\
 &  & Wanda-sp & keep 0.6 & 61.5 & 61.4 & 2.880 & 20.9 & 25.0 & 70.6 & -45.6 \\
 &  & Wanda-sp & keep 0.5 & 51.1 & 51.1 & 3.518 & 20.0 & 24.4 & 70.6 & -46.2 \\
 &  & Wanda-sp & keep 0.4 & 43.0 & 42.8 & 2.542 & 21.3 & 25.3 & 70.6 & -45.3 \\
 &  & Wanda-sp & keep 0.3 & 32.5 & 32.3 & 2.982 & 23.5 & 25.4 & 70.6 & -45.2 \\
 &  & Random & seed 2 & 28.3 & 63.4 & 1.407 & 29.1 & 26.0 & 70.6 & -44.6 \\
 &  & Random & seed 0 & 28.0 & 63.6 & 1.398 & 29.1 & 26.0 & 70.6 & -44.6 \\
 &  & Random & seed 1 & 26.8 & 63.6 & 1.550 & 29.1 & 26.0 & 70.6 & -44.6 \\
\midrule
Gemma-3-4B & MME & Ours & $\tol=0.25$ & 30.4 & 44.8 & 0.228 & 89.8 & 77.1 & 78.3 & -1.2 \\
 &  & Ours & $\tol=0.15$ & 55.9 & 77.9 & 0.137 & 93.3 & 77.9 & 78.3 & -0.4 \\
 &  & Haider-Mean & $\lambda_H=0.8$ & 55.8 & 55.3 & 0.247 & 88.4 & 77.1 & 78.3 & -1.2 \\
 &  & Haider-Mean & $\lambda_H=0.9$ & 40.4 & 44.7 & 0.292 & 88.0 & 76.6 & 78.3 & -1.7 \\
 &  & Haider-Mean & $\lambda_H=0.95$ & 34.7 & 36.7 & 0.321 & 86.7 & 77.2 & 78.3 & -1.1 \\
 &  & GC & $\alpha=0.001$ & 83.0 & 88.5 & 0.009 & 99.3 & 78.2 & 78.3 & -0.1 \\
 &  & GC & $\alpha=0.003$ & 72.2 & 73.1 & 0.066 & 97.5 & 78.5 & 78.3 & +0.2 \\
 &  & GC & $\alpha=0.01$ & 65.7 & 66.8 & 0.270 & 90.1 & 77.5 & 78.3 & -0.8 \\
 &  & ShortV & $L_f=6$ & 96.8 & 100.0 & 0.000 & 99.9 & 78.4 & 78.3 & +0.1 \\
 &  & ShortV & $L_f=10$ & 94.3 & 100.0 & 0.000 & 99.9 & 78.4 & 78.3 & +0.1 \\
 &  & ShortV & $L_f=14$ & 91.7 & 100.0 & 0.004 & 99.8 & 78.5 & 78.3 & +0.2 \\
 &  & ShortV & $L_f=18$ & 89.2 & 100.0 & 0.115 & 97.1 & 78.7 & 78.3 & +0.4 \\
 &  & ShortV & $L_f=22$ & 86.6 & 100.0 & 0.367 & 94.1 & 78.0 & 78.3 & -0.3 \\
 &  & ShortV & $L_f=26$ & 84.1 & 100.0 & 1.473 & 84.0 & 74.3 & 78.3 & -4.0 \\
 &  & ShortV & $L_f=30$ & 81.6 & 100.0 & 1.721 & 76.1 & 70.7 & 78.3 & -7.6 \\
 &  & FLAP & keep 0.6 & 61.0 & 60.7 & 0.842 & 50.7 & 51.4 & 78.3 & -26.9 \\
 &  & FLAP & keep 0.5 & 51.2 & 50.8 & 1.234 & 52.2 & 50.1 & 78.3 & -28.2 \\
 &  & FLAP & keep 0.4 & 41.3 & 40.9 & 1.519 & 51.8 & 50.0 & 78.3 & -28.3 \\
 &  & FLAP & keep 0.3 & 31.3 & 31.0 & 1.299 & 51.8 & 50.0 & 78.3 & -28.3 \\
 &  & Wanda-sp & keep 0.6 & 61.5 & 61.4 & 2.083 & 51.7 & 49.9 & 78.3 & -28.4 \\
 &  & Wanda-sp & keep 0.5 & 51.0 & 51.0 & 2.620 & 51.8 & 50.0 & 78.3 & -28.3 \\
 &  & Wanda-sp & keep 0.4 & 43.0 & 42.8 & 2.174 & 51.8 & 50.0 & 78.3 & -28.3 \\
 &  & Wanda-sp & keep 0.3 & 32.4 & 32.2 & 1.899 & 52.0 & 49.7 & 78.3 & -28.6 \\
 &  & Random & seed 0 & 20.0 & 45.2 & 0.788 & 48.2 & 50.0 & 78.3 & -28.3 \\
 &  & Random & seed 2 & 18.7 & 45.3 & 0.687 & 48.2 & 50.0 & 78.3 & -28.3 \\
 &  & Random & seed 1 & 18.1 & 45.3 & 0.685 & 46.5 & 49.9 & 78.3 & -28.4 \\
\midrule
Qwen3-VL-8B & GQA & Ours & $\tol=0.25$ & 16.9 & 28.2 & 0.212 & 89.8 & 86.0 & 87.1 & -1.1 \\
 &  & Ours & $\tol=0.15$ & 23.9 & 37.3 & 0.139 & 91.8 & 86.7 & 87.1 & -0.4 \\
 &  & Haider-Mean & $\lambda_H=0.8$ & 50.6 & 50.1 & 0.136 & 92.0 & 87.4 & 87.1 & +0.3 \\
 &  & Haider-Mean & $\lambda_H=0.9$ & 43.5 & 43.0 & 0.184 & 90.2 & 86.6 & 87.1 & -0.5 \\
 &  & Haider-Mean & $\lambda_H=0.95$ & 35.0 & 34.6 & 0.290 & 87.9 & 85.2 & 87.1 & -1.9 \\
 &  & GC & $\alpha=0.001$ & 84.7 & 85.6 & 0.110 & 94.3 & 86.8 & 87.1 & -0.3 \\
 &  & GC & $\alpha=0.003$ & 75.1 & 76.3 & 0.243 & 89.3 & 84.0 & 87.1 & -3.1 \\
 &  & GC & $\alpha=0.01$ & 65.7 & 76.1 & 0.849 & 66.7 & 65.4 & 87.1 & -21.7 \\
 &  & ShortV & $L_f=6$ & 95.8 & 100.0 & 0.000 & 99.9 & 87.1 & 87.1 & +0.0 \\
 &  & ShortV & $L_f=10$ & 92.4 & 100.0 & 0.000 & 99.8 & 87.2 & 87.1 & +0.1 \\
 &  & ShortV & $L_f=14$ & 89.0 & 100.0 & 0.000 & 99.7 & 87.1 & 87.1 & +0.0 \\
 &  & ShortV & $L_f=18$ & 85.6 & 100.0 & 0.011 & 98.2 & 87.2 & 87.1 & +0.1 \\
 &  & ShortV & $L_f=22$ & 82.2 & 100.0 & 0.061 & 96.2 & 86.1 & 87.1 & -1.0 \\
 &  & ShortV & $L_f=26$ & 78.8 & 100.0 & 0.323 & 89.0 & 81.9 & 87.1 & -5.2 \\
 &  & ShortV & $L_f=30$ & 75.4 & 100.0 & 0.632 & 82.2 & 77.4 & 87.1 & -9.7 \\
 &  & FLAP & keep 0.6 & 60.1 & 60.7 & 1.302 & 47.6 & 48.1 & 87.1 & -39.0 \\
 &  & FLAP & keep 0.5 & 50.2 & 51.1 & 1.530 & 35.0 & 35.6 & 87.1 & -51.5 \\
 &  & FLAP & keep 0.4 & 40.1 & 41.3 & 1.839 & 35.6 & 36.5 & 87.1 & -50.6 \\
 &  & FLAP & keep 0.3 & 29.8 & 31.3 & 2.071 & 35.6 & 36.5 & 87.1 & -50.6 \\
 &  & Wanda-sp & keep 0.6 & 61.8 & 61.7 & 0.722 & 73.6 & 72.5 & 87.1 & -14.6 \\
 &  & Wanda-sp & keep 0.5 & 51.5 & 51.5 & 2.514 & 37.3 & 37.1 & 87.1 & -50.0 \\
 &  & Wanda-sp & keep 0.4 & 41.8 & 41.8 & 1.792 & 35.3 & 35.4 & 87.1 & -51.7 \\
 &  & Wanda-sp & keep 0.3 & 31.9 & 31.9 & 1.630 & 35.5 & 36.0 & 87.1 & -51.1 \\
\midrule
Qwen3-VL-8B & POPE & Ours & $\tol=0.25$ & 5.1 & 8.1 & 0.132 & 93.7 & 88.9 & 88.4 & +0.5 \\
 &  & Ours & $\tol=0.15$ & 7.4 & 11.2 & 0.092 & 95.1 & 89.2 & 88.4 & +0.8 \\
 &  & Haider-Mean & $\lambda_H=0.8$ & 35.6 & 35.2 & 0.073 & 95.1 & 89.1 & 88.4 & +0.7 \\
 &  & Haider-Mean & $\lambda_H=0.9$ & 26.0 & 25.8 & 0.097 & 94.3 & 89.1 & 88.4 & +0.7 \\
 &  & Haider-Mean & $\lambda_H=0.95$ & 21.0 & 20.9 & 0.115 & 93.9 & 89.2 & 88.4 & +0.8 \\
 &  & GC & $\alpha=0.001$ & 61.6 & 69.4 & 0.072 & 95.8 & 88.4 & 88.4 & +0.0 \\
 &  & GC & $\alpha=0.003$ & 57.3 & 61.0 & 0.151 & 93.2 & 87.5 & 88.4 & -0.9 \\
 &  & GC & $\alpha=0.01$ & 48.1 & 52.1 & 0.656 & 54.5 & 50.0 & 88.4 & -38.4 \\
 &  & ShortV & $L_f=6$ & 96.4 & 100.0 & 0.000 & 99.9 & 88.4 & 88.4 & +0.0 \\
 &  & ShortV & $L_f=10$ & 93.5 & 100.0 & 0.000 & 99.9 & 88.4 & 88.4 & +0.0 \\
 &  & ShortV & $L_f=14$ & 90.6 & 100.0 & 0.000 & 99.8 & 88.3 & 88.4 & -0.1 \\
 &  & ShortV & $L_f=18$ & 87.7 & 100.0 & 0.011 & 98.6 & 88.2 & 88.4 & -0.2 \\
 &  & ShortV & $L_f=22$ & 84.8 & 100.0 & 0.039 & 97.7 & 88.0 & 88.4 & -0.4 \\
 &  & ShortV & $L_f=26$ & 82.0 & 100.0 & 0.060 & 96.6 & 87.5 & 88.4 & -0.9 \\
 &  & ShortV & $L_f=30$ & 79.1 & 100.0 & 0.133 & 94.9 & 87.0 & 88.4 & -1.4 \\
 &  & FLAP & keep 0.6 & 60.1 & 60.7 & 0.488 & 69.8 & 71.7 & 88.4 & -16.7 \\
 &  & FLAP & keep 0.5 & 50.2 & 51.0 & 1.178 & 45.5 & 50.0 & 88.4 & -38.4 \\
 &  & FLAP & keep 0.4 & 40.0 & 41.2 & 0.930 & 45.5 & 50.0 & 88.4 & -38.4 \\
 &  & FLAP & keep 0.3 & 29.9 & 31.4 & 3.044 & 45.5 & 50.0 & 88.4 & -38.4 \\
 &  & Wanda-sp & keep 0.6 & 61.8 & 61.8 & 0.330 & 84.4 & 83.9 & 88.4 & -4.5 \\
 &  & Wanda-sp & keep 0.5 & 51.6 & 51.6 & 1.284 & 47.7 & 50.9 & 88.4 & -37.5 \\
 &  & Wanda-sp & keep 0.4 & 41.8 & 41.8 & 1.236 & 45.3 & 49.9 & 88.4 & -38.5 \\
 &  & Wanda-sp & keep 0.3 & 31.9 & 31.8 & 0.900 & 50.2 & 48.8 & 88.4 & -39.6 \\
\midrule
Qwen3-VL-8B & AI2D & Ours & $\tol=0.25$ & 32.8 & 44.9 & 0.257 & 87.3 & 81.9 & 83.9 & -2.0 \\
 &  & Ours & $\tol=0.15$ & 37.7 & 50.5 & 0.160 & 90.5 & 83.6 & 83.9 & -0.3 \\
 &  & Haider-Mean & $\lambda_H=0.8$ & 55.7 & 55.9 & 0.194 & 89.5 & 83.5 & 83.9 & -0.4 \\
 &  & Haider-Mean & $\lambda_H=0.9$ & 49.5 & 49.0 & 0.266 & 87.0 & 81.6 & 83.9 & -2.3 \\
 &  & Haider-Mean & $\lambda_H=0.95$ & 44.6 & 43.9 & 0.363 & 82.8 & 78.7 & 83.9 & -5.2 \\
 &  & GC & $\alpha=0.001$ & 81.9 & 82.7 & 0.169 & 91.9 & 80.9 & 83.9 & -3.0 \\
 &  & GC & $\alpha=0.003$ & 66.2 & 75.9 & 0.451 & 79.8 & 74.2 & 83.9 & -9.7 \\
 &  & GC & $\alpha=0.01$ & 54.9 & 60.8 & 1.163 & 41.0 & 37.9 & 83.9 & -46.0 \\
 &  & ShortV & $L_f=6$ & 97.0 & 100.0 & 0.000 & 99.6 & 83.9 & 83.9 & +0.0 \\
 &  & ShortV & $L_f=10$ & 94.6 & 100.0 & 0.000 & 99.5 & 83.8 & 83.9 & -0.1 \\
 &  & ShortV & $L_f=14$ & 92.2 & 100.0 & 0.001 & 99.1 & 83.7 & 83.9 & -0.2 \\
 &  & ShortV & $L_f=18$ & 89.8 & 100.0 & 0.079 & 95.0 & 83.3 & 83.9 & -0.6 \\
 &  & ShortV & $L_f=22$ & 87.4 & 100.0 & 0.438 & 84.3 & 76.7 & 83.9 & -7.2 \\
 &  & ShortV & $L_f=26$ & 85.0 & 100.0 & 0.551 & 78.5 & 72.0 & 83.9 & -11.9 \\
 &  & ShortV & $L_f=30$ & 82.6 & 100.0 & 0.833 & 72.6 & 67.6 & 83.9 & -16.3 \\
 &  & FLAP & keep 0.6 & 59.9 & 60.5 & 2.430 & 22.2 & 25.0 & 83.9 & -58.9 \\
 &  & FLAP & keep 0.5 & 50.0 & 50.8 & 2.152 & 22.0 & 24.0 & 83.9 & -59.9 \\
 &  & FLAP & keep 0.4 & 39.8 & 41.0 & 2.400 & 23.0 & 25.4 & 83.9 & -58.5 \\
 &  & FLAP & keep 0.3 & 29.6 & 31.0 & 2.834 & 24.4 & 25.7 & 83.9 & -58.2 \\
 &  & Wanda-sp & keep 0.6 & 61.9 & 61.8 & 1.499 & 49.5 & 48.1 & 83.9 & -35.8 \\
 &  & Wanda-sp & keep 0.5 & 51.5 & 51.5 & 2.228 & 22.6 & 24.6 & 83.9 & -59.3 \\
 &  & Wanda-sp & keep 0.4 & 41.7 & 41.7 & 1.969 & 22.2 & 24.3 & 83.9 & -59.6 \\
 &  & Wanda-sp & keep 0.3 & 31.9 & 31.9 & 2.646 & 22.7 & 24.2 & 83.9 & -59.7 \\
\midrule
Qwen3-VL-8B & MME & Ours & $\tol=0.25$ & 25.5 & 38.4 & 0.253 & 87.2 & 82.5 & 87.3 & -4.8 \\
 &  & Ours & $\tol=0.15$ & 36.5 & 51.3 & 0.158 & 91.6 & 85.8 & 87.3 & -1.5 \\
 &  & Haider-Mean & $\lambda_H=0.8$ & 48.2 & 47.7 & 0.161 & 91.6 & 86.4 & 87.3 & -0.9 \\
 &  & Haider-Mean & $\lambda_H=0.9$ & 43.1 & 43.0 & 0.236 & 88.6 & 84.3 & 87.3 & -3.0 \\
 &  & Haider-Mean & $\lambda_H=0.95$ & 36.4 & 36.1 & 0.370 & 82.3 & 76.5 & 87.3 & -10.8 \\
 &  & GC & $\alpha=0.001$ & 74.1 & 75.2 & 0.166 & 93.1 & 84.7 & 87.3 & -2.6 \\
 &  & GC & $\alpha=0.003$ & 61.8 & 67.6 & 0.273 & 86.7 & 80.5 & 87.3 & -6.8 \\
 &  & GC & $\alpha=0.01$ & 49.4 & 55.8 & 0.681 & 58.2 & 50.0 & 87.3 & -37.3 \\
 &  & ShortV & $L_f=6$ & 95.9 & 100.0 & 0.000 & 99.6 & 87.0 & 87.3 & -0.3 \\
 &  & ShortV & $L_f=10$ & 92.6 & 100.0 & 0.000 & 99.4 & 86.8 & 87.3 & -0.5 \\
 &  & ShortV & $L_f=14$ & 89.4 & 100.0 & 0.000 & 99.2 & 86.8 & 87.3 & -0.5 \\
 &  & ShortV & $L_f=18$ & 86.1 & 100.0 & 0.022 & 97.7 & 86.2 & 87.3 & -1.1 \\
 &  & ShortV & $L_f=22$ & 82.8 & 100.0 & 0.138 & 94.8 & 85.8 & 87.3 & -1.5 \\
 &  & ShortV & $L_f=26$ & 79.6 & 100.0 & 0.603 & 84.8 & 75.3 & 87.3 & -12.0 \\
 &  & ShortV & $L_f=30$ & 76.3 & 100.0 & 1.487 & 74.7 & 65.0 & 87.3 & -22.3 \\
 &  & FLAP & keep 0.6 & 60.0 & 60.5 & 2.189 & 41.8 & 50.0 & 87.3 & -37.3 \\
 &  & FLAP & keep 0.5 & 50.1 & 50.9 & 1.735 & 41.9 & 49.9 & 87.3 & -37.4 \\
 &  & FLAP & keep 0.4 & 40.0 & 41.2 & 1.401 & 42.2 & 49.6 & 87.3 & -37.7 \\
 &  & FLAP & keep 0.3 & 29.8 & 31.2 & 1.479 & 42.0 & 50.2 & 87.3 & -37.1 \\
 &  & Wanda-sp & keep 0.6 & 61.9 & 61.8 & 0.596 & 61.9 & 63.7 & 87.3 & -23.6 \\
 &  & Wanda-sp & keep 0.5 & 51.6 & 51.6 & 1.347 & 42.3 & 49.8 & 87.3 & -37.5 \\
 &  & Wanda-sp & keep 0.4 & 41.8 & 41.8 & 1.478 & 43.8 & 51.0 & 87.3 & -36.3 \\
 &  & Wanda-sp & keep 0.3 & 31.9 & 31.9 & 1.116 & 50.5 & 49.7 & 87.3 & -37.6 \\
\midrule
InternVL3-8B & GQA & Ours & $\tol=0.25$ & 13.9 & 21.2 & 0.222 & 86.9 & 83.5 & 87.0 & -3.5 \\
 &  & Ours & $\tol=0.15$ & 19.8 & 29.1 & 0.146 & 89.9 & 85.2 & 87.0 & -1.8 \\
 &  & Haider-Mean & $\lambda_H=0.8$ & 52.1 & 51.9 & 0.115 & 92.0 & 87.2 & 87.0 & +0.2 \\
 &  & Haider-Mean & $\lambda_H=0.9$ & 44.2 & 43.2 & 0.179 & 88.7 & 85.5 & 87.0 & -1.5 \\
 &  & Haider-Mean & $\lambda_H=0.95$ & 38.4 & 37.6 & 0.237 & 86.9 & 84.9 & 87.0 & -2.1 \\
 &  & GC & $\alpha=0.001$ & 90.0 & 90.6 & 0.101 & 92.6 & 86.2 & 87.0 & -0.8 \\
 &  & GC & $\alpha=0.003$ & 81.6 & 85.7 & 0.257 & 85.7 & 81.7 & 87.0 & -5.3 \\
 &  & GC & $\alpha=0.01$ & 76.5 & 80.9 & 0.734 & 63.3 & 63.0 & 87.0 & -24.0 \\
 &  & ShortV & $L_f=4$ & 97.3 & 100.0 & 0.000 & 99.7 & 87.1 & 87.0 & +0.1 \\
 &  & ShortV & $L_f=8$ & 93.8 & 100.0 & 0.001 & 99.5 & 87.1 & 87.0 & +0.1 \\
 &  & ShortV & $L_f=12$ & 90.2 & 100.0 & 0.009 & 98.2 & 86.8 & 87.0 & -0.2 \\
 &  & ShortV & $L_f=16$ & 86.7 & 100.0 & 0.042 & 95.4 & 86.5 & 87.0 & -0.5 \\
 &  & ShortV & $L_f=20$ & 83.1 & 100.0 & 0.168 & 88.9 & 83.1 & 87.0 & -3.9 \\
 &  & ShortV & $L_f=24$ & 79.6 & 100.0 & 1.133 & 66.1 & 64.1 & 87.0 & -22.9 \\
\midrule
InternVL3-8B & POPE & Ours & $\tol=0.25$ & 4.1 & 5.9 & 0.136 & 92.4 & 88.7 & 89.5 & -0.8 \\
 &  & Ours & $\tol=0.15$ & 6.4 & 9.1 & 0.118 & 93.7 & 89.5 & 89.5 & +0.0 \\
 &  & Haider-Mean & $\lambda_H=0.8$ & 37.5 & 36.7 & 0.072 & 94.7 & 89.6 & 89.5 & +0.1 \\
 &  & Haider-Mean & $\lambda_H=0.9$ & 27.8 & 27.2 & 0.083 & 94.2 & 89.6 & 89.5 & +0.1 \\
 &  & Haider-Mean & $\lambda_H=0.95$ & 18.8 & 18.7 & 0.138 & 92.2 & 88.2 & 89.5 & -1.3 \\
 &  & GC & $\alpha=0.001$ & 84.6 & 88.7 & 0.064 & 95.6 & 88.6 & 89.5 & -0.9 \\
 &  & GC & $\alpha=0.003$ & 79.9 & 84.2 & 0.167 & 92.5 & 87.7 & 89.5 & -1.8 \\
 &  & GC & $\alpha=0.01$ & 68.6 & 73.3 & 0.630 & 54.6 & 50.0 & 89.5 & -39.5 \\
 &  & ShortV & $L_f=4$ & 97.3 & 100.0 & 0.000 & 99.7 & 89.5 & 89.5 & +0.0 \\
 &  & ShortV & $L_f=8$ & 93.6 & 100.0 & 0.001 & 99.7 & 89.4 & 89.5 & -0.1 \\
 &  & ShortV & $L_f=12$ & 90.0 & 100.0 & 0.002 & 99.3 & 89.5 & 89.5 & +0.0 \\
 &  & ShortV & $L_f=16$ & 86.3 & 100.0 & 0.021 & 97.6 & 88.9 & 89.5 & -0.6 \\
 &  & ShortV & $L_f=20$ & 82.7 & 100.0 & 0.325 & 86.4 & 81.1 & 89.5 & -8.4 \\
 &  & ShortV & $L_f=24$ & 79.0 & 100.0 & 1.979 & 60.5 & 55.8 & 89.5 & -33.7 \\
\midrule
InternVL3-8B & AI2D & Ours & $\tol=0.25$ & 33.3 & 46.2 & 0.250 & 84.6 & 79.0 & 83.0 & -4.0 \\
 &  & Ours & $\tol=0.15$ & 38.9 & 51.5 & 0.174 & 88.0 & 80.6 & 83.0 & -2.4 \\
 &  & Haider-Mean & $\lambda_H=0.8$ & 65.8 & 64.7 & 0.197 & 86.5 & 81.0 & 83.0 & -2.0 \\
 &  & Haider-Mean & $\lambda_H=0.9$ & 58.5 & 56.8 & 0.232 & 85.6 & 80.3 & 83.0 & -2.7 \\
 &  & Haider-Mean & $\lambda_H=0.95$ & 51.9 & 50.5 & 0.383 & 79.3 & 75.3 & 83.0 & -7.7 \\
 &  & GC & $\alpha=0.001$ & 91.0 & 91.5 & 0.174 & 87.5 & 79.4 & 83.0 & -3.6 \\
 &  & GC & $\alpha=0.003$ & 82.8 & 83.7 & 0.394 & 79.4 & 73.8 & 83.0 & -9.2 \\
 &  & GC & $\alpha=0.01$ & 80.2 & 81.5 & 0.843 & 58.8 & 57.3 & 83.0 & -25.7 \\
 &  & ShortV & $L_f=4$ & 97.5 & 100.0 & 0.001 & 99.2 & 82.7 & 83.0 & -0.3 \\
 &  & ShortV & $L_f=8$ & 94.2 & 100.0 & 0.012 & 97.1 & 82.5 & 83.0 & -0.5 \\
 &  & ShortV & $L_f=12$ & 90.8 & 100.0 & 0.022 & 95.9 & 81.9 & 83.0 & -1.1 \\
 &  & ShortV & $L_f=16$ & 87.5 & 100.0 & 0.142 & 89.2 & 80.2 & 83.0 & -2.8 \\
 &  & ShortV & $L_f=20$ & 84.2 & 100.0 & 0.506 & 75.8 & 71.1 & 83.0 & -11.9 \\
 &  & ShortV & $L_f=24$ & 80.8 & 100.0 & 0.775 & 66.8 & 64.1 & 83.0 & -18.9 \\
\midrule
InternVL3-8B & MME & Ours & $\tol=0.25$ & 25.1 & 36.8 & 0.229 & 84.1 & 80.1 & 85.1 & -5.0 \\
 &  & Ours & $\tol=0.15$ & 39.7 & 55.9 & 0.105 & 92.8 & 84.1 & 85.1 & -1.0 \\
 &  & Haider-Mean & $\lambda_H=0.8$ & 56.2 & 54.7 & 0.168 & 87.9 & 83.6 & 85.1 & -1.5 \\
 &  & Haider-Mean & $\lambda_H=0.9$ & 49.0 & 47.9 & 0.216 & 86.3 & 81.6 & 85.1 & -3.5 \\
 &  & Haider-Mean & $\lambda_H=0.95$ & 38.4 & 37.5 & 0.324 & 80.7 & 78.2 & 85.1 & -6.9 \\
 &  & GC & $\alpha=0.001$ & 86.3 & 90.1 & 0.108 & 91.9 & 82.7 & 85.1 & -2.4 \\
 &  & GC & $\alpha=0.003$ & 80.6 & 84.9 & 0.249 & 82.7 & 76.7 & 85.1 & -8.4 \\
 &  & GC & $\alpha=0.01$ & 48.7 & 53.9 & 0.611 & 58.2 & 50.0 & 85.1 & -35.1 \\
 &  & ShortV & $L_f=4$ & 97.4 & 100.0 & 0.000 & 99.7 & 85.0 & 85.1 & -0.1 \\
 &  & ShortV & $L_f=8$ & 94.0 & 100.0 & 0.000 & 99.3 & 85.2 & 85.1 & +0.1 \\
 &  & ShortV & $L_f=12$ & 90.6 & 100.0 & 0.003 & 99.3 & 85.1 & 85.1 & +0.0 \\
 &  & ShortV & $L_f=16$ & 87.1 & 100.0 & 0.055 & 95.4 & 83.1 & 85.1 & -2.0 \\
 &  & ShortV & $L_f=20$ & 83.7 & 100.0 & 0.288 & 85.4 & 74.2 & 85.1 & -10.9 \\
 &  & ShortV & $L_f=24$ & 80.3 & 100.0 & 0.527 & 78.8 & 69.3 & 85.1 & -15.8 \\
\end{longtable}}

\end{document}